%% file: main.tex
\documentclass[12pt,letterpaper, runningheads]{llncs}

\usepackage{eccv}
\usepackage{eccvabbrv}

\usepackage[margin=1.0in, letterpaper, total={6.5in,9in}, centering]{geometry}
\usepackage{graphicx}
\usepackage{booktabs}
\usepackage{amsmath}
\usepackage{algorithm}
\usepackage{algpseudocode}
\usepackage{mathtools}
\usepackage{wrapfig}
\newcommand{\method}{DrivingVoxels}

\algrenewcommand\algorithmiccomment[1] {\hfill$\triangleright$ #1}
\usepackage[accsupp]{axessibility}  

\usepackage{hyperref}
\usepackage{multirow} 
\usepackage{orcidlink}

\usepackage{xcolor}

\definecolor{citecolor}{rgb}{0.0,1.0,0.0}   
\definecolor{linkcolor}{rgb}{1.0,0.0,0.0}   
\definecolor{urlcolor}{rgb}{0.0,0.0,1.0}    

\hypersetup{
    colorlinks=true,
    citecolor=citecolor,
    linkcolor=linkcolor,
    urlcolor=urlcolor
}

\usepackage{caption}
\makeatletter
\renewcommand\section{\@startsection
  {section}{1}{\z@}
  {-18\p@ \@plus -4\p@ \@minus -4\p@}
  {12\p@ \@plus 4\p@ \@minus 4\p@}
  {\normalfont\Large\bfseries}}

\renewcommand\subsection{\@startsection
  {subsection}{2}{\z@}
  {-18\p@ \@plus -4\p@ \@minus -4\p@}
  {8\p@ \@plus 4\p@ \@minus 4\p@}
  {\normalfont\large\bfseries}}

\renewcommand\subsubsection{\@startsection
  {subsubsection}{3}{\z@}
  {-12\p@ \@plus -4\p@ \@minus -4\p@}
  {6\p@ \@plus 4\p@ \@minus 4\p@}
  {\normalfont\normalsize\bfseries}}
\makeatother

\usepackage{overpic}

\begin{document}

\title{\method: Compositional Sparse Voxel Rasterization for Dynamic Driving Scene Reconstruction} 

\titlerunning{DrivingVoxels}

\author{
    \large Tania Aguirre$^{1,2}$ \qquad
     Luis Roldão$ ^{1}$ \qquad \\
    \large Moussab Bennehar$^{1}$ \qquad
    Nathan Piasco$^{1}$ \qquad 
    \large Dzmitry Tsishkou$^{1}$ \qquad \\
    Simone Rossi $^{2}$  \qquad
    Pietro Michiardi$^{2}$ 
    \\
    }
\authorrunning{T.~Aguirre et al.}

\institute{\large Noah's Ark, Huawei Paris Research Center, France \and
EURECOM, France
}

\maketitle

\begin{figure}[h]
  \centering
  \begin{overpic}[width=1.00\linewidth]{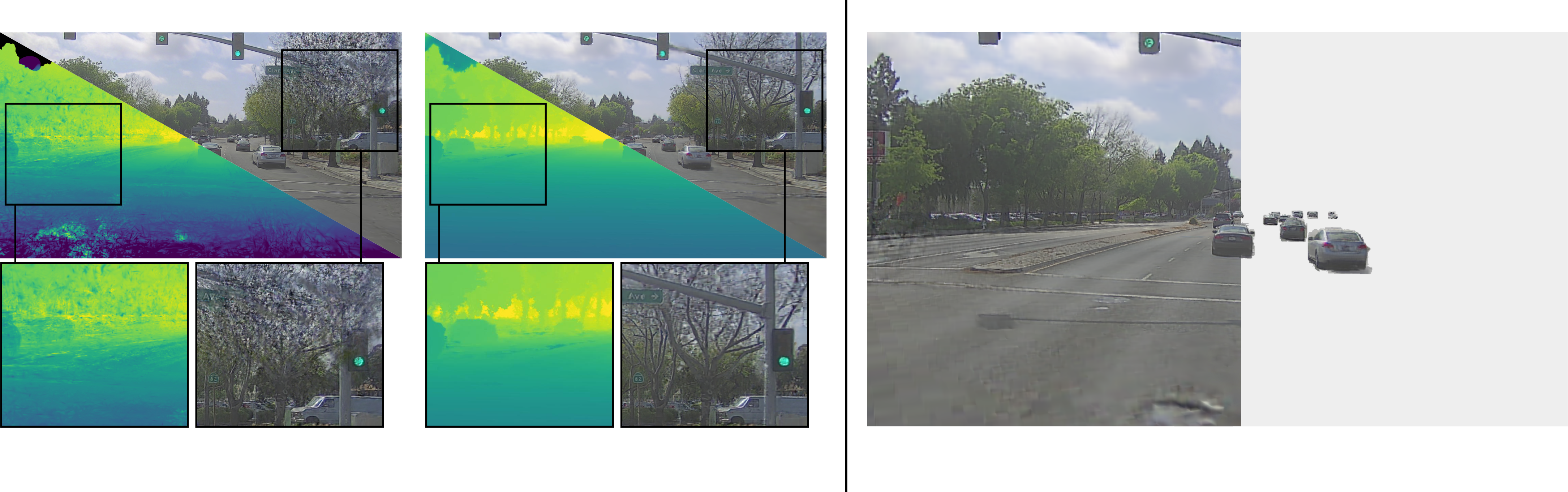}
    \put(7.0,30.3){
      \scriptsize StreetGS~\cite{yan2024street}
    }
    \put(30.4,30.3){
      \scriptsize\method~(ours)
    }
    \put(61.7,30.3){
      \scriptsize Static / Dynamic Scene Decomposition
    }
    \put(-0.5,2.5){
      \fontsize{5.5}{5}\selectfont
      $\text{Train Time:}~102~\text{min.}$ -- $\text{PSNR}=23.89$
    }
    \put(-0.5,0.7){
      \fontsize{5.5}{5}\selectfont $\text{CD}=0.19$ -- $\text{F1@0.1m}=31.8\%$
    }

    \put(26.5,2.5){
      \fontsize{5.5}{5}\selectfont $\text{Train Time:}~\textbf{30~\text{min.}}$ -- $\text{PSNR}=\textbf{23.90}$
    }
    \put(26.5,0.7){
      \fontsize{5.5}{5}\selectfont $\text{CD}=\textbf{0.06}$ -- $\text{F1@0.1m}=\textbf{59.9\%}$
    }
  \end{overpic}
  \caption{\textbf{Dynamic Scene Reconstruction and Decomposition Performance.} Compared to state-of-the-art 3DGS approaches such as StreetGS~\cite{yan2024street} (left), \method~(center) eliminates floating artifacts and geometric degradation, achieving superior structural accuracy ($\text{CD} = \mathbf{0.06}$, $\text{F1} = \mathbf{59.9\%}$) while reducing total optimization overhead by over $3.3\times$ (from 102 minutes down to 30 minutes). Our compositional multi-octree formulation cleanly decouples stationary background from independent moving actors (right).}
  \label{fig:teaser}
\end{figure}
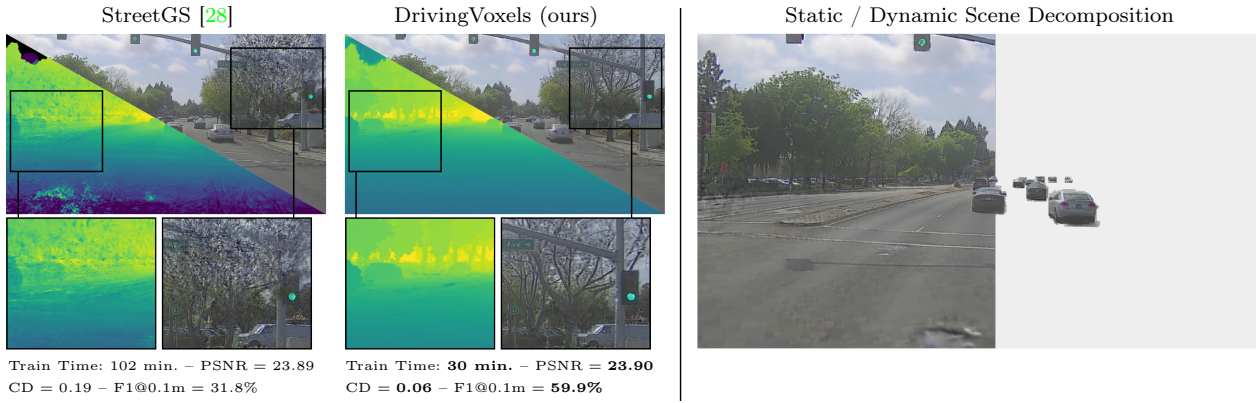
\input{secs/0_abstract}

\input{secs/1_intro}
\input{secs/2_related_work}

\input{secs/3_method}

\input{secs/4_experiments}

\input{secs/5_conclusion}

%
%
\bibliographystyle{splncs04}
\bibliography{main}
\newpage

\input{secs/supp}

\end{document}

%% file: secs/0_abstract.tex
\begin{abstract}
    Reconstructing dynamic urban scenes remains challenging due to the unbounded nature of driving environments and the presence of multiple dynamic objects.
    Currently, potentially faster sparse voxel methods are mainly designed for static scenarios. On the other hand, dynamic approaches based on 3D Gaussian Splatting, despite their high-fidelity, are often time-consuming for driving scenarios and exhibit uncontrollable memory growth in large scenes. 
    To address these limitations, we present \textbf{\method}, a compositional sparse voxel rendering framework for dynamic driving scenes. Our method jointly rasterizes sparse voxels from multiple independent octrees within a single rendering pass. Each rigid dynamic object is represented by an octree defined in its local coordinate frame, while a separate static octree models the stationary background. \method~adopts a fully explicit, neural-free representation together with a LiDAR-guided structural initialization that efficiently captures scene geometry. We evaluate our framework on the PandaSet benchmark, 
    demonstrating that \method~performs on par on perceptual metrics and better on structural metrics for NVS and reconstruction—while requiring shorter training times than previous 3DGS-base methods to an efficient optimization workflow anchored by a strong LiDAR prior.
\end{abstract}

%% file: secs/1_intro.tex
\section{Introduction}
Photorealistic reconstruction of driving environments is a key component for autonomous driving simulation, enabling applications such as novel view synthesis, sensor simulation, and closed-loop policy evaluation. Achieving this goal requires scene representations that are simultaneously scalable, geometrically consistent, and efficient to render.

Novel view synthesis (NVS) has seen rapid progress with the introduction of Neural Radiance Fields (NeRF)~\cite{mildenhall2021nerf} and 3D Gaussian Splatting (3DGS)~\cite{3dgs}. 
NeRF offers a continuous, physically grounded volumetric representation but is bottlenecked by expensive ray-marching. Conversely, 3DGS achieves real-time speeds via explicit anisotropic Gaussians splatting, but frequently suffers from floating artifacts due to a lack of structural consistency. Sparse Voxel Rasterization (SVRaster)~\cite{svr}  was recently proposed to combine 
the fast tile-based rasterization of 3DGS with explicit sparse voxel 
primitives, eliminating floating artifacts via Morton-ordered depth 
sorting while maintaining a well-structured volumetric representation similar to NeRF.

However, existing SVRaster formulations~\cite{svr, li2025geosvrtamingsparsevoxels, oh2025svreconsparsevoxelrasterization} assume a completely static world, preventing their use in autonomous driving simulations where scenes contain many moving actors like vehicles and pedestrians. This limitation comes from two main constraints in SVRaster. First, the efficient Morton-ordered depth sorting requires a single, fixed coordinate system, which means moving objects cannot easily be combined into a single global tree. Second, the sparse voxel octree structure is locked into place after it is created. While the system can prune or subdivide voxels during training, all new child voxels must still follow the rigid grid of the parent octree. Furthermore, gradients cannot change the actual positions of the voxel corners. As a result, a single static octree lacks the mathematical flexibility to track independent movements over time, restricting its capability entirely to static scenes.

Latest Gaussian-based dynamic methods for driving scenarios, such as OmniRe~\cite{chen2025omnire} and StreetGaussians~\cite{yan2024street}, address dynamic scenes by building scene graphs of per-object Gaussian representations in canonical spaces. While these methods are highly expressive, they still keep the main limitations of Gaussian primitives, such as ill-defined density fields, inaccurate surface normals, and unpredictable memory spikes during runtime. 
On the other hand, SaLF~\cite{chen2025salf} uses a sparse voxel layout for simulation but places a local implicit field inside each voxel node. This design requires explicit hierarchical octree traversal alongside costly neural network calculations, which means it reintroduces the slow computational bottlenecks that explicit primitives are supposed to eliminate.

Therefore, extending sparse voxel rasterization to dynamic urban environments requires overcoming the rigid single-octree assumption without sacrificing the efficiency advantages of voxel-based explicit rendering. To address such limitation, we introduce \textbf{\method}, a compositional sparse voxel rendering framework for dynamic driving  (\cref{fig:teaser}). Our method extends the original SVRaster architecture~\cite{svr} to handle independently moving actors without losing its neural-free, explicit representation. During rendering, we explicitly solve the depth ordering by computing the ray-box intersections between the camera rays and the bounding boxes of the moving vehicles. The traversal order between the static background voxels and the dynamic foreground voxels along each ray is fully defined by sorting these entry and exit points by camera distance. Using this sorting strategy, each moving actor is reconstructed in its own local canonical octree, mapped into world coordinates via rigid transformations, and smoothly composited into the final scene while correctly propagating transmittance across all octree boundaries.\method\ also scales the scene representation to match the layout of driving environments and integrates LiDAR data to guide the training process.  To summarize, the main contributions of our work are:
\begin{itemize}
    \item A novel compositional multi-octree voxel renderer that jointly rasterizes independent voxel volumes while resolving cross-volume depth ordering without merging tree structures.

    \item A LiDAR-guided background initialization pipeline combining visibility pruning, proximity-based voxel subdivision, and soft density priors to accelerate convergence in large outdoor scenes.

    \item Our method achieves competitive perceptual quality and state-of-the-art geometry reconstruction with faster training times on PandaSet~\cite{xiao2021pandaset}. 
    
\end{itemize}

%% file: secs/2_related_work.tex
\section{Related Work}
\paragraph{Neural Radiance Fields (NeRF).}
Neural Radiance Fields (NeRF)~\cite{mildenhall2021nerf} provide a foundation for photorealistic 3D scene reconstruction from 2D images by modeling scenes as continuous volumetric fields, mapping 3D coordinates and viewing directions to volumetric properties via a multilayer perceptron (MLP). 
While NeRF variants offer continuous, physically grounded representations, scaling them to large scenes is heavily constrained by the expensive computational overhead of volumetric ray-marching. Various explicit and hybrid structures have been introduced such as multiresolution hash encodings and tensor factorizations to accelerate this bottleneck~\cite{tensorf, muller2022instant, nsvf, plenoxels, sun2022direct}. Despite these optimization advancements, continuous volumetric sampling remains a fundamental challenge for scaling implicit representations to unbounded environments.  

\paragraph{3D Gaussian Splatting (3DGS).}
The introduction of 3D Gaussian Splatting (3DGS)~\cite{3dgs} shifted the research focus toward explicit point-based rasterization to achieve real-time performance, projecting learnable anisotropic Gaussians directly onto the image plane.
However, because multiple overlapping Gaussians can cover the same spatial coordinate, the actual volume density is mathematically ill-defined \cite{svr}. While this unconstrained optimization excels at interpolating training views, the lack of structural constraints frequently causes primitives to fall into disorganized spatial configurations under novel perspectives. This can lead to inconsistent surface normal approximations and severe geometric degradation when rendering extrapolated viewpoints. Furthermore, this spatial overlap introduces severe ambiguity during multi-modal feature fusion, causing semantic bleeding where distinct feature field values are blended across overlapping primitives \cite{wang2026lesv}. As a result, unconstrained point-based splatting is fundamentally ill-suited for downstream applications that require exact physical boundaries, structurally sound constraints, or discrete spatial partitioning, such as physical simulation \cite{gsslam}, autonomous navigation \cite{sugar}, or language-embedded scene understanding \cite{wang2026lesv}. 

To enforce strict structural boundaries, SVRaster~\cite{svr} models the scene using explicit, axis-aligned sparse voxels with discrete radiance values. This non-overlapping spatial partitioning  is rendered by a hardware-accelerated tile rasterization via Morton-ordered depth sorting, which successfully eliminates floating artifacts common in standard 3DGS. 
We build directly upon this explicit architecture to handle complex environments. A neural-free, bounded voxel representation provides deterministic control over scene complexity and a predictable quality-efficiency tradeoff unavailable to unconstrained Gaussian primitives.

\paragraph{Urban Scene Reconstruction.}

Photorealistic reconstruction of dynamic urban scenes is crucial for closed-loop autonomous driving simulation~\cite{wang2026jneus,Djeghim_2025_CVPR,wang2024planerf}. Early implicit frameworks separate scene elements into static backgrounds and dynamic agents via object-centric NeRF models to handle moving actors~\cite{scene-graph}, scaling up via multi-resolution hash tables in SUDS~\cite{turki2023suds} or modeling complex sensor characteristics in NeuRAD~\cite{tancik2023neurad}. However, these continuous volumetric methods remain bottlenecked by MLP ray-marching, preventing interactive simulation scales. While recent urban frameworks have pivoted toward explicit 3DGS~\cite{3dgs} to achieve real-time rendering and temporal dynamics through pipelines like StreetSurf~\cite{guo2023streetsurf}, PVG~\cite{chen2026periodic}, DrivingGaussian~\cite{zhou2024drivinggaussian}, StreetGaussians~\cite{yan2024street}, and OmniRe~\cite{chen2025omnire}, their unconstrained primitives introduce geometric floaters and unpredictable memory footprints.

As an alternative, sparse voxel layouts enforce strict physical boundaries to mitigate these geometric artifacts. Representing a major step forward in this category, SaLF \cite{chen2025salf} proposes a multi-sensor layout built upon 3D voxel primitives. However, SaLF relies on optimizing a hybrid local implicit field within each voxel node, which requires complex hierarchical traversals and neural queries. In contrast, our framework, \method, maintains a strictly neural-free, explicit voxel representation. By executing a ray-interval sorting strategy across independent, moving octrees, we resolve depth-correct composition and exact transmittance propagation.

%% file: secs/3_method.tex
\section{Method}

We tackle the challenge of dynamic urban scene synthesis by modeling the environment as a multi-octree composition of explicit sparse voxel fields.  Rather than using unconstrained point cloud primitives like 3D Gaussian Splatting~\cite{3dgs}, which lead to ill-defined density fields, floaters and unconstrained memory spikes, our framework builds on top of an explicit sparse voxel layout introduced by SVRaster~\cite{svr}.

\begin{figure}[!t]
    \centering
    \includegraphics[width=0.95\linewidth]{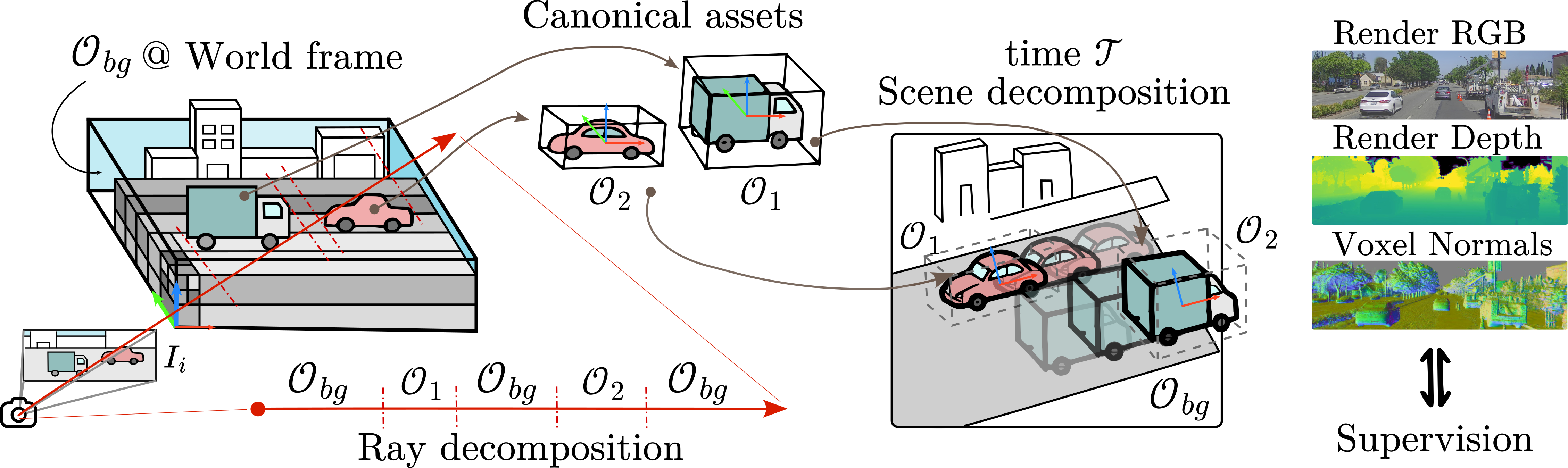}
   \caption{\textbf{Overview of the \method~pipeline.} Camera rays are cast into a compositional scene of a global static background octree $\mathcal{O}_{bg}$ in the world frame and independent dynamic assets octrees ($\mathcal{O}_{i}$) in local canonical spaces. The ray path is decomposed into sequential front-to-back segments based on 3D bounding box intersections. Background segments are integrated in world space, while asset segments are transformed into local coordinates via inverse tracking matrices. These segments are rasterized in a single pass while continuously propagating transmittance across coordinate boundaries to resolve correct depth ordering for final RGB, depth, and normal supervision.
   }   %
    \label{fig:pipeline}
\end{figure}

As illustrated in \cref{fig:pipeline}, our method allows us to decompose dynamic sequences into a high quality stationary background ($\mathcal{O}_{bg}$) representation in the world frame (\cref{sec:background}),  while simultaneously having a canonical asset representation ($\mathcal{O}_{i}$) for individual moving vehicles relying on tracking data (\cref{sec:dyn-assets}). After a proper initialization of each representation (\cref{sec:initialization}), we feed these octrees into our rendering function (\cref{sec:render}) to achieve scene reconstruction in a driving scenario. This architecture jointly rasterizes voxels from multiple independent grids in a single pass, correctly solving depth ordering and occlusion across octrees without requiring any modifications to their independent voxel structures.

\subsection{Static Background Modeling} \label{sec:background}

The global background is represented by an octree $\mathcal{O}_{bg}$ following the architectural primitives of SVRaster. The spatial layout of this hierarchical 3D voxel grid is aligned with an axis-aligned bounding box (AABB), parameterized by the global scene center $\mathbf{w}_\mathbf{c}$ and the root octree scale $\mathbf{w}_\mathbf{s}$ in world coordinates, which defines the boundaries of the scene. Each grid node corresponds to a voxel at a specific octree level that dictates its physical dimensions. Following SVRaster, only the active leaf nodes are preserved in memory without keeping their parent ancestor structures. The higher level of the octree defines the finest grid resolution for the scene. Each leaf node voxel stores a view-dependent appearance $\mathbf{c}$ parameterized via Spherical Harmonics (SH) coefficients, alongside discrete raw volume densities ($\mathbf{v}_\mathrm{geo}$) at its eight corners to define an internal continuous volume via trilinear interpolation. These corner densities and color are optimized during training. The structure is only modified during an adaptive pruning and progressive subdivision allowing the explicit voxel layout to dynamically refine and align with the structural details.

\paragraph{Sky Modeling.} Following prior work on sky-decoupled scene representation~\cite{chen2025omnire}, we model the sky as a 2D spherical environment map surrounding the scene. The sky color $\mathbf{C}_{\text{sky}}$ for a given camera ray is obtained by mapping its 3D viewing direction $\mathbf{d}$ into 2D map coordinates to sample the texture via bilinear interpolation. During training, the texture is updated via backpropagation. This design minimizes computational overhead by avoiding explicit voxel allocation, making it well-suited for unbounded outdoor autonomous driving environments.

\subsection{Dynamic Asset Modeling} \label{sec:dyn-assets}
Each dynamic asset octree $\mathcal{O}_{i}$ models the persistent geometry and view-dependent appearance of an individual moving actor. We adopt the core primitive layout from SVRaster~\cite{svr}, representing each asset in an isolated canonical space where only leaf nodes are preserved. In this setup, the origin is centered on the object’s 3D bounding box, and the coordinates are aligned with the box axes. We leverage the annotated bounding box dimensions as structural prior to bound the volume before optimization. We avoid initializing dynamic assets with LiDAR point cloud to mitigate optimization errors due to temporal sensor misalignment and bounding box inaccuracy.
 
The spatial state of $N$ dynamic assets $\{\mathcal{O}_{i}\}_{i=1}^N$ at any given timestep $t$ is defined by a set of time-varying rigid transformations $\{T_t^i \in SE(3)\}_{i=1}^N$ taken from 3D tracking annotations. At runtime, these matrices map each asset octree from its local canonical frame into the global world space. This allows our framework to aggregate sparse multi-view observations of moving objects across the entire driving sequence.

\subsection{Initialization}\label{sec:initialization}
To accurately model unbounded dynamic driving environments we propose a decoupled, multi-octree structural initialization strategy (\cref{fig:init}). We apply separate initialization priors for the static background and the independent dynamic actors.

\paragraph{Background.} Mirroring standard 3D Gaussian Splatting frameworks that naturally seed primitive positions using available LiDAR points or sparse keypoints, we extend this practice to explicit voxel layouts. We introduce a LiDAR-guided structural initialization pipeline:

\begin{figure}[!t]
    \centering
    
    \begin{overpic}[width=0.95\linewidth]{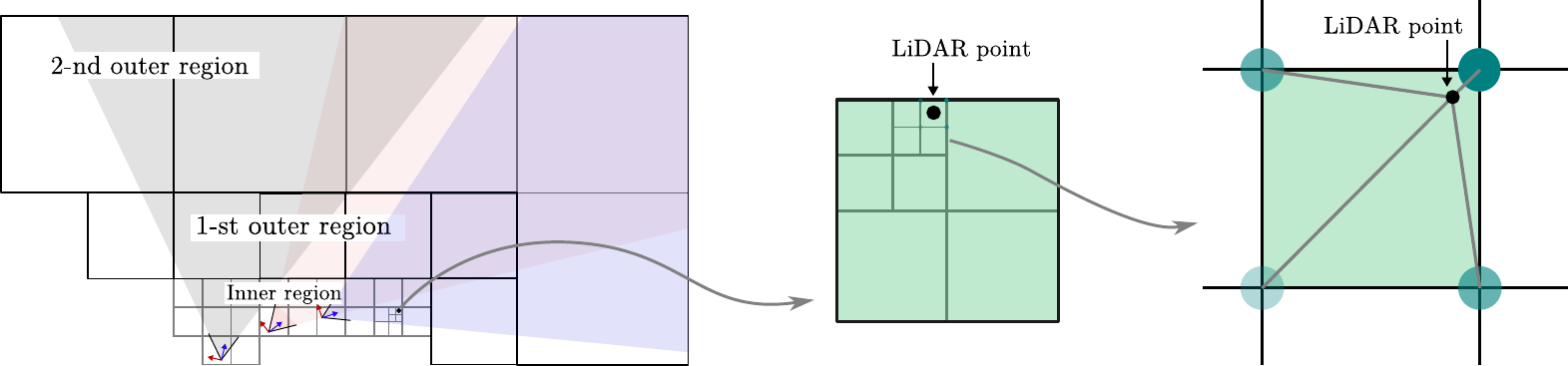}
    \put(21,-3.0){\small a)}
    \put(59.5,-3.0){\small b)}
    \put(87,-3.0){\small c)}
    \end{overpic}
    \vspace{0.2cm}

   \caption{\textbf{Overview of the \method~initialization.}We introduce a LiDAR-guided structural initialization pipeline: (a) 
   Inner and outer bound volumes are initialized considering camera poses and frustums akin SVRaster~\cite{svr} with additional height filtering to adapt to driving scenarios.
   (b) Leveraging LiDAR information, the octree is iteratively subdivided to create a finer surface shell around physical structures. (c) Raw corner densities are initialized based on an exponential decay prior according to the distance to the closest LiDAR point.}

    \label{fig:init}
\end{figure}

\begin{enumerate}
    \item \textbf{Visibility Pruning:} The volume is divided into an inner and an outer region following SVRaster implementation~\cite{svr}. The inner region models the main scene volume, containing all camera poses, while the outer region models far-away structures at a coarser resolution. The inner region is initialized first as a coarse uniform grid. Voxels are  pruned if they are not visible by any training camera or fall below a minimum camera-height reference threshold. The outer region is represented at two coarse resolution levels to capture far away structures, with the same frustum and height filter applied.
    \item \textbf{Proximity-Based Voxel Subdivision:} The remaining grid is then refined through rounds of iterative subdivision. In each round, voxels closer than a distance threshold to a LiDAR point are subdivided, producing a progressively finer shell on physical surface while retaining a coarse resolution in empty space. The subdivision continues until the minimum voxel sizes falls below $30$ cm.
    \item \textbf{Soft Density Initialization Prior:} Once the structural refinement is finished, voxel corner densities are initialized by proximity to the LiDAR point cloud. For each grid corner, the euclidean distance $d$ to the nearest LiDAR point is computed and mapped to an initial density: 
    \begin{equation}
        \mathbf{v}_{\mathrm{geo}}^{\mathrm{init}} = \mathbf{v}_{\mathrm{far}} + (\mathbf{v}_{\mathrm{close}} - \mathbf{v}_{\mathrm{far}}) \cdot \exp\left(-\frac{2d^2}{s^2}\right),
    \end{equation}

where $s$ is minimum voxel size. Voxel corners closer to the LiDAR points are initialized with a density  $\mathbf{v}_{\mathrm{close}}$  while distant corners are assigned an empty density $\mathbf{v}_{\mathrm{far}}$. We set heuristically $\mathbf{v}_{\mathrm{close}} = -2.0$ and  $\mathbf{v}_{\mathrm{far}} = -10.0$, the empty density defined by SVRaster~\cite{svr}. This approach creates smooth density gradients that provide the optimizer with a geometrically plausible starting point, even for coarse structures, while preserving the ability to recover surfaces missed by the LiDAR sensor. 
\end{enumerate}

\paragraph{Assets.} Unlike the static background, each asset begins as a uniform voxel grid aligned with bound local coordinates system, where the asset's 3D bounding box serves as the prior geometric information. The boundary of each asset matches the largest dimension of the bounding of the box, scaled outward by  $20\%$, to ensure a small margin around the asset geometry. Furthermore, all asset grids are initialized with a uniform empty density $\mathbf{v}_{\mathrm{geo}}^{\mathrm{init}} = -10$, independent of asset size or category. 

\subsection{Compositional Voxel Rendering}\label{sec:render}

To render a novel view at timestep $t$, we cast rays into our multi-octree scene representation. In the baseline framework, high rendering efficiency relies on a tile-based rasterizer that maps all intersected voxels into a single, strictly ordered front-to-back sequence via Morton ordering~\cite{svr}. However, because independent moving objects change coordinates arbitrarily over time, their voxels cannot be sorted within a single global tree without costly runtime tree reconstruction. To overcome this limitation and enable the representation of dynamic scenes,  we propose a macro-level ray-interval sorting strategy.

For each ray $\mathbf{r}(\tau) = \mathbf{o} + \tau\,\mathbf{d}$, we calculate ray-box intersections against the oriented 3D bounding boxes of all active dynamic assets (cf. \cref{fig:pipeline}, left), yielding camera distance intervals $\{\tau_{\text{in}}^i, \tau_{\text{out}}^i\}$. Sorting these entry and exit thresholds by camera distance maps the ray path into an ordered front-to-back sequence of different segments: 
\begin{itemize}
      \item \textbf{Background Segments:} Ray intervals falling entirely outside all dynamic bounding boxes are integrated using the static background octree $\mathcal{O}_{bg}$ in world space.
      \item \textbf{Asset Segments:} Within an active intersection interval $[\tau_{\text{in}}^i, \tau_{\text{out}}^i]$, the ray is transformed into the asset's local canonical frame via $(\mathbf{T}^i_t)^{-1}$ and evaluated inside the canonical octree $\mathcal{O}_i$.
\end{itemize}

Each consecutive ray segment is processed by an accumulation function $\mathcal{F}$, which performs local, hardware-accelerated Morton-ordered voxel traversals within that segment's bounded interval. Throughout this process, we track a running global pixel state along the ray path, including its color, transmittance, depth, and normal vectors (we refer to supplementary material 
\cref{sec:algorithm} for further details)
. For any intersected voxel within a segment, we adopt the standard numerical integration, exponential-linear density activation, and alpha-compositing formulas from the baseline framework~\cite{svr}. These localized segment properties are then blended step-by-step into the global pixel state, strictly weighted by the accumulated transmittance. Passing this running state continuously between consecutive segment calls ensures mathematically correct multi-volume occlusion and precise transmittance tracking across all coordinate boundaries.

After aggregating all foreground and background voxel segments into a total voxel color $\mathbf{C}_{\text{voxels}}$ and final residual transmittance $T_{\text{final}}$, the final pixel color representation $\mathbf{C}$ is obtained by compositing with an infinite-depth sky model:
\begin{equation}
   \mathbf{C} = \mathbf{C}_{\text{voxels}} + \text{T}_{\text{final}} \cdot \mathbf{C}_{\text{sky}}
\end{equation}
where $\mathbf{C}_{\text{sky}}$ is the view-dependent color of the infinite sky background. Gradients flow through each voxel segment independently following SVRaster \cite{svr}.

\subsection{Training Details}\label{sec:train_details}
\paragraph{Multi-Modal Supervision Losses.} 
We jointly optimize the multi-octree representation framework including background and dynamic octrees as well as our sky model end-to-end by using the following loss function:

\begin{equation}
\mathcal{L}
= \lambda_{\text{color}}\mathcal{L}_{\text{color}} + \lambda_{\text{depth}}\mathcal{L}_{\text{depth}} + \lambda_{\text{normal}}\mathcal{L}_{\text{normal}} + \lambda_{\text{sky}}\mathcal{L}_{\text{sky}},
\end{equation}
where $\mathcal{L}_{\text{color}}$ is the reconstruction loss between rendered and observed images as defined in \cite{svr}. $\mathcal{L}_{\text{depth}}$ is a L1 loss between rendered depth 
and sparse depth generated from LiDAR points projected into the camera. $\mathcal{L}_{\text{normal}}$ maximizes the cosine similarity between rendered voxel normals and pseudo-ground-truth normals obtained with DepthAnythingV2~\cite{depth_anything_v2} and $\mathcal{L}_{\text{sky}}$ identifies sky pixels via zero-depth values and applies a sky prior mask that explicitly enforces maximum global ray transmittance through an MSE loss to eliminate floating voxels within sky region. Please refer to the supplementary material \cref{sec:losses} for details of each loss term.

\paragraph{Training Schedule.} 

Rather than using a fixed iteration budget for all scenes, our method employs an adaptive training schedule that dynamically determines when to stop voxel subdivision. Since driving environments exhibit varying geometric complexity and dynamic content, the optimal training duration differs across scenes. To account for this variability, we periodically monitor reconstruction quality using SSIM. Once improvements plateau, the framework terminates background refinement and freezes the voxel structure, preventing unnecessary subdivision while reducing training time. Additional details are provided in the supplementary material \cref{sec:train_details}. 

%% file: secs/4_experiments.tex
\section{Experiments}
We evaluate our proposed framework,  \method, across different driving scenarios to demonstrate its performance in dynamic scene reconstruction, geometry and downstream applications.

\paragraph{Dataset and Evaluation metrics.} We perform evaluation on the PandaSet dataset~\cite{xiao2021pandaset} which contains 103 driving scenes captured at 1920×1080 resolution together with 360° LiDAR measurements. We follow the literature~\cite{chen2025salf, yang2023unisim, tancik2023neurad} and use the standard 10 sequence evaluation set selecting every 2nd image in the sequence as training frames and the rest as test using the same protocol defined in SaLF~\cite{chen2025salf}.

To evaluate novel view synthesis quality, we report standard perceptual metrics: PSNR, SSIM, and LPIPS (with a VGG backbone). To validate reconstructed geometric accuracy, we report Chamfer Distance (CD), Median error, and F1@0.1 scores computed against the accumulated ground-truth LiDAR point clouds following the StreetSurf protocol~\cite{guo2023streetsurf}. Runtime efficiency is verified via rendering frame rates (FPS)
and mean per-scene training time across all sequences.
Since the official SaLF codebase is not publicly available, we are unable to reproduce geometry metrics (CD, F1) for this baseline. We therefore restrict our direct quantitative comparison to the metrics reported in their original manuscript for this baseline, and acknowledge this as a limitation of our evaluation.

\paragraph{Baselines and Implementation Details.} We compare our approach against several state-of-the-art dynamic urban reconstruction methods including radiance fields based UniSim~\cite{yang2023unisim} and NeuRAD~\cite{tancik2023neurad}, 3DGS-based OmniRe\cite{chen2025omnire} and Street Gaussians~\cite{yan2024street} and recent voxel-based SaLF~\cite{chen2025salf} (including both Base and Large variants). For OmniRe and Street Gaussians implementations we use the original DriveStudio codebase\footnote{\url{https://github.com/ziyc/drivestudio}}. We benchmark our method with the original training schedule defined in \cref{sec:train_details} and additionally report our results with fixed number of iterations set to 20K to evaluate performance gains resulting from larger training times, we refer to the later as \method@20K.

\subsection{Comparison with State-of-the-Art Methods}

\input{tabs/pandaset_dynamic_salf}

Evaluation data in \cref{tab:pandaset_dynamic_salf} demonstrate that our framework better reconstructs the geometry of the scene, outperforming all baselines in all geometric metric.
Specifically, \method~achieves an $L_1$ error of 0.044 m, nearly halving that of the closest competitor NeuRAD~\cite{tancik2023neurad}, while improving the $F1@0.1$ score by over 10 points. 
Regarding efficiency, our model trains faster than most competing methods; although SaLF (base) trains slightly faster, our approach yields significantly higher geometric precision and structural fidelity (SSIM and LPIPS).
\method @ 20k demonstrates that our training speed is not merely a result of cutting training short. Under its extended configuration, the framework reaches its highest performance, yielding the best overall Chamfer distance (0.235 m) and competitive perceptual scores (0.787 SSIM and 0.255 LPIPS).
Compared to explicit point-based approaches (OmniRe, StreetGS), \method~ maintains competitive rendering quality while delivering superior geometric accuracy and efficiency; we attribute this advantage to the rigid spatial constraints of SVRaster, which limit unconstrained voxel placement~\cite{svr}. Finally, while SaLF (large) marginally leads in PSNR, \method @ 20k retains superior geometric consistency and structural fidelity.

\cref{fig:qualitative_results} shows the qualitative results of our method and baselines on the same scenes reported by SaLF~\cite{chen2025salf}. In most scenarios StreetGS~\cite{yan2024street} achieves higher visual quality for individual foreground objects compared to the other baselines, reconstructing them with sharp details. However, this method frequently introduces severe floating artifacts and geometric degradation in complex regions. As illustrated in the third row of \cref{fig:qualitative_results}, the truck rendered by StreetGS and OmniRe displays a  noisy appearance. In contrast, both SaLF~\cite{chen2025salf} and our method handle these difficult areas significantly better, preserving a solid and continuous geometry for 
the vehicle without introducing floating noise.

\input{figs_tex/qualitative_main}

\paragraph{Geometry Extrapolation Quality.}

\input{tabs/fid_shift_lane}

To evaluate geometric generalization beyond the training trajectory, we render novel views from an ego-trajectory laterally shifted by $\pm2$ meters from the original camera path. In the absence of ground-truth imagery for this shifted-lane scenario, we report the Fréchet Inception Distance (FID) to evaluate the perceptual realism of the extrapolated views. As shown in ~\cref{tab:geometry_extrapolation}, our model optimized for 20k iterations achieves a lower FID score (${53.5}$) than StreetGS ($72.6$) and OmniRe ($71.6$). This improvement indicates that our explicit voxel layout maintains more reliable geometric constraints than unconstrained point-splatting methods during viewpoint extrapolation. We note that the baseline results for UniSim and NeuRAD are cited directly from the original NeuRAD manuscript \cite{tancik2023neurad}; while they utilize identical training scenes, they differ slightly in their train test splits, yet our framework still demonstrates a clear performance advantage.

\subsection{Ablation Study}

\paragraph{Background Grid Initialization.}
To cleanly isolate the geometric impact of our LiDAR-guided initialization pipeline on the static background representation ($\mathcal{O}_{bg}$) we use five background-dominant PandaSet sequences containing sparse dynamic actors (for implementation details, see 
Supplementary Material 
). To ensure a fair comparison, all variants are trained for a fixed window of 20k iterations. The evaluation compares our method against two distinct variants: 1) \textbf{w/o LiDAR subdivision} which initializes the model from a pruned, coarse uniform grid without iterative LiDAR-guided subdivision. And 2) \textbf{Binary density init.} which replaces the soft normal distribution prior with a binary assignment ($\mathbf{v}_{\mathrm{close}}$ or $\mathbf{v}_{\mathrm{far}}$) based on a hard distance threshold.

\input{tabs/ablation_init}

As shown in \cref{tab:ablation_init}, removing LiDAR-guided subdivision results in the largest performance drop across all metrics ($\Delta \text{PSNR} = 0.21$~dB, $\Delta \text{LPIPS} = 0.017$, $\Delta \text{CD} = 0.012$~m). On the other hand, binary density initializations performs comparably to our method with slight decrease on perceptual metrics and slight gain in geometry. We attribute this comparable performance to the subdivision attribute of our octree structure during training which enables to adapt the voxel sizes and their densities without the need of soft initialization. Nevertheless, we argue that a soft density initialization is generally better for scenes with very sparse LiDAR data or coarse initial grid.

\paragraph{Supervision Losses.}
\cref{tab:ablation_losses} ablates the contribution of our LiDAR and normal supervision losses as well as our sky-model supervision by individually removing each of such components and train each variant for 20K iterations. 
Results show that the most significant contribution in geometric accuracy comes from $\mathcal{L}_{\text{depth}}$ contribution. In addition, $\mathcal{L}_{\text{normal}}$  improves the F1@0.1 score from $49.0$ \% to $66.3$ \%. LiDAR provides accurate sparse supervision while dense normals improves geometry on areas with no LiDAR or with lack of texture (see \cref{fig:ablation_depth_images}). Both supervision signals complement each other improving geometry but penalizing perceptual metrics. We argue this trade-off results from the spatial misalignment between the LiDAR and camera sensor calibration. Although our sky model brings small contribution to quantitative metrics, it avoids allocation of large number of voxels for the sky, saving computational resources. Furthermore, the absence of groundtruth geometry at sky region hinders correct quantitative assessment of this component.

\input{tabs/ablation_losses}

\input{figs_tex/depth_ablation}

\subsection{Applications}
Our 3D reconstruction framework naturally enables three downstream tasks for autonomous driving simulation, without requiring extra training pipelines: traffic scene editing, 3D semantic segmentation and open vocabulary scene understanding. While these initial results leave room for future improvement, they show that our method is immediately useful for simulation task.

\begin{figure*}[t] 
    \centering
    \def\fgsize{0.24}
    \scriptsize
    \setlength{\tabcolsep}{0.0020\linewidth}
    \renewcommand{\arraystretch}{0.00}
    \begin{tabular}{c @{\hspace{7pt}} ccc}%
         Render  &  Removal & Pose Change & Replacement \\ 
         
        \includegraphics[clip=false, trim={0 0 0 0},width=\fgsize\columnwidth]{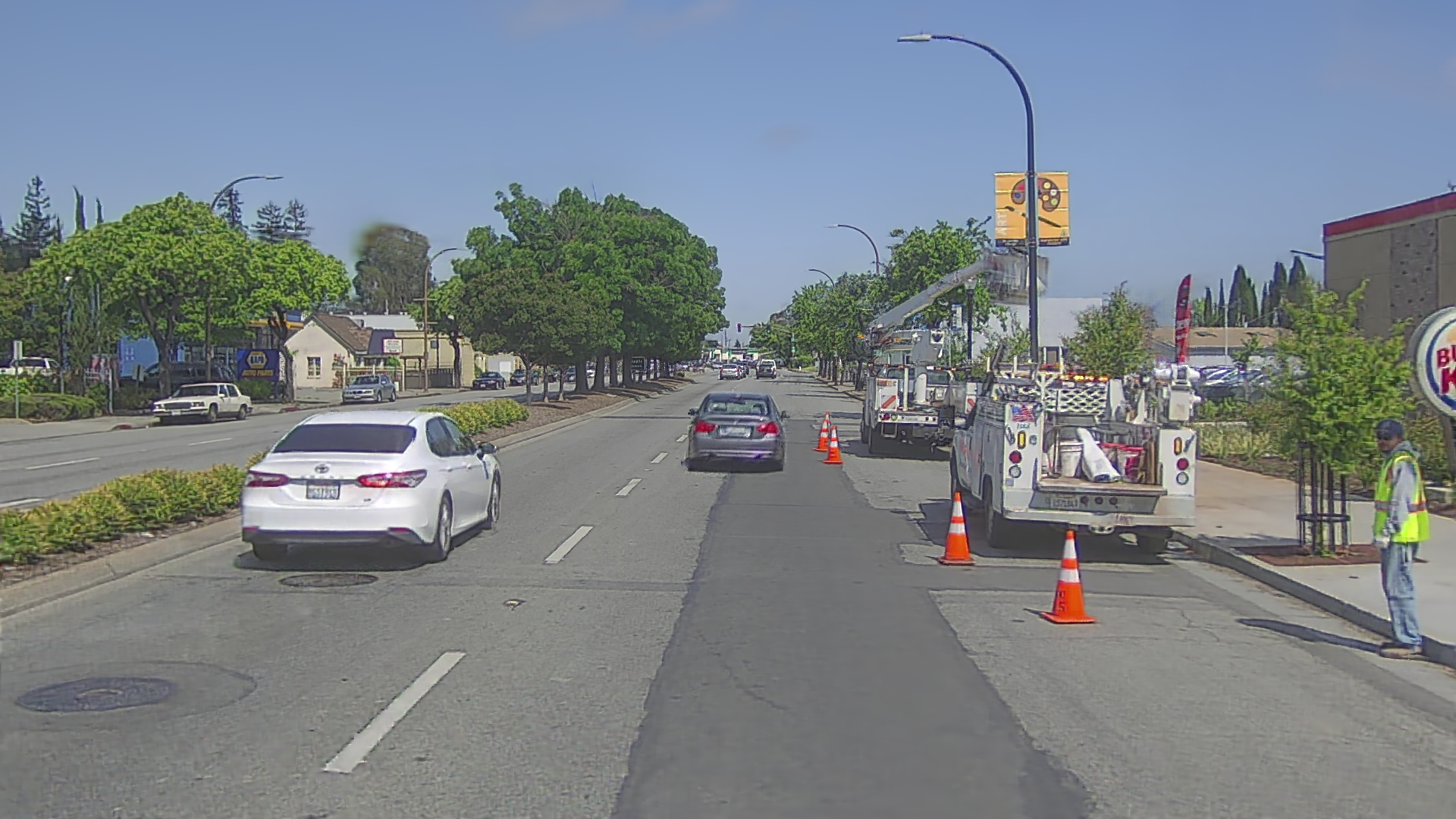} & 
        \includegraphics[clip=false, trim={0 0 0 0},,width=\fgsize\columnwidth]{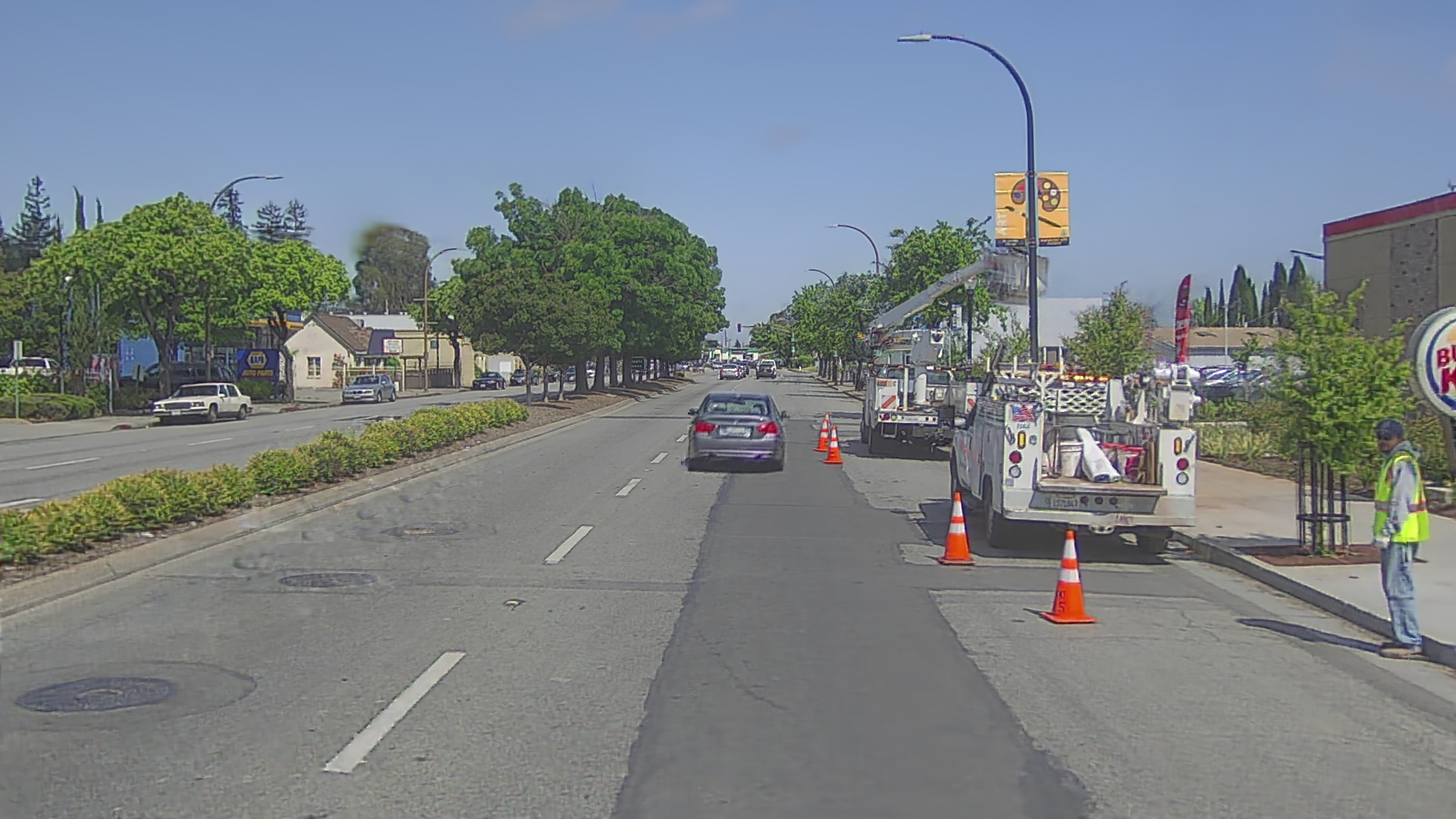} & 
        \includegraphics[clip=false, trim={0 0 0 0},,width=\fgsize\columnwidth]{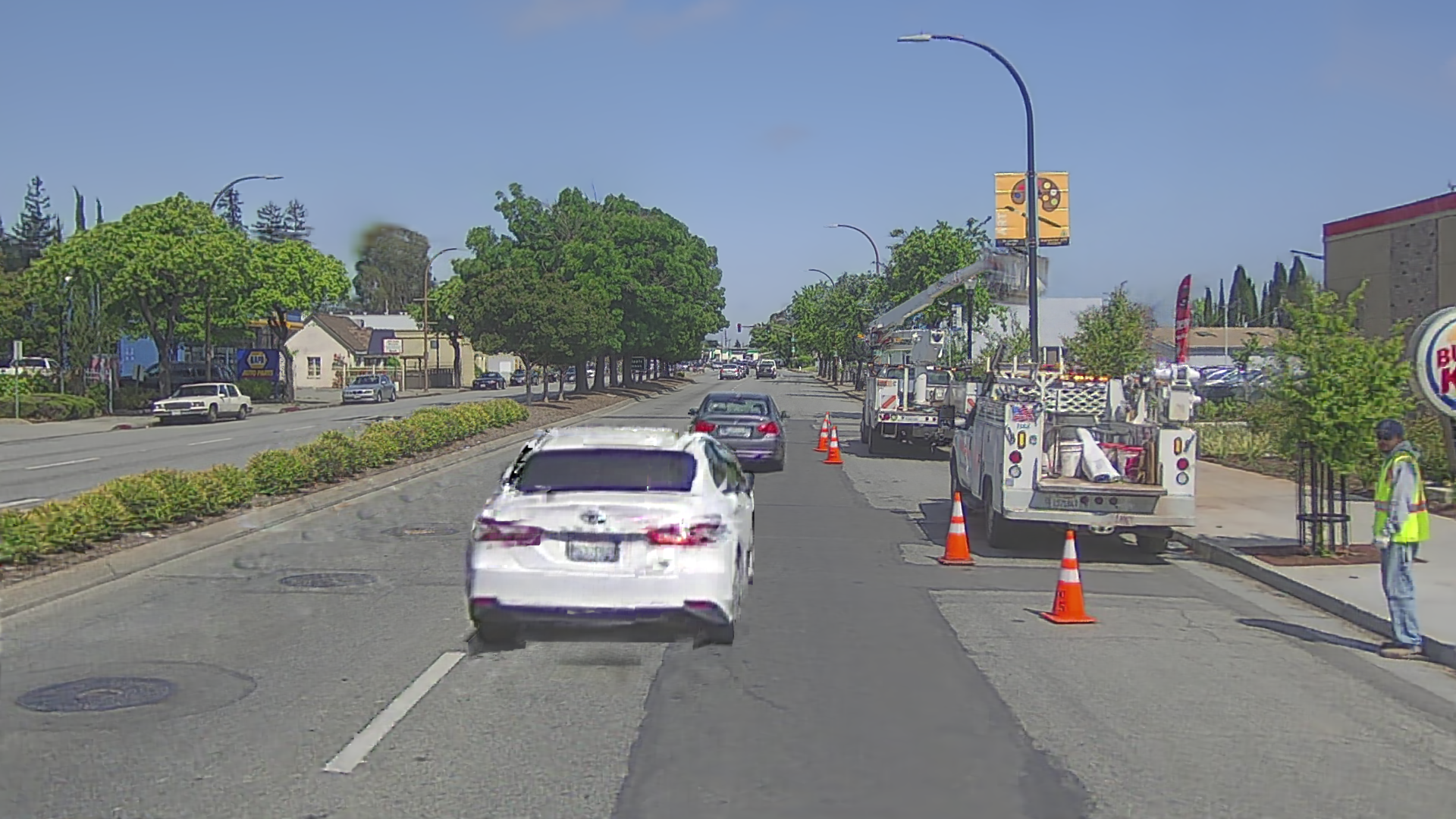} & 
        \includegraphics[clip=false, trim={0 0 0 0},,width=\fgsize\columnwidth]{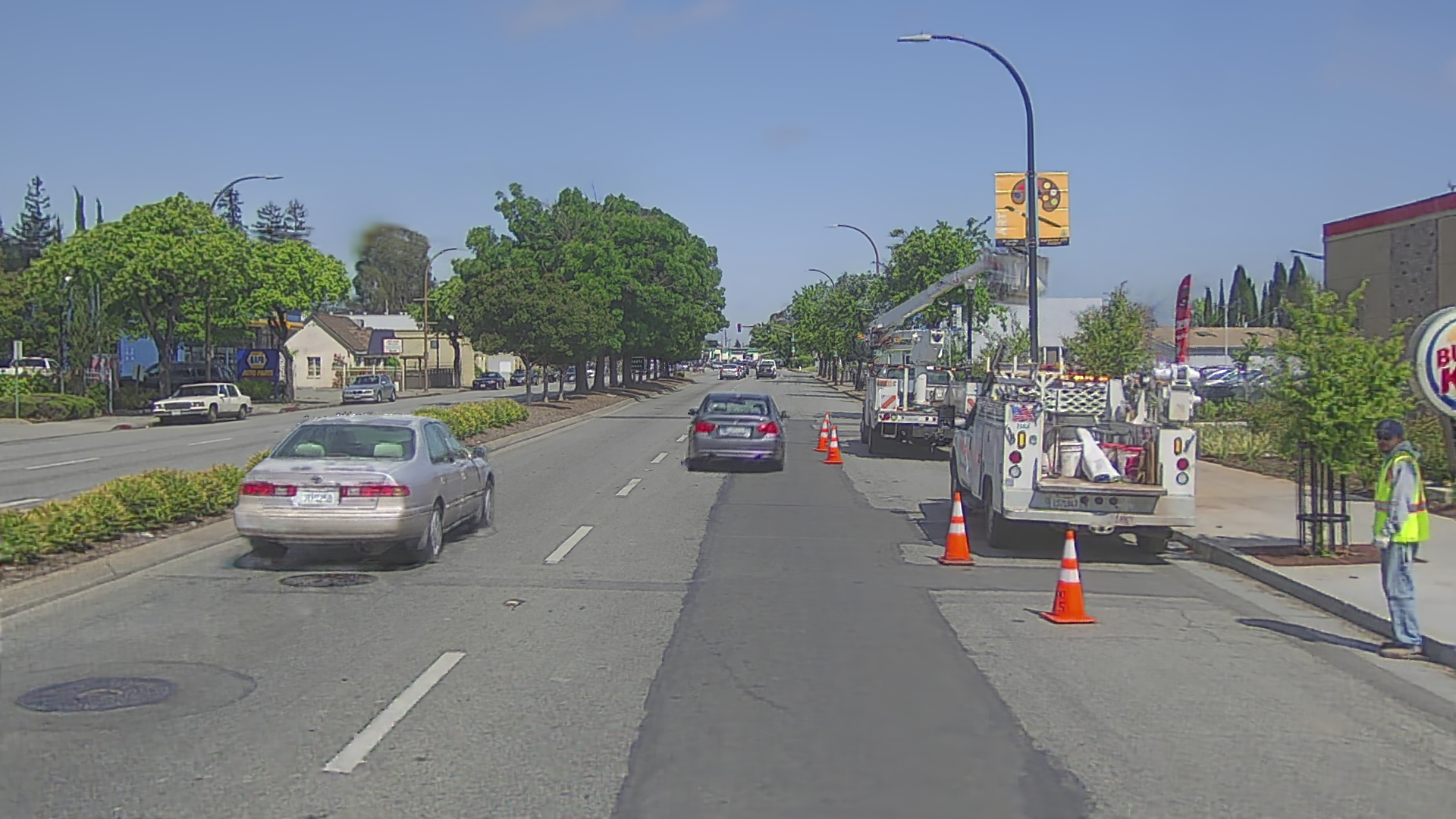} \\

    \end{tabular}
    \caption{\textbf{Post-Training Scene Editing with Explicit Voxel Assets.} Because each dynamic actor is stored in its own voxel grid, \method~supports direct scene manipulation after training. The examples show object removal, pose modification, and asset replacement without retraining while preserving consistent compositing with the surrounding scene.}

    \label{fig:edition_images}

\end{figure*}

\paragraph{Dynamic Scene Editing Operations.}
Our explicit voxel representation makes it easy to edit the scene after training. Since every moving object is saved in its own independent grid, we can simply remove an object to change the scene layout. In addition, by simply modifying the object transformations $\{T_t^i\}$, we can move the object along a new trajectory. We can also swap one object with another by replacing its voxel data in the scene graph, allowing us to replace a target vehicle with a different model while maintaining accurate rendering boundaries. In \cref{fig:edition_images}, we can visualize examples of these operations.

\paragraph{3D Semantic Segmentation.}
Building on top of SVRaster \cite{svr} allows us to exploit its structured and non overlapping sparse voxel representation to map 2D pixel coordinates and explicit 3D primitives. As demonstrated by recent studies~\cite{wang2026lesv}, feature fusion in a sparse voxel architecture outperforms unstructured 3DGS alternatives, as it eliminates the ambiguity and prevents semantic bleeding across object boundaries.

\begin{figure*}[!t]
    \centering
    \def\fgsize{0.28}
    \def\fgheight{1.8cm}
    \scriptsize
    \setlength{\tabcolsep}{0.0020\linewidth}
    \renewcommand{\arraystretch}{0.00}
    \begin{tabular}{c @{\hspace{7pt}} cc}%
         GT  &  Segformer & \method \\ 
         
        \includegraphics[clip=false, trim={0 0 0 0},
        width=\fgsize\columnwidth
        ]{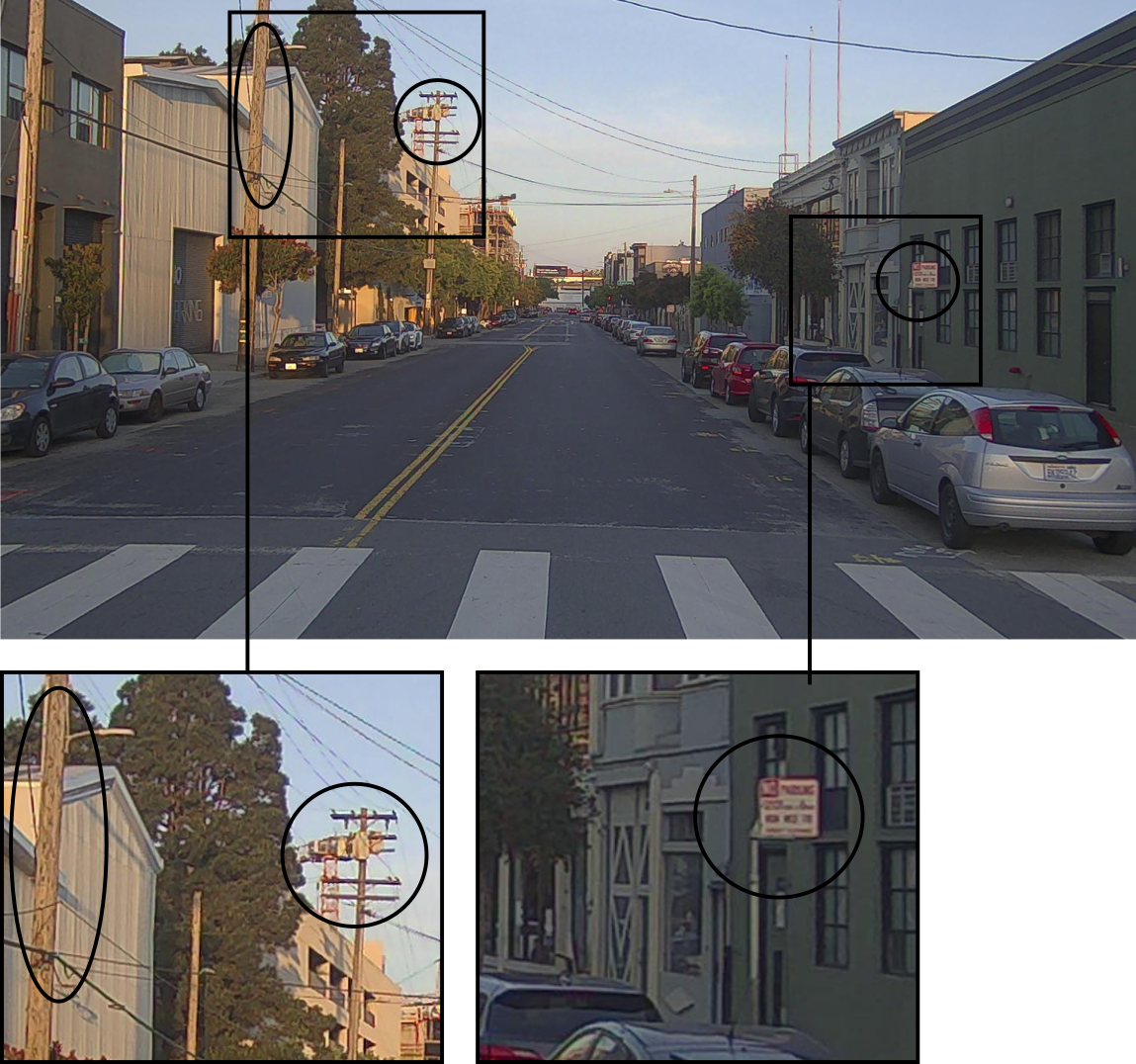} & 
        \includegraphics[clip=false, trim={0 0 0 0},
        width=\fgsize\columnwidth
        ]{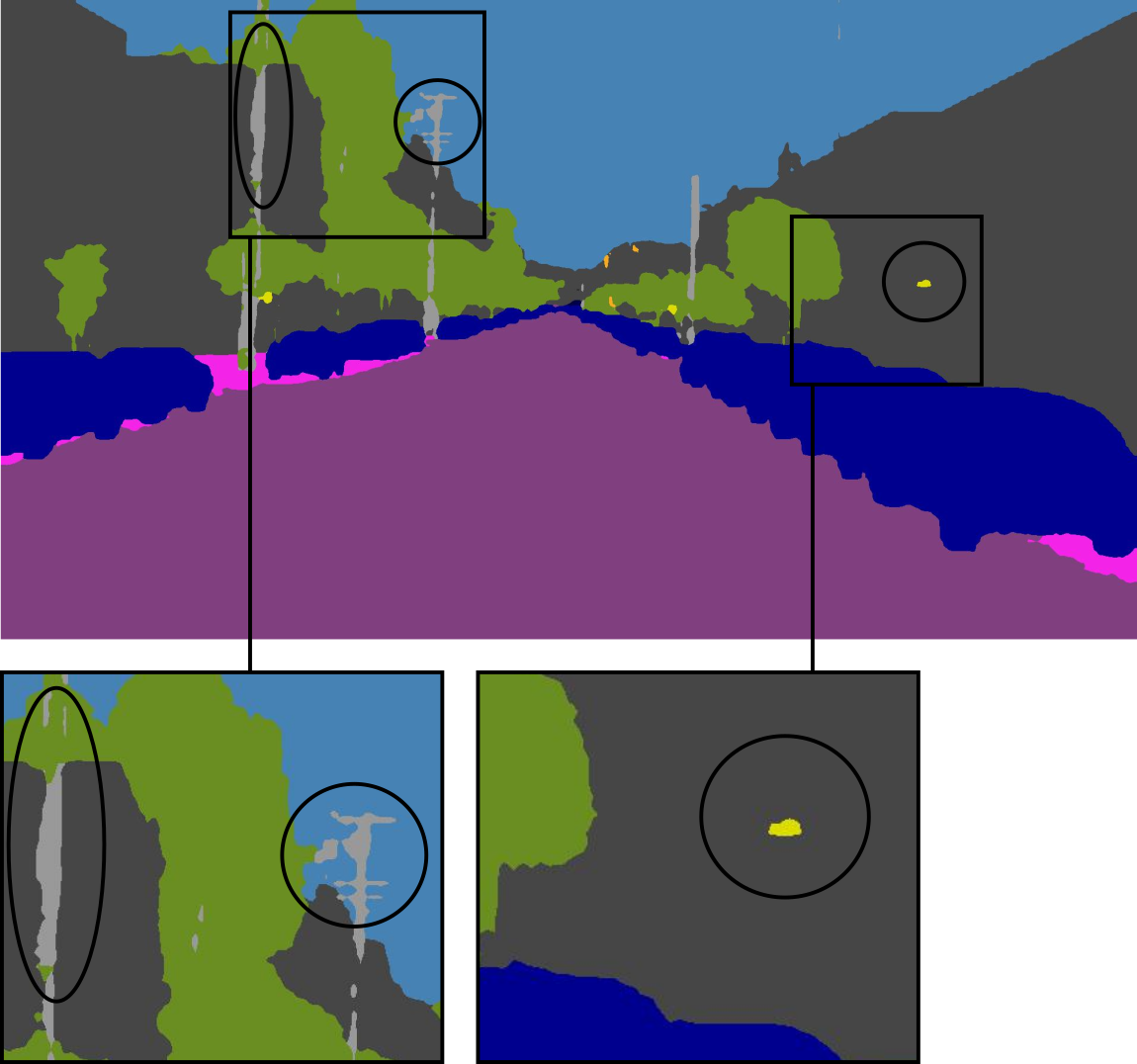} & 
        \includegraphics[clip=false, trim={0 0 0 0},
        width=\fgsize\columnwidth
        ]{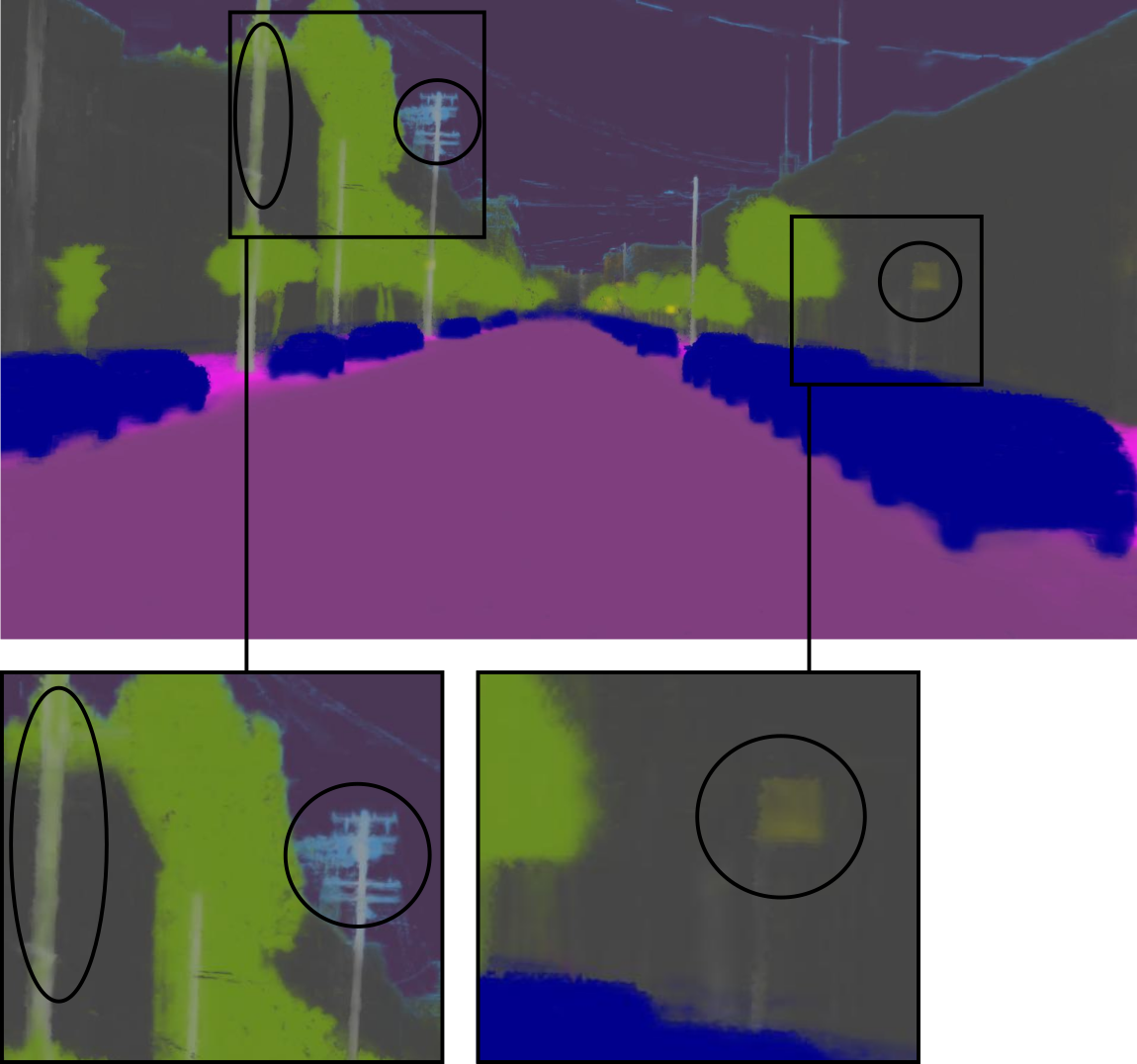} \\      
       
    \end{tabular}
    \caption{\textbf{Structured 3D Semantic Fusion Improves Scene Labels.} We project per-view SegFormer predictions into the sparse voxel volume and fuse them across views using the reconstructed geometry. Compared with the raw monocular segmentation, the fused result is more spatially consistent and better aligned with the ground-truth layout, especially along roads, sidewalks, vegetation, and thin vertical structures.}

    \label{fig:segmentation_fusion}
\end{figure*}

We use the core rasterization engine by substituting the standard RGB color rendering step with a high-dimensional feature accumulation pass during the volume rendering pipeline~\cite{svr}. This enables our model to create a robust multi-view feature fusion by projecting dense 2D semantic mask, obtained from SegFormer~\cite{segformer}, into the 3D grid space. 
As shown in ~\cref{fig:segmentation_fusion}, the fused 3D segmentation map successfully corrects errors from the raw monocular inputs. 

\begin{figure*}[!t] 
    \centering
    \def\fgsize{0.24}
    \def\fgheight{2.5cm}
    \scriptsize
    \setlength{\tabcolsep}{0.0020\linewidth}
    \renewcommand{\arraystretch}{0.00}
    \begin{tabular}{c @{\hspace{7pt}} cc}%
         Render  &  "Cross Walk" & "Traffic Light" \\ 
         
        \includegraphics[clip=false, trim={0 0 0 0},
        height=\fgheight
        ]{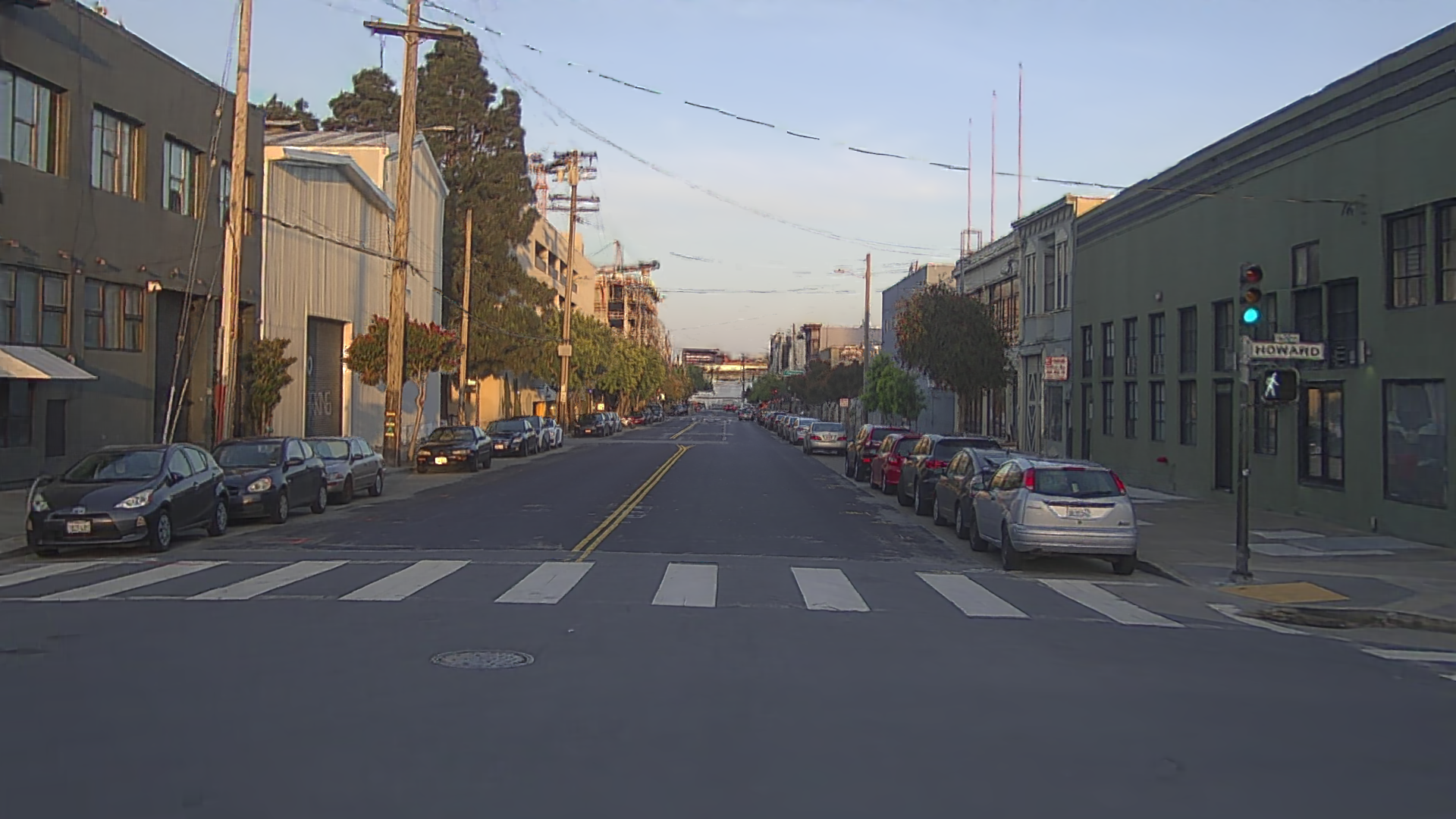} & 
        \includegraphics[clip=false, trim={0 0 0 0},
        height=\fgheight
        ]{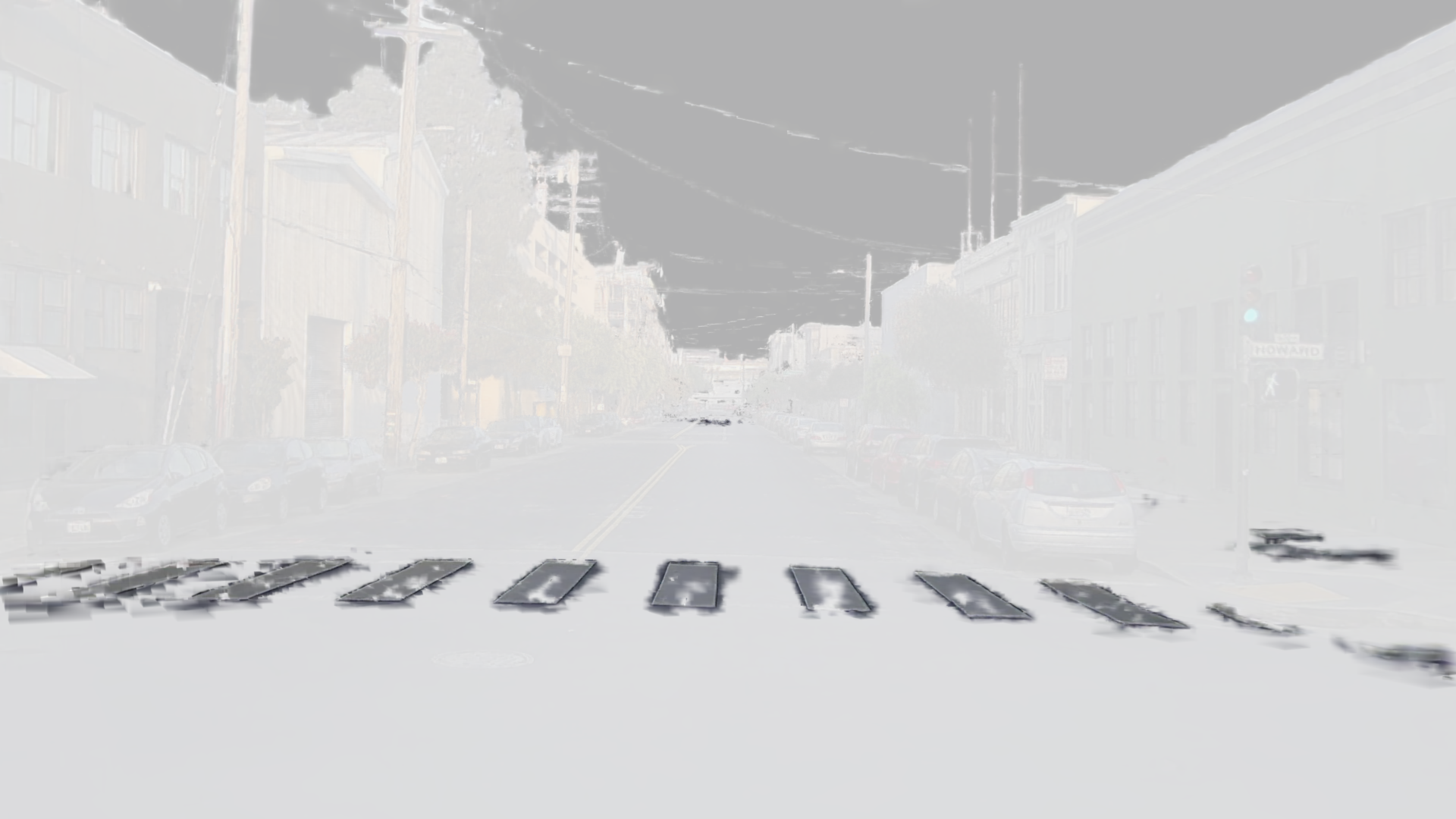} & 
        \includegraphics[clip=false, trim={0 0 0 0},
        height=\fgheight
        ]{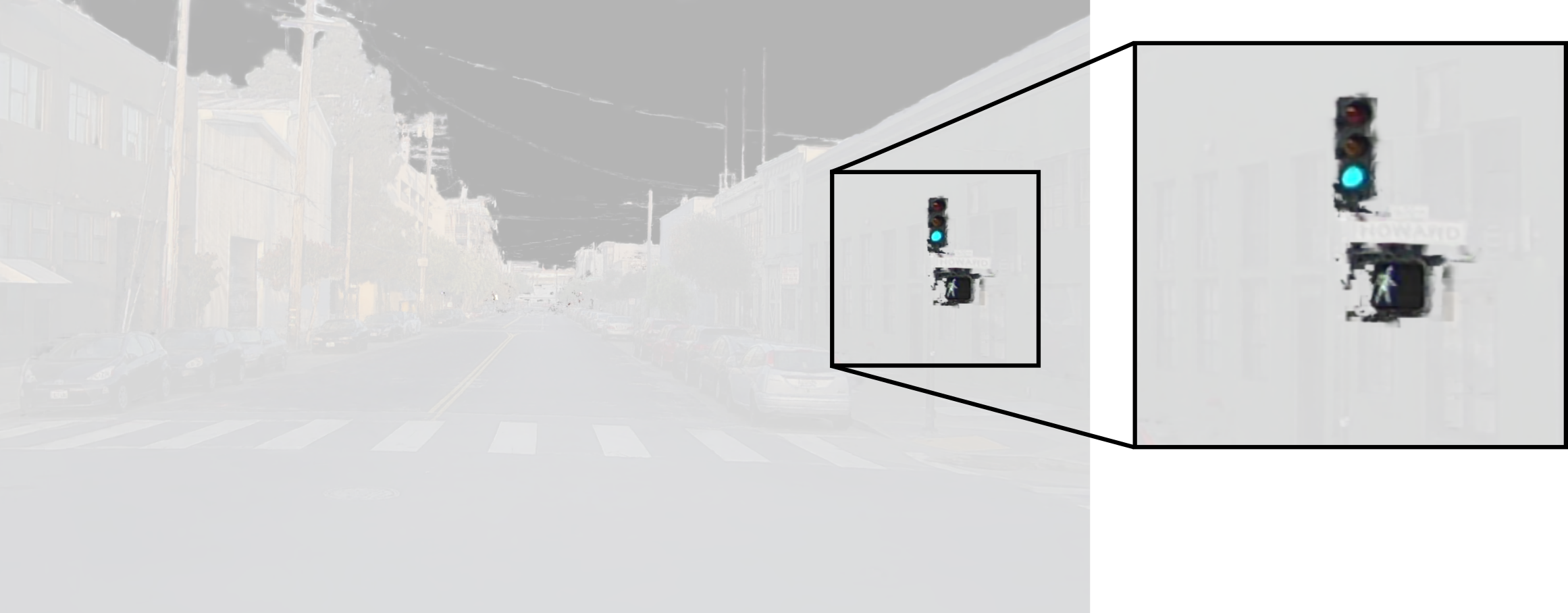} \\      
       
    \end{tabular}
    \caption{\textbf{Open-Vocabulary Scene Queries in 3D.} By fusing language-aligned features into the reconstructed voxel volume, \method{} supports text-driven localization at inference time. The queries \textit{"Cross Walk"} and \textit{"Traffic Light"} activate the corresponding scene regions with accurate spatial grounding from the same reconstruction.}

    \label{fig:open_vocabulary_query}
\end{figure*}

\paragraph{Open Vocabulary Scene Understanding.}
Furthermore, the feature volume fuser can be expanded to open-vocabulary language fields~\cite{wang2026lesv}. We obtained a fuse 3D feature from the language-aligned features from the AM-RADIO foundation model \cite{ranzinger2024radio}. Our architecture allows the user to perform unconstrained text-based queries at inference time without requiring network training by computing the pixel-wise cosine similarity between a CLIP-encoded text embedding vector and our fused 3D feature. As shown in \cref{fig:open_vocabulary_query}, we obtain highly accurate 3D segmentations for queries such as ``Cross Walk'' or ``Traffic Light''.

%% file: tabs/pandaset_dynamic_salf.tex
\begin{table*}[t]
\centering

\label{tab:pandaset_dynamic_salf}

\resizebox{0.91\textwidth}{!}{
\begin{tabular}{lccccccccc}
\cmidrule(l{0pt}r{0pt}){2-9}
& \multicolumn{3}{c}{Perceptual} 
& \multicolumn{3}{c}{Geometry} 
& \multicolumn{2}{c}{Efficiency} \\
\cmidrule(lr){2-4} \cmidrule(lr){5-7} \cmidrule(lr){8-9}
Method 
& PSNR$\uparrow$ 
& SSIM$\uparrow$ 
& LPIPS$\downarrow$
& L1 (m)$\downarrow$
& CD (m)$\downarrow$
& F1@0.1 (\%)$\uparrow$
& FPS$\uparrow$ 
& Train Time (h)$\downarrow$ \\
\midrule

UniSim~\cite{yang2023unisim}$^{\ddagger}$ & \underline{25.63} & 0.745 & 0.288 & 0.100 & - & - & 1.3 & 
1.67
\\

NeuRAD~\cite{tancik2023neurad}$^{\ddagger}$ & 26.60 & 0.770 & 0.297 & 0.085 & - & - & 1.7 & 
3.48
\\

OmniRe \cite{chen2025omnire}$^{*}$ & 25.60 & \textbf{0.789} & \underline{0.248} &  0.129 &  0.594 & 54.9 & 44.9 & 
1.94
\\

StreetGS~\cite{yan2024street}$^{*}$ & 25.56 & \textbf{0.789} & \textbf{0.249} & 0.130 & 0.605  & 54.8 & \textbf{57.0} & 
1.80
\\

SaLF (base)~\cite{chen2025salf}$^{\ddagger}$ & 25.48 & 0.744 & 0.373 &  0.142 & - & - & \underline{54.5} & 
\textbf{0.31}
\\

SaLF (large)~\cite{chen2025salf}$^{\ddagger}$ & \textbf{25.78} & 0.762 & 0.344 & 0.111 & -  & - & 34.3 & 
0.48
\\

\midrule

\method~(ours) & 24.99 & 0.772 & 0.299 & \textbf{0.044} &  \underline{0.242}  & \textbf{65.6} & 26.0 & 
\underline{0.41}
\\ 

\method @20k~(ours) & 25.34 & \underline{0.787} & 0.255 & \underline{0.046} & \textbf{0.235}  & \underline{65.3} & 26.3 & 
1.01
\\
\midrule
\end{tabular}

}
\\{\scriptsize $^{\ddagger}$ Results repoted on \cite{chen2025salf} original manuscript. $^{*}$ Obtained using DriveStudio codebase~\cite{chen2025omnire}}

\vspace{0.3cm}

\caption{
\textbf{Dynamic PandaSet Benchmark.} On ten dynamic PandaSet sequences, \method~achieves the best geometric accuracy and competitive perceptual quality.
}

\end{table*}

%% file: figs_tex/qualitative_main.tex
\newcommand{\missingfigure}[2]{
\begin{tikzpicture}
    \draw[gray, line width=0.3pt] (0,0) rectangle (#1,#2);
    \node[gray] at (#1/2,#2/2) {\scriptsize N/A};
\end{tikzpicture}
}

\begin{figure*}[t] 
    \centering
    \def\fgsize{0.20}
    \scriptsize
    \setlength{\tabcolsep}{0.0020\linewidth}
    \renewcommand{\arraystretch}{0.00}
    \begin{tabular}{c @{\hspace{4pt}} cccc}%
         Ground Truth  & OmniRe~\cite{chen2025omnire} & StreetGS~\cite{yan2024street} & SaLF~\cite{chen2025salf}$^{\dag}$ & \method ~(ours) \\ 
         
        \includegraphics[clip=false, trim={0 0 0 0},width=\fgsize\columnwidth]{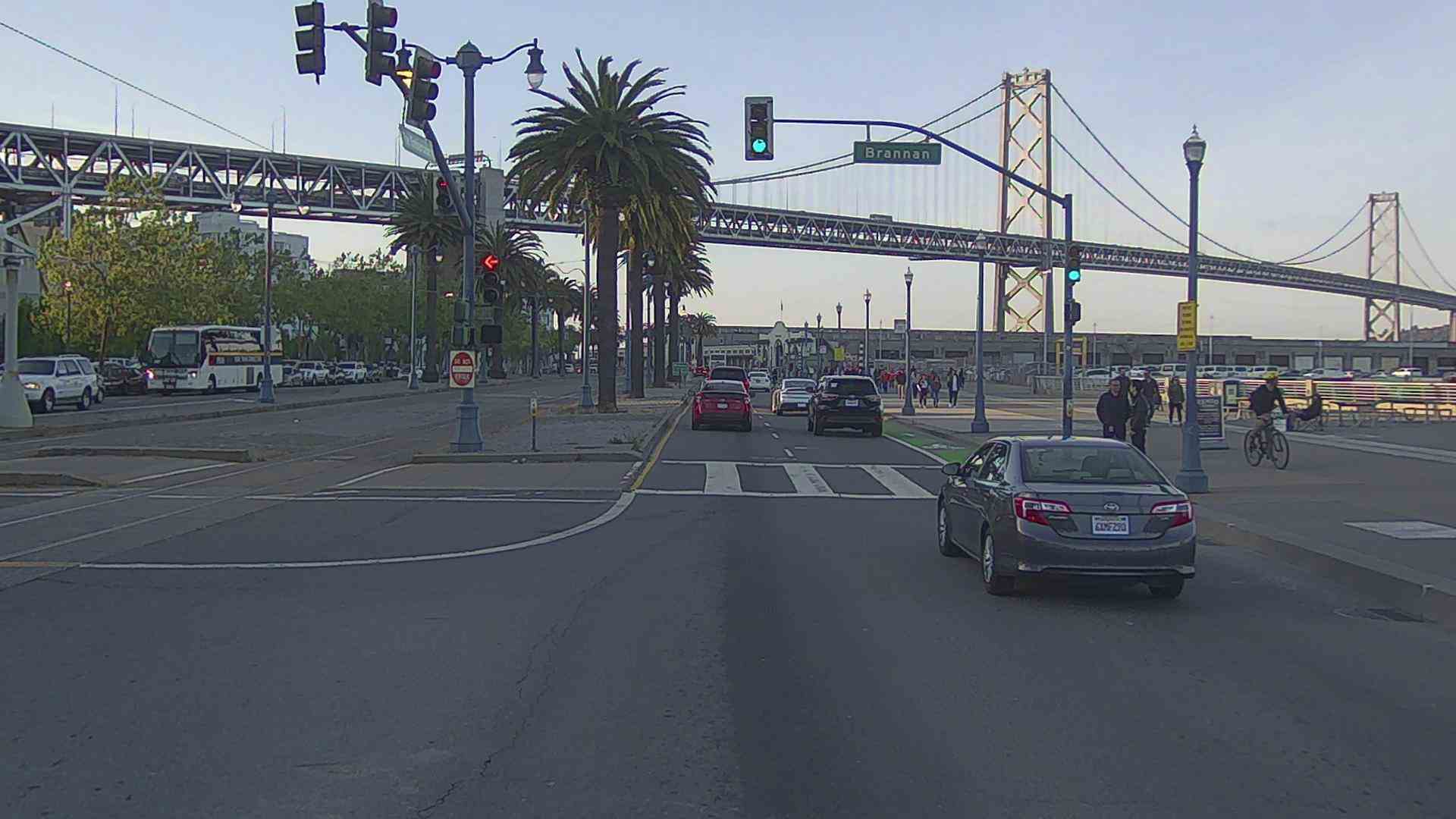} & 
        \includegraphics[clip=false, trim={0 0 0 0},,width=\fgsize\columnwidth]{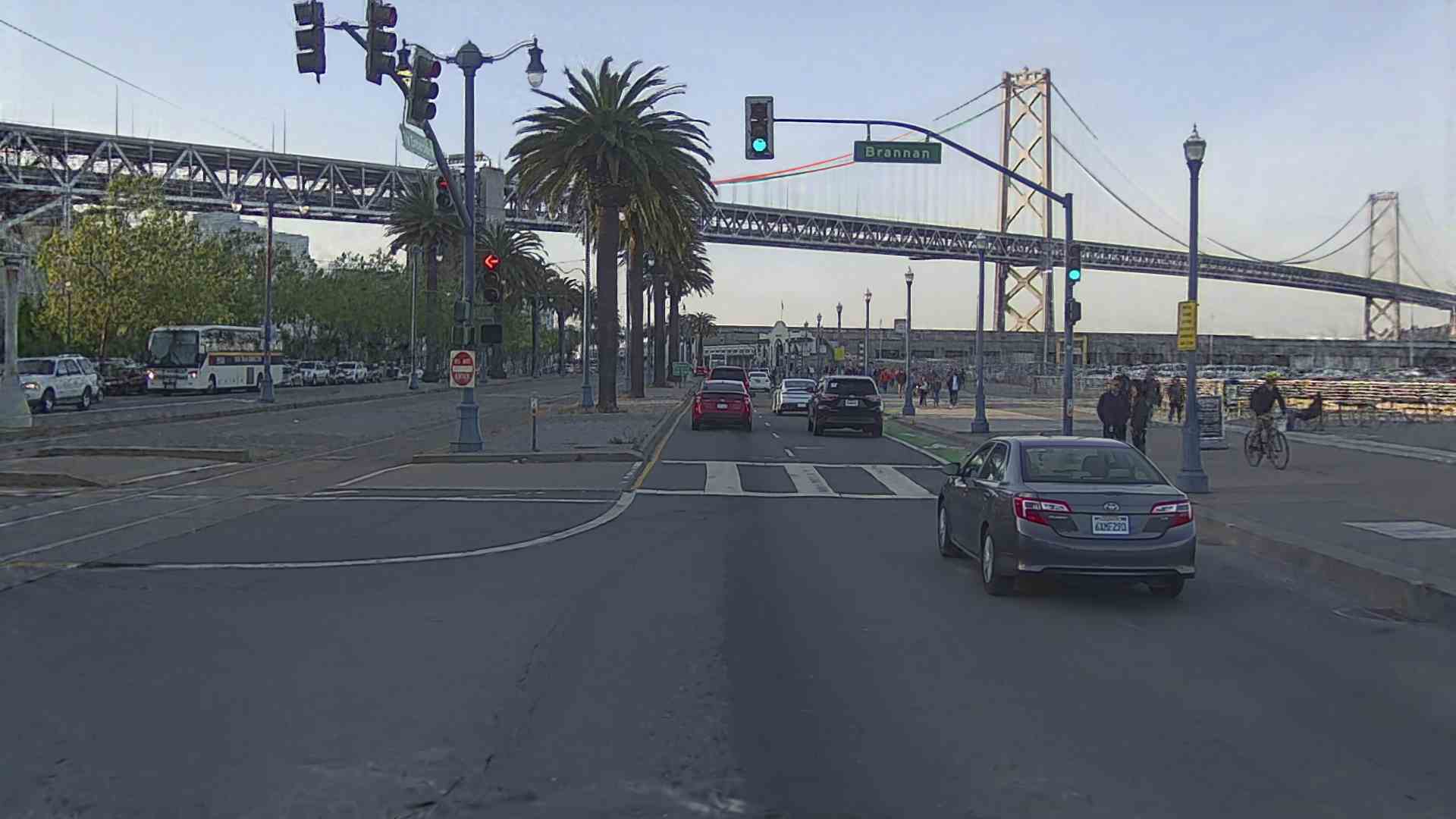} & 
        \includegraphics[clip=false, trim={0 0 0 0},,width=\fgsize\columnwidth]{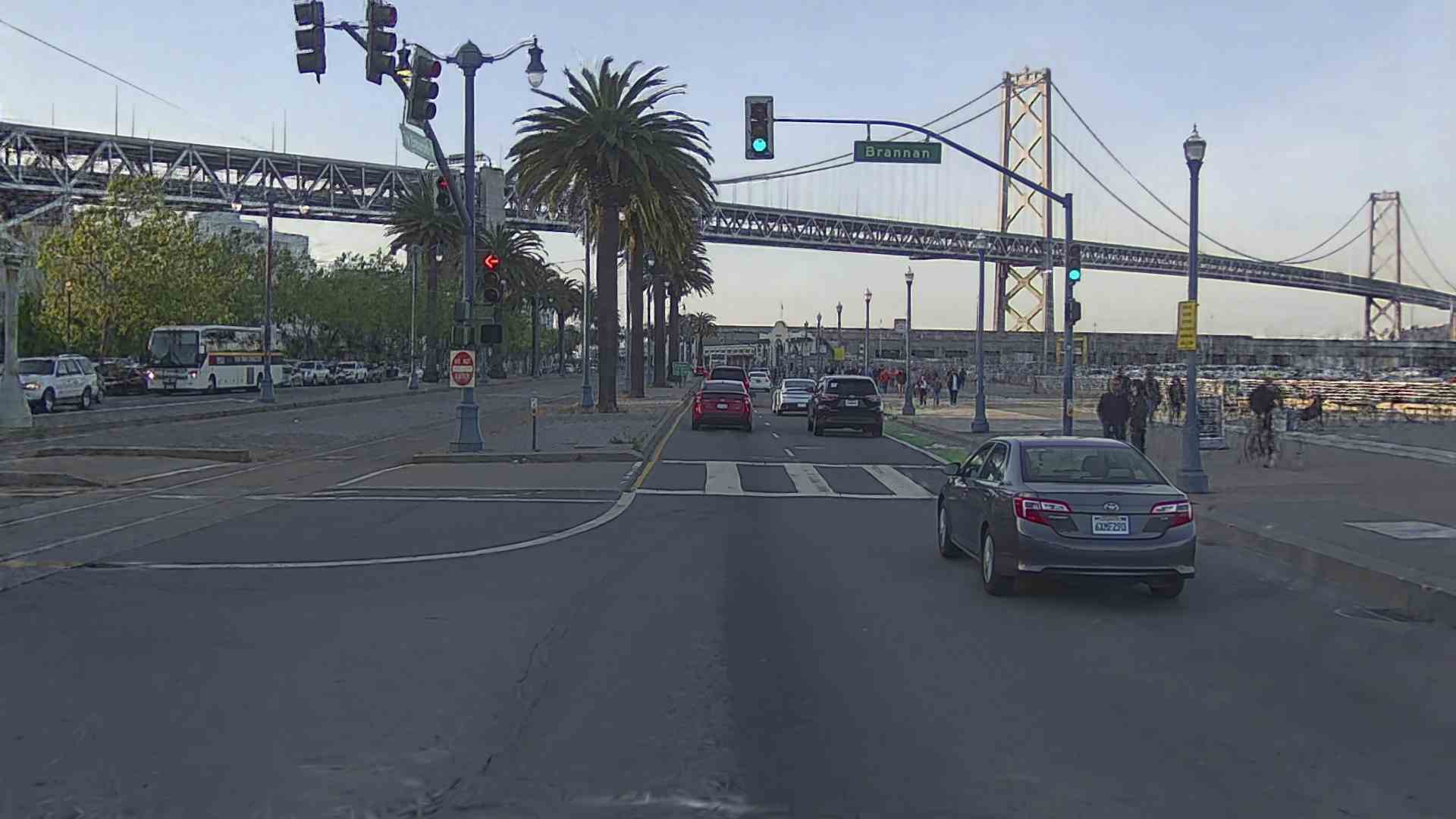} & 
        \includegraphics[clip=false, trim={0 0 0 0},,width=\fgsize\columnwidth]{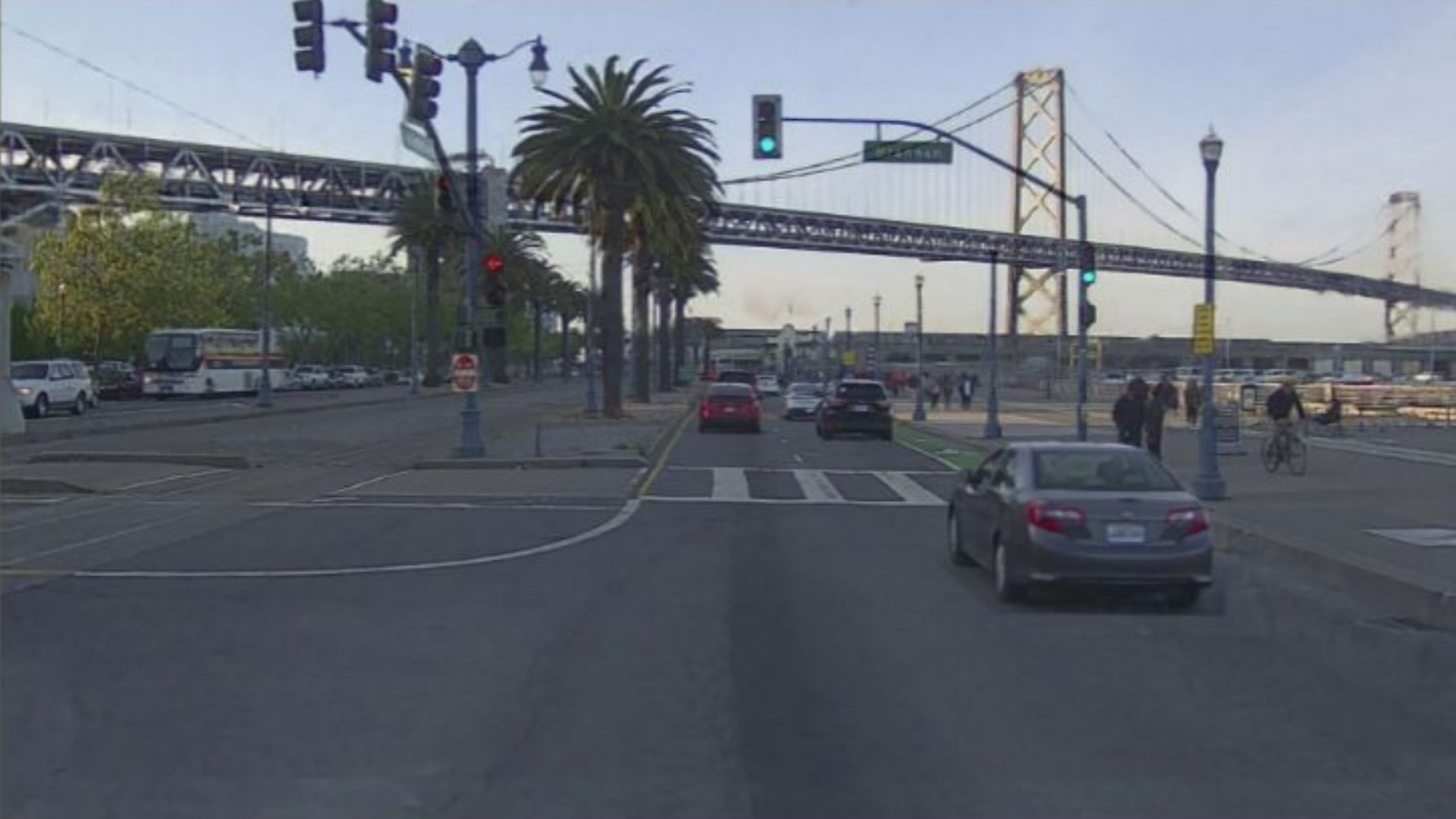} & 
        \includegraphics[clip=false, trim={0 0 0 0},,width=\fgsize\columnwidth]{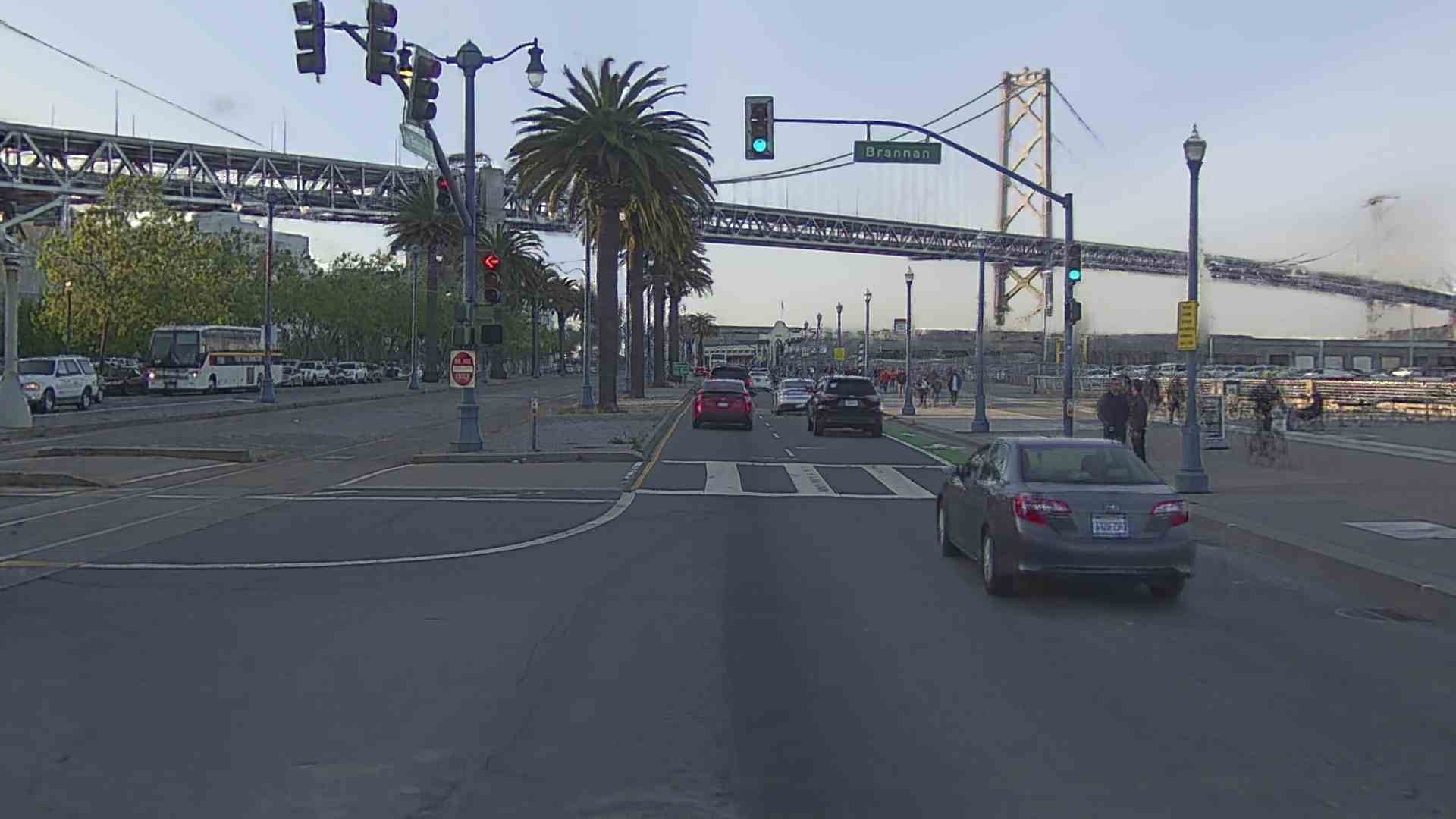} \\      

        \includegraphics[clip=false, trim={0 0 0 0},width=\fgsize\columnwidth]{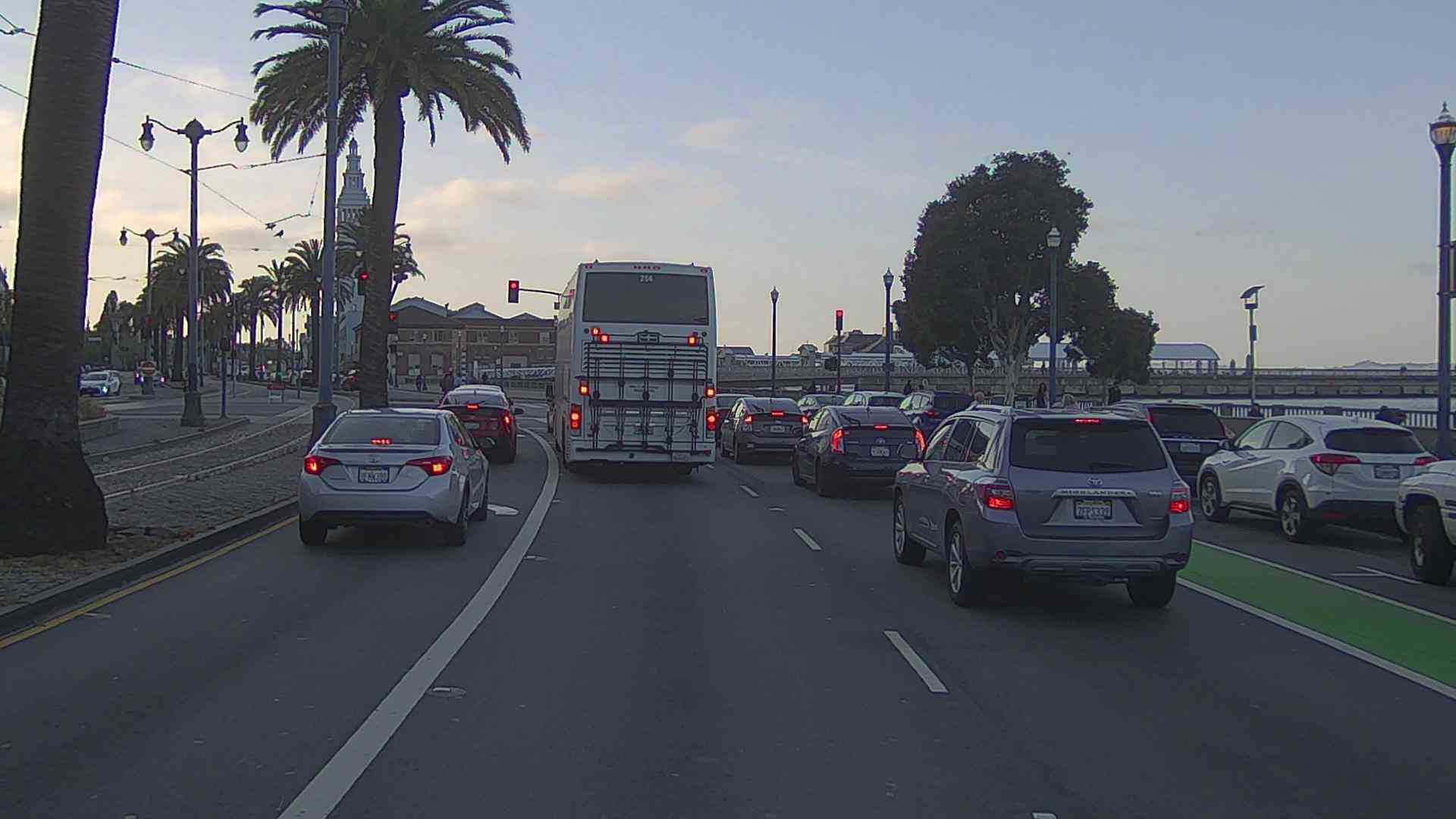} & 
        \includegraphics[clip=false, trim={0 0 0 0},,width=\fgsize\columnwidth]{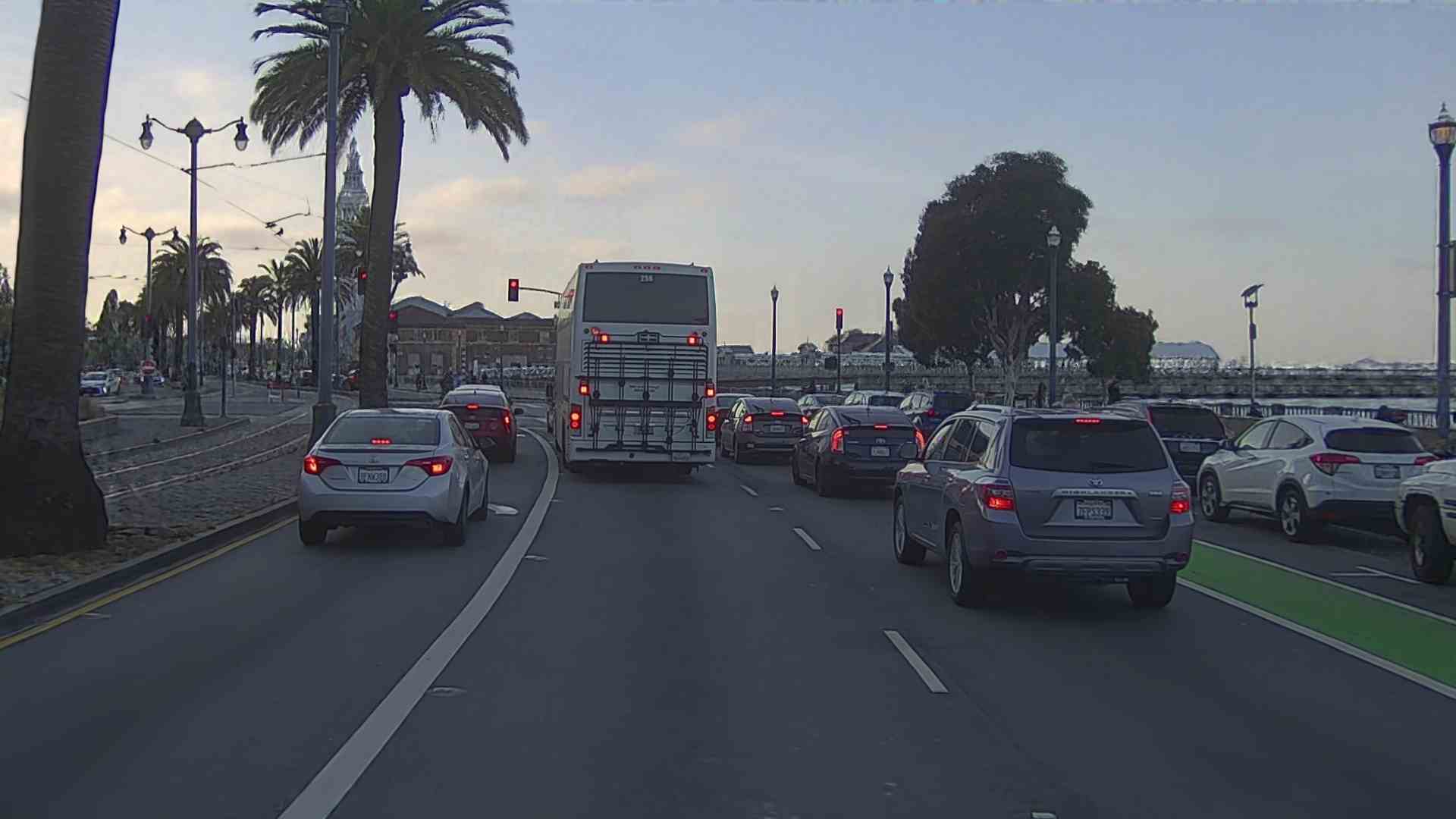} &     
        \includegraphics[clip=false, trim={0 0 0 0},,width=\fgsize\columnwidth]{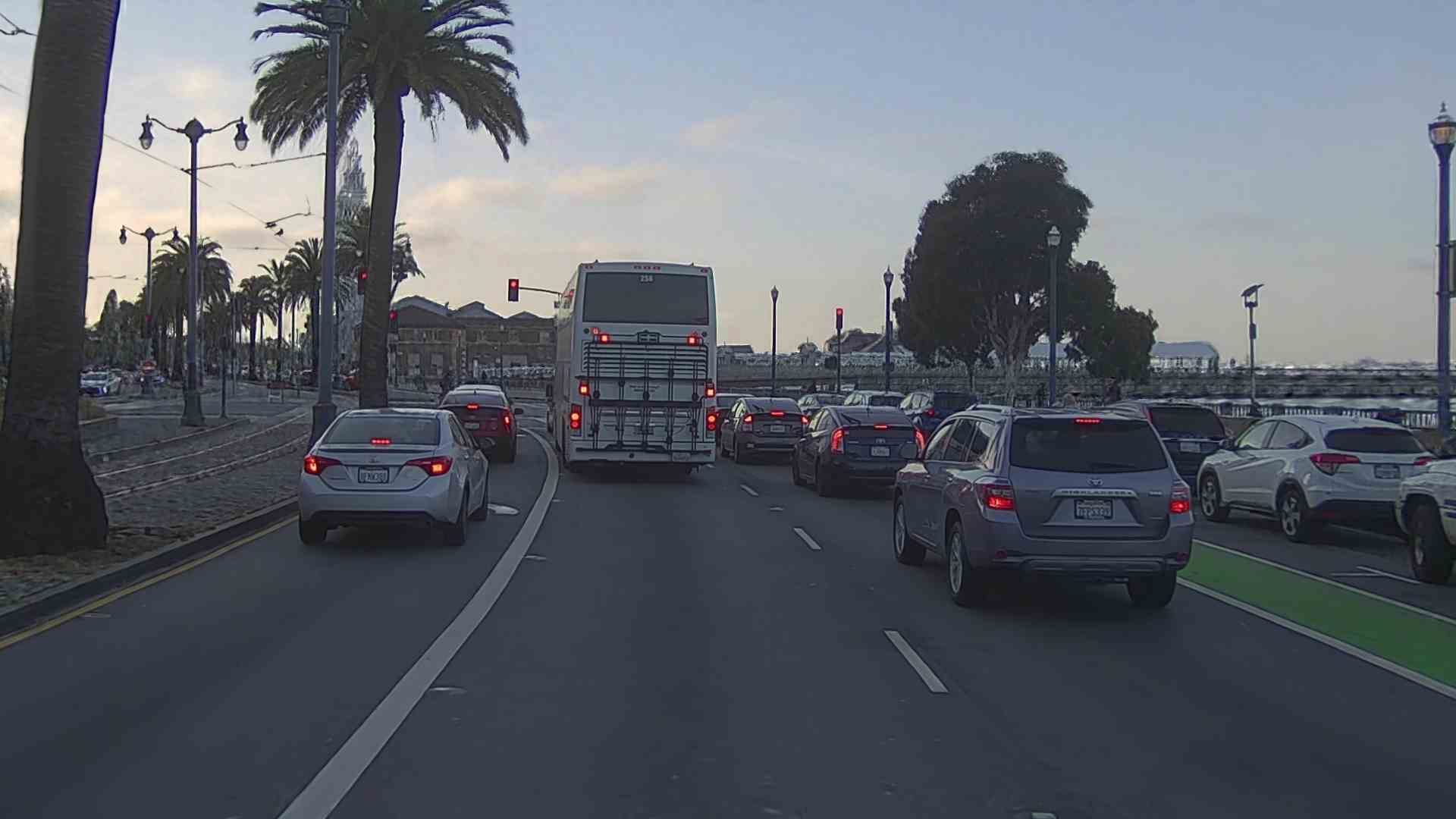} & 
        \includegraphics[clip=false, trim={0 0 0 0},,width=\fgsize\columnwidth]{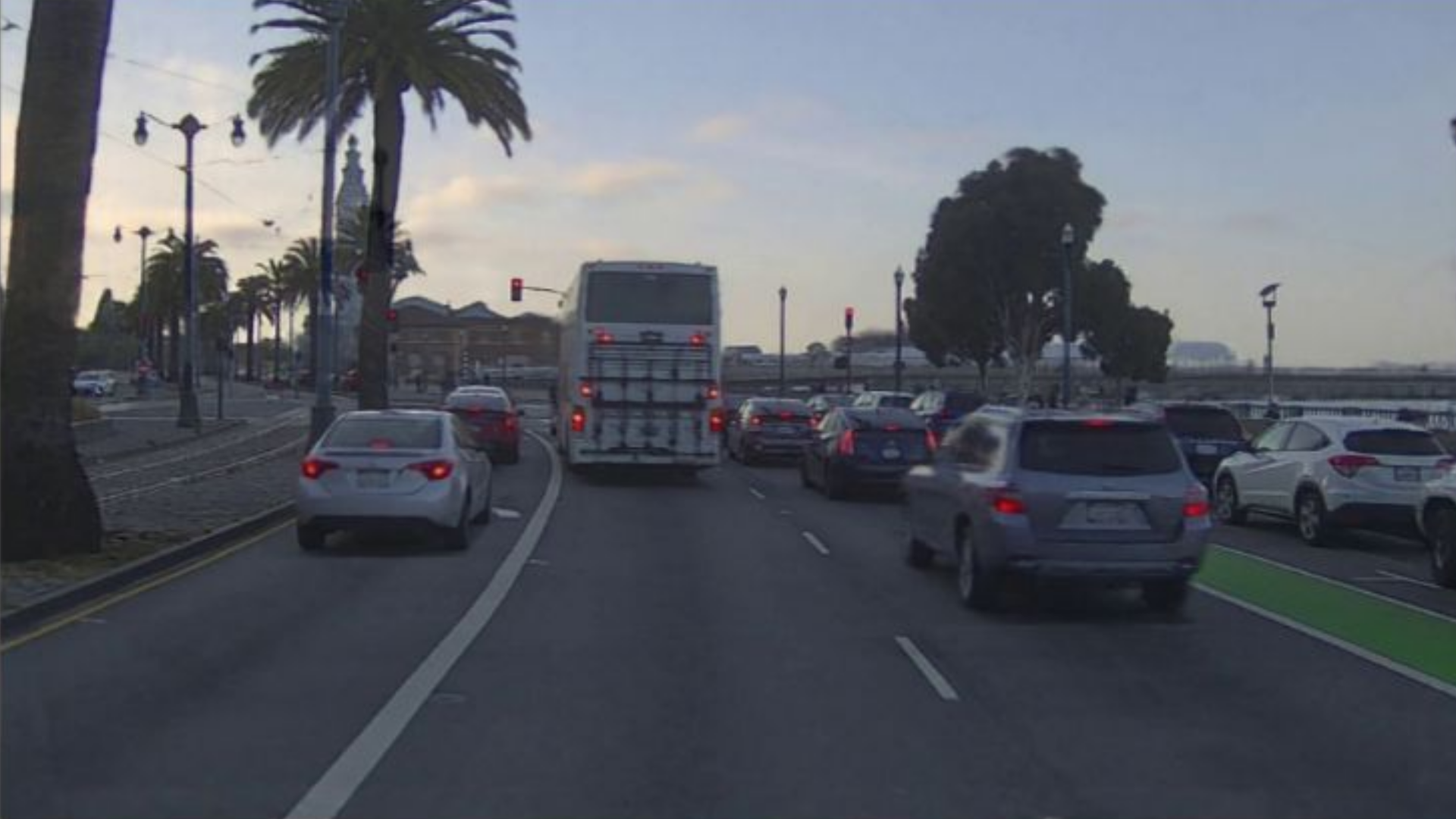} & 
        \includegraphics[clip=false, trim={0 0 0 0},,width=\fgsize\columnwidth]{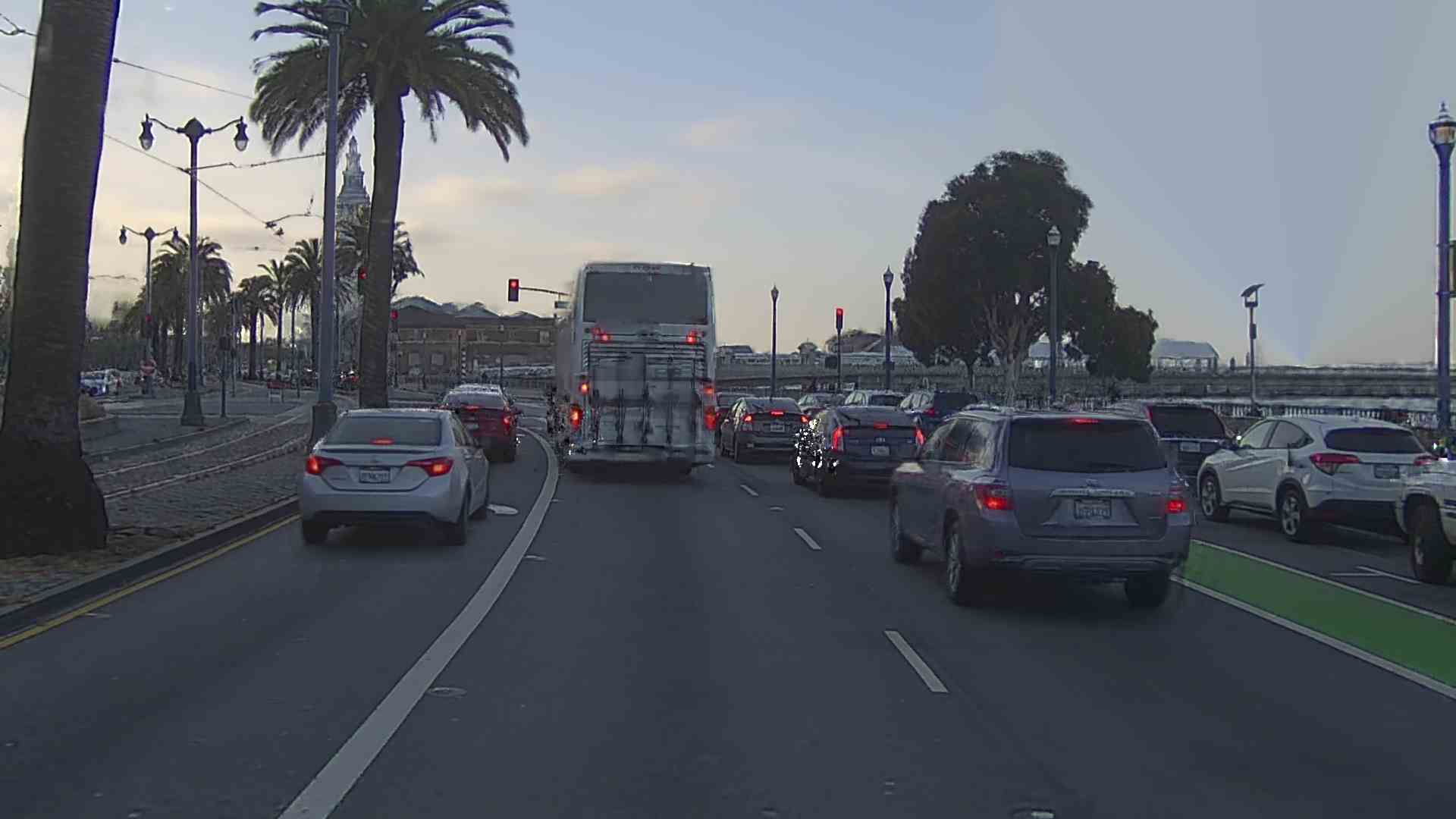} \\

        \includegraphics[clip=false, trim={0 0 0 0},width=\fgsize\columnwidth]{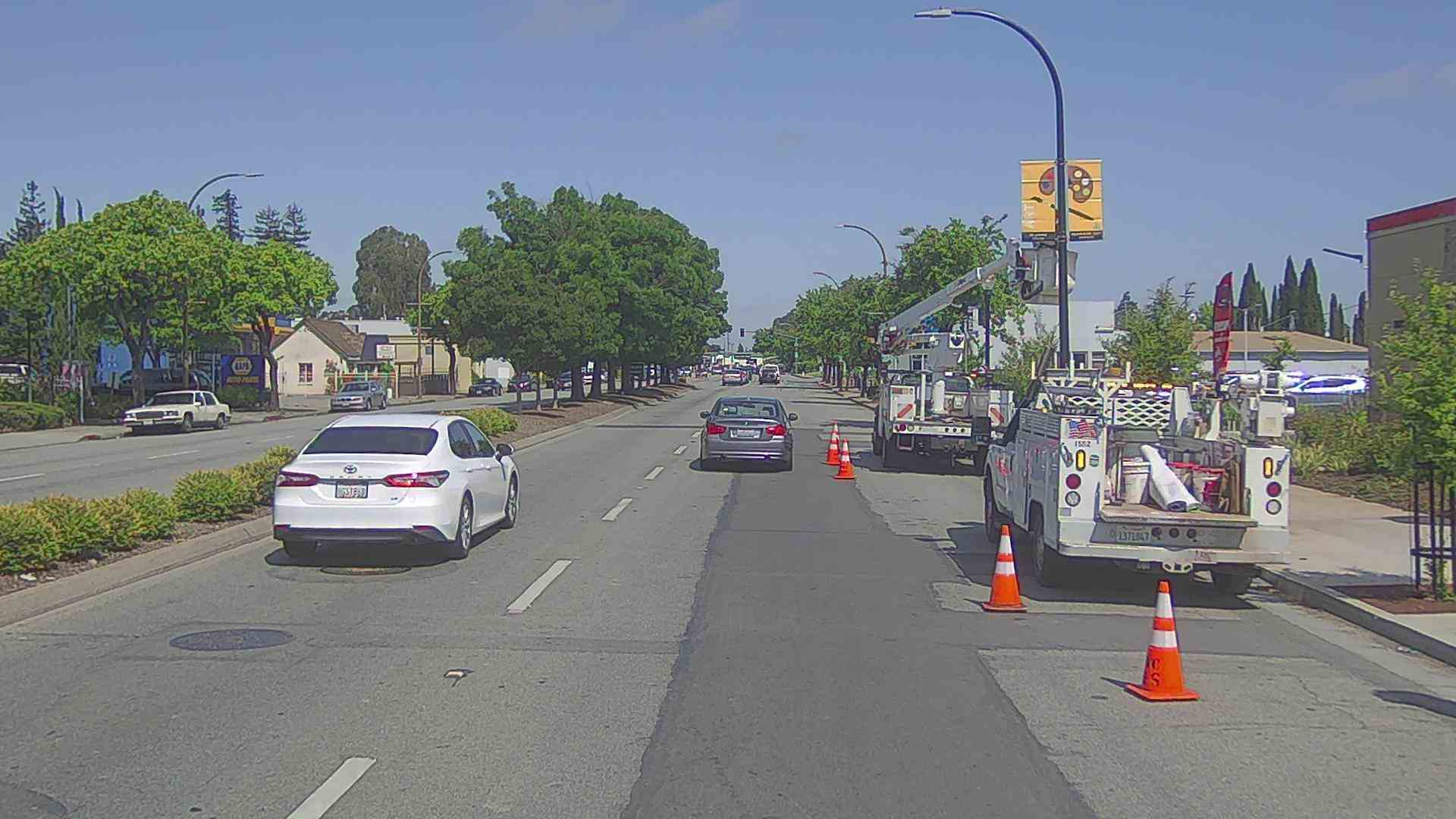} & 
        \includegraphics[clip=false, trim={0 0 0 0},,width=\fgsize\columnwidth]{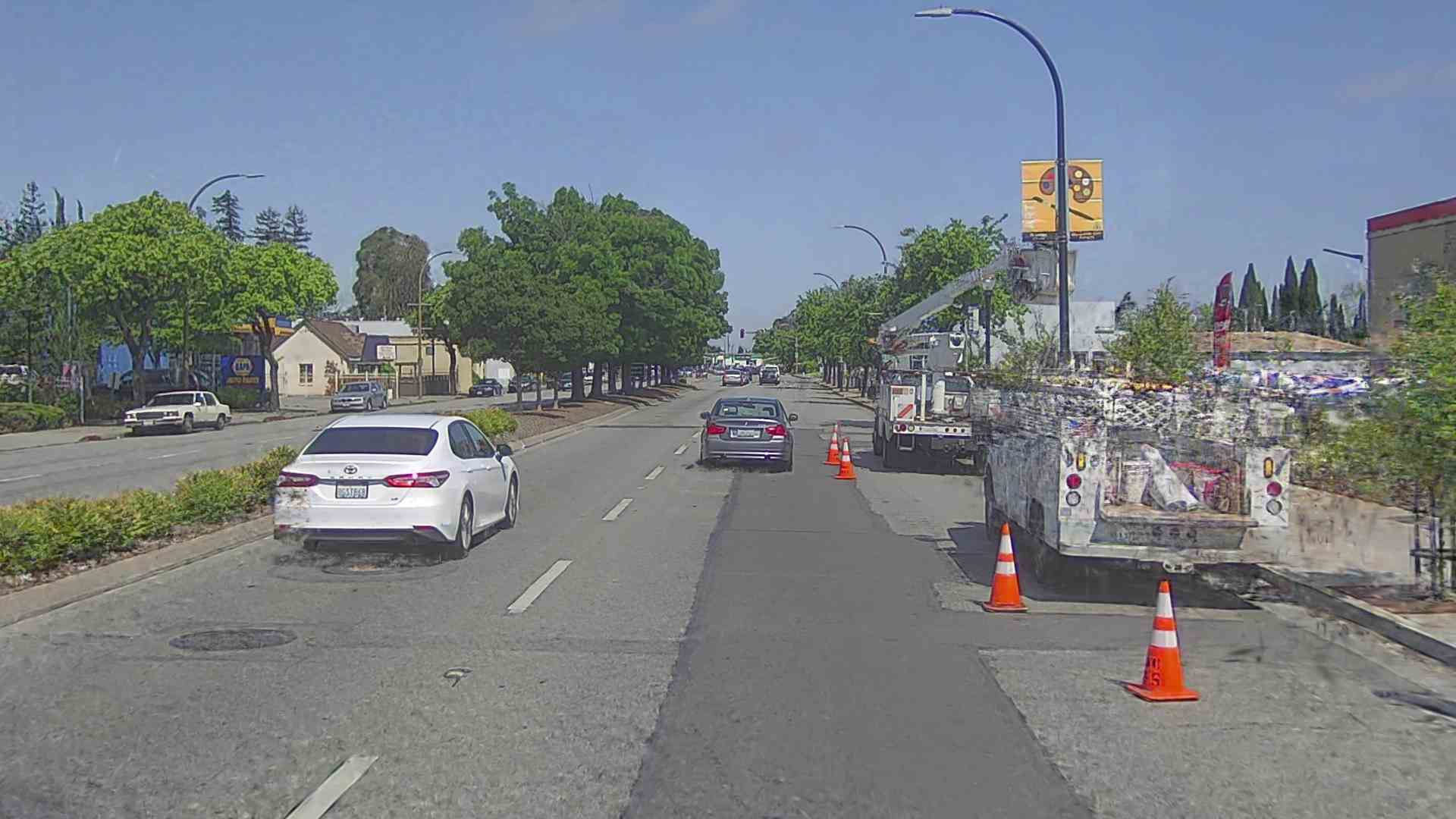} & 
        \includegraphics[clip=false, trim={0 0 0 0},,width=\fgsize\columnwidth]{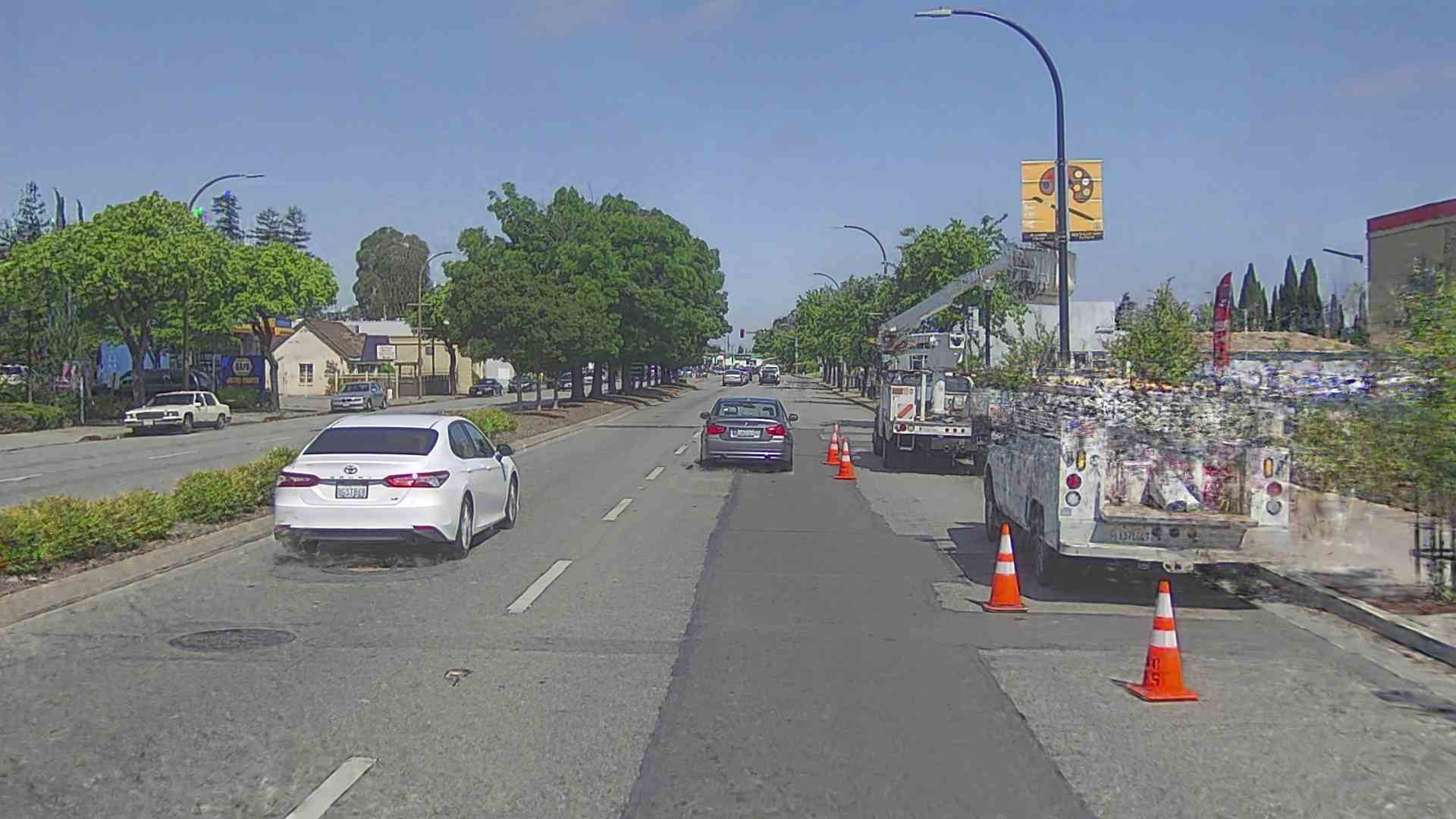} & 
        \includegraphics[clip=false, trim={0 0 0 0},,width=\fgsize\columnwidth]{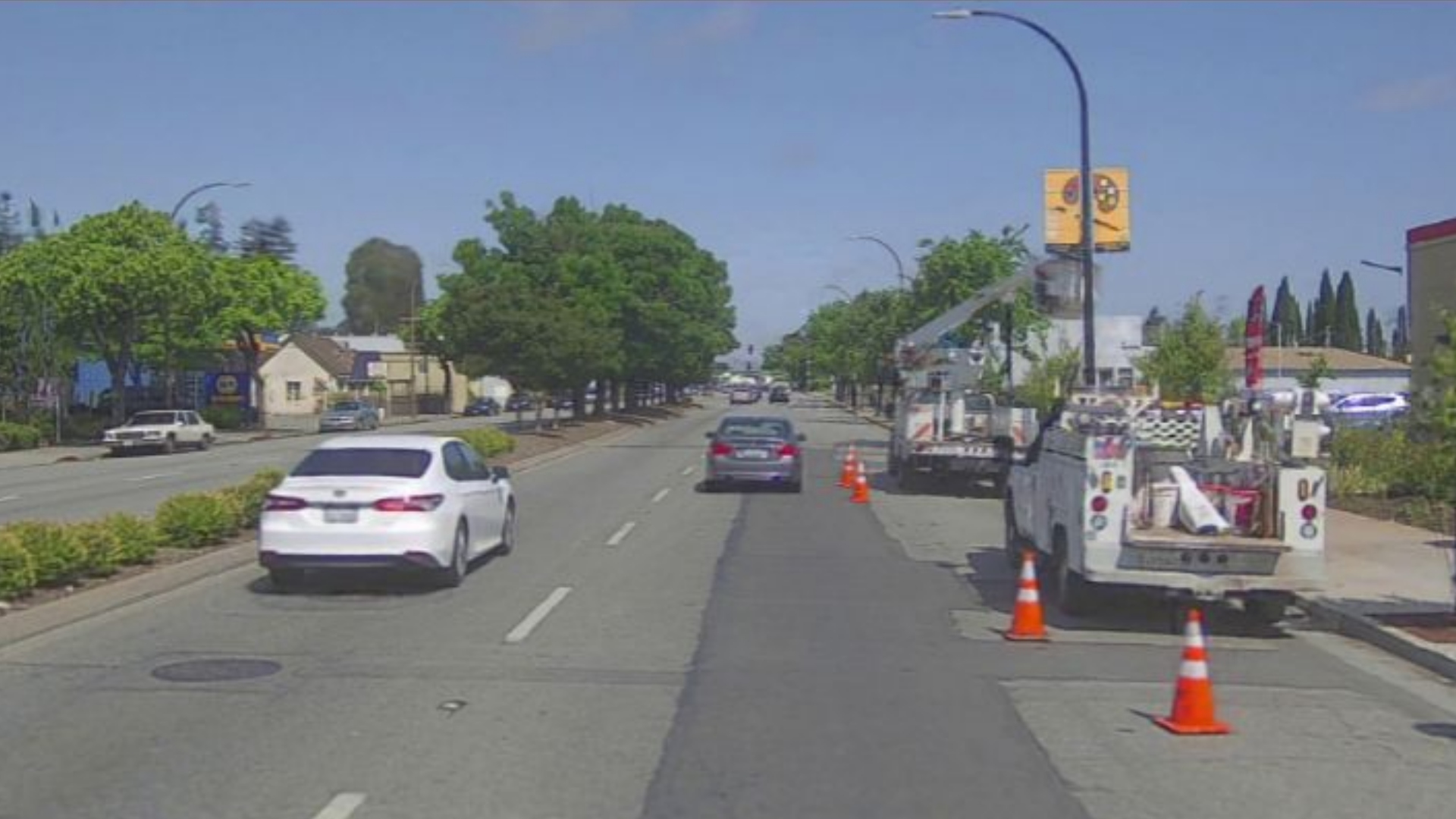} & 
        \includegraphics[clip=false, trim={0 0 0 0},,width=\fgsize\columnwidth]{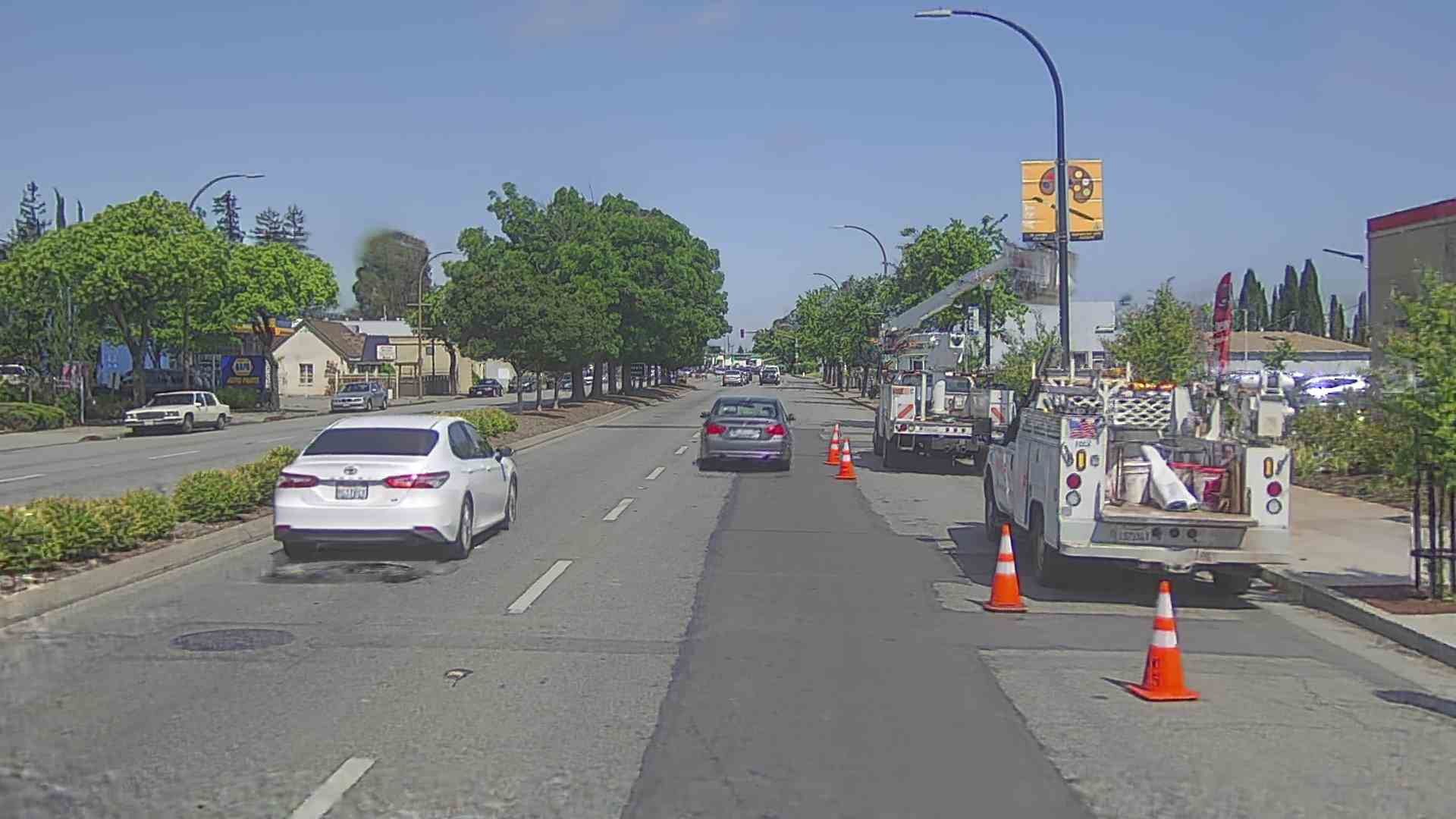} \\

        \includegraphics[clip=false, trim={0 0 0 0},width=\fgsize\columnwidth]{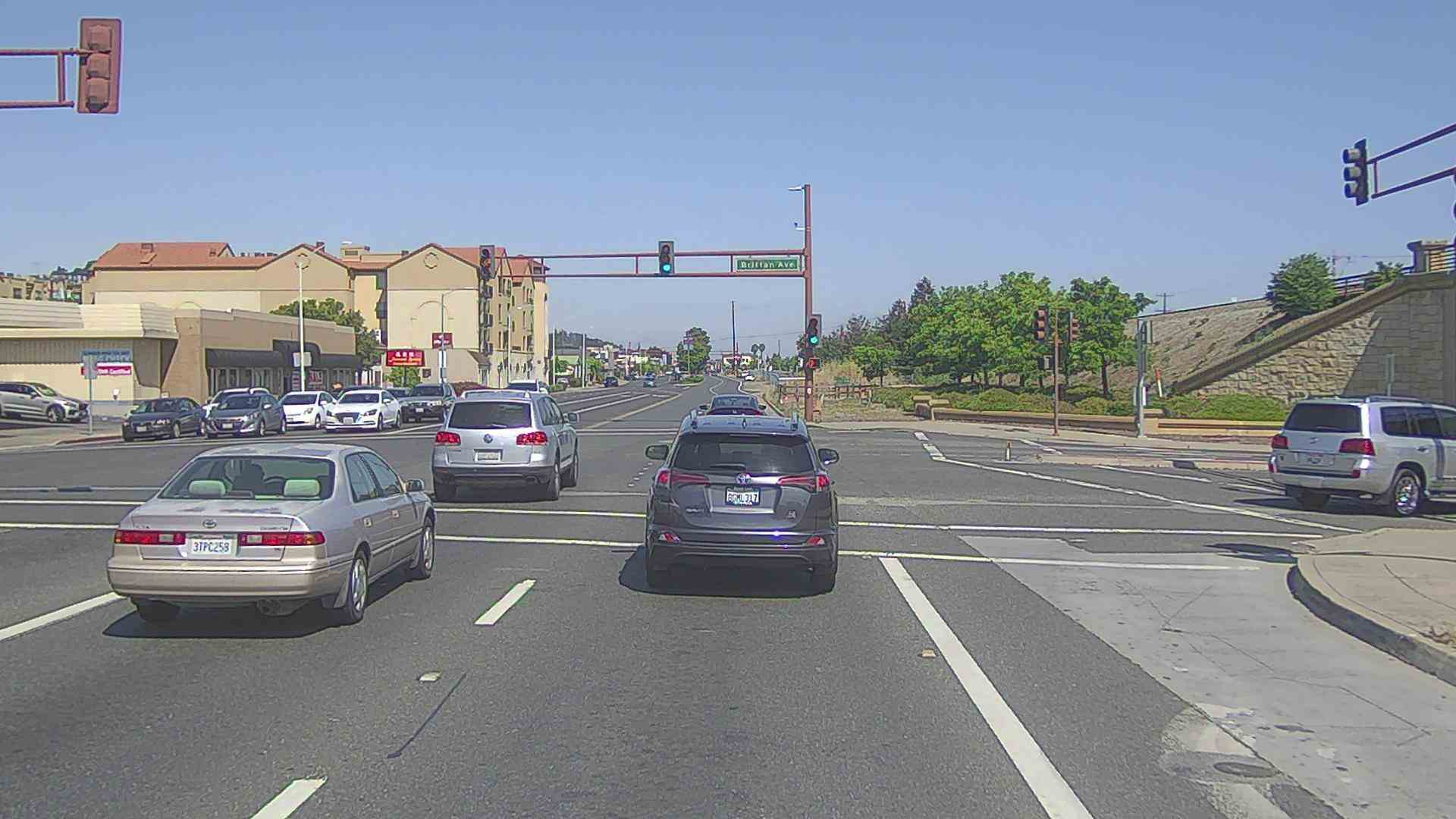} & 
        \includegraphics[clip=false, trim={0 0 0 0},,width=\fgsize\columnwidth]{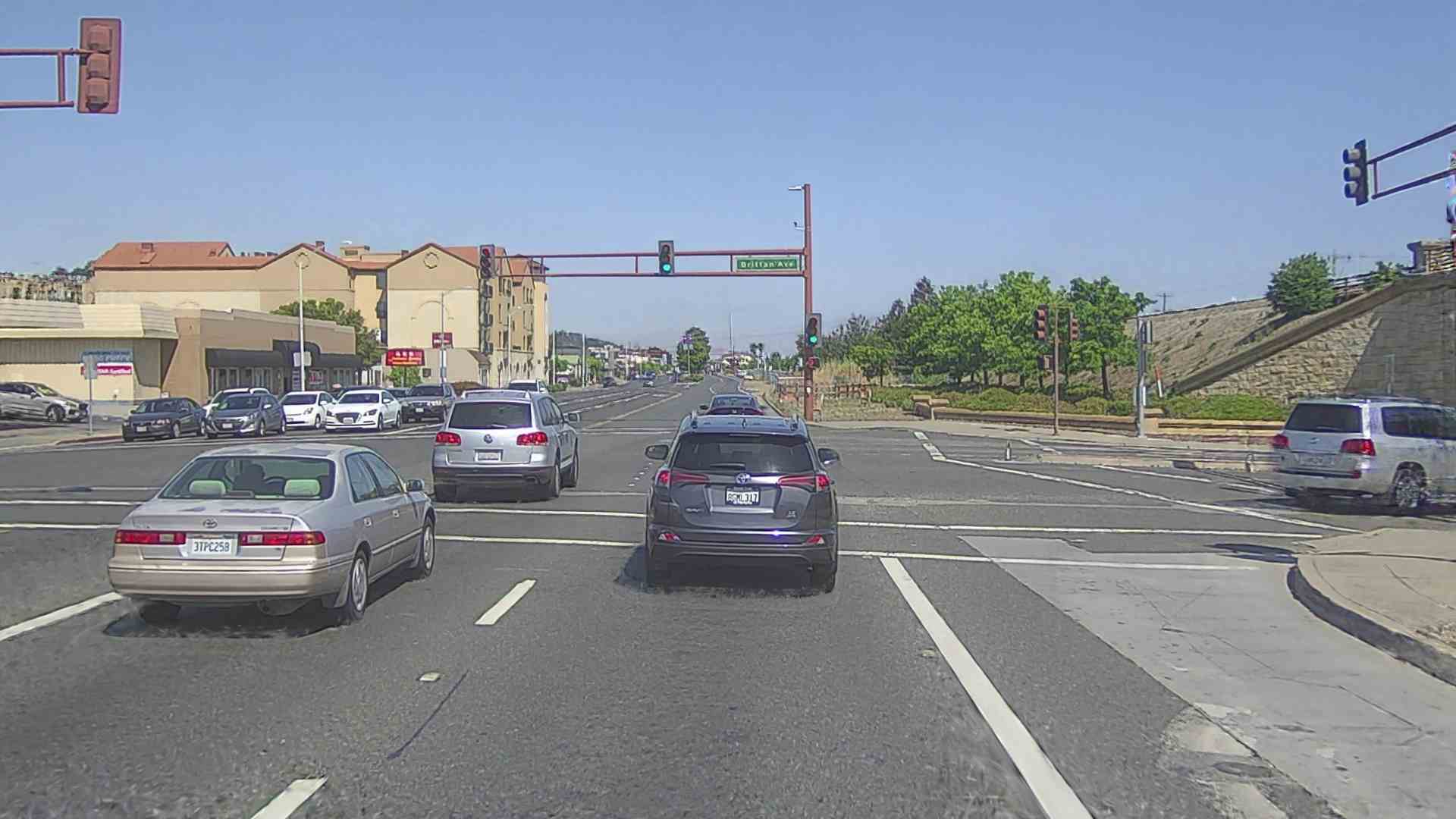} & 
        \includegraphics[clip=false, trim={0 0 0 0},,width=\fgsize\columnwidth]{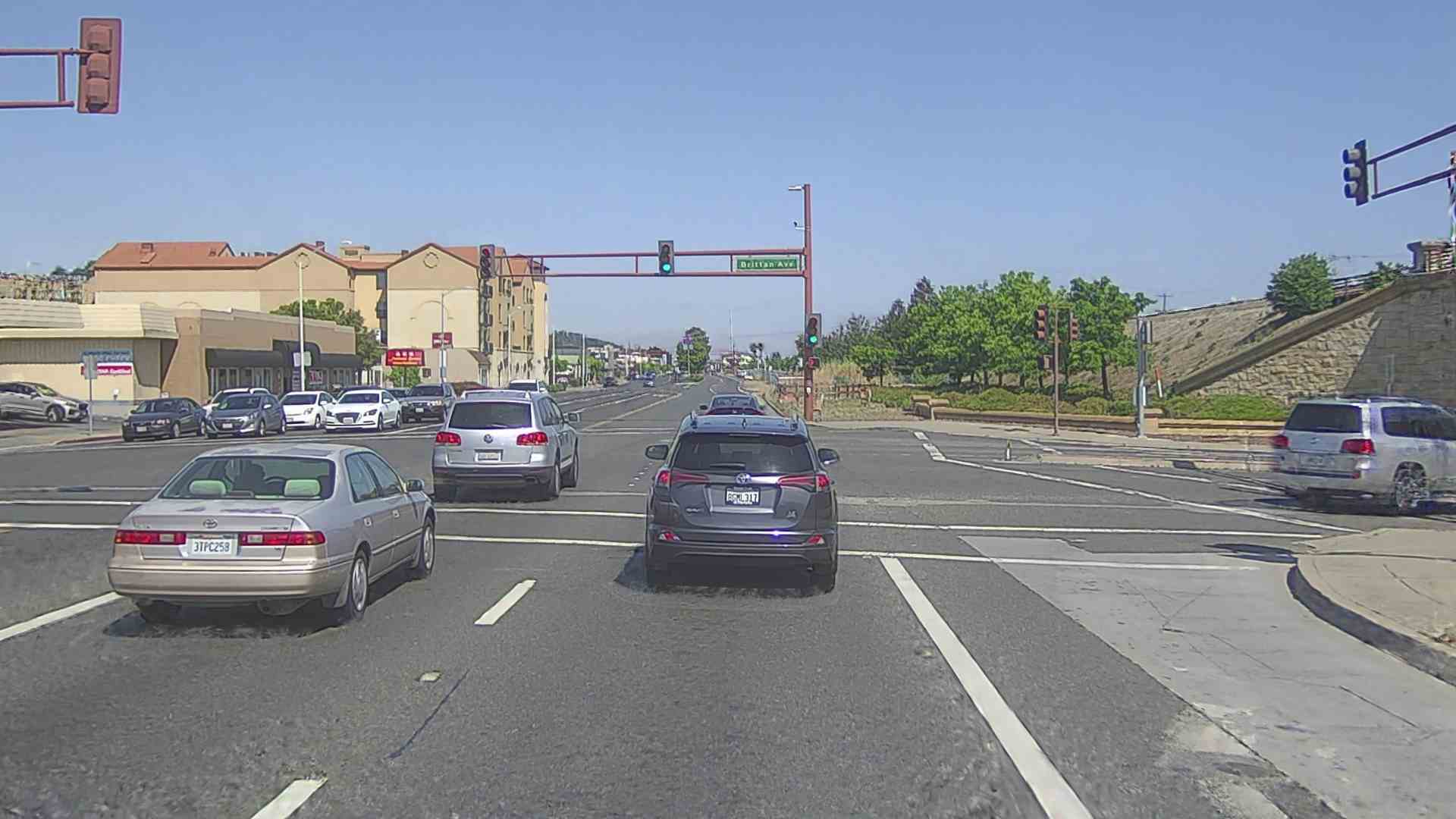} & 
        \missingfigure{0.19\linewidth}{1.75cm} & 
        \includegraphics[clip=false, trim={0 0 0 0},,width=\fgsize\columnwidth]{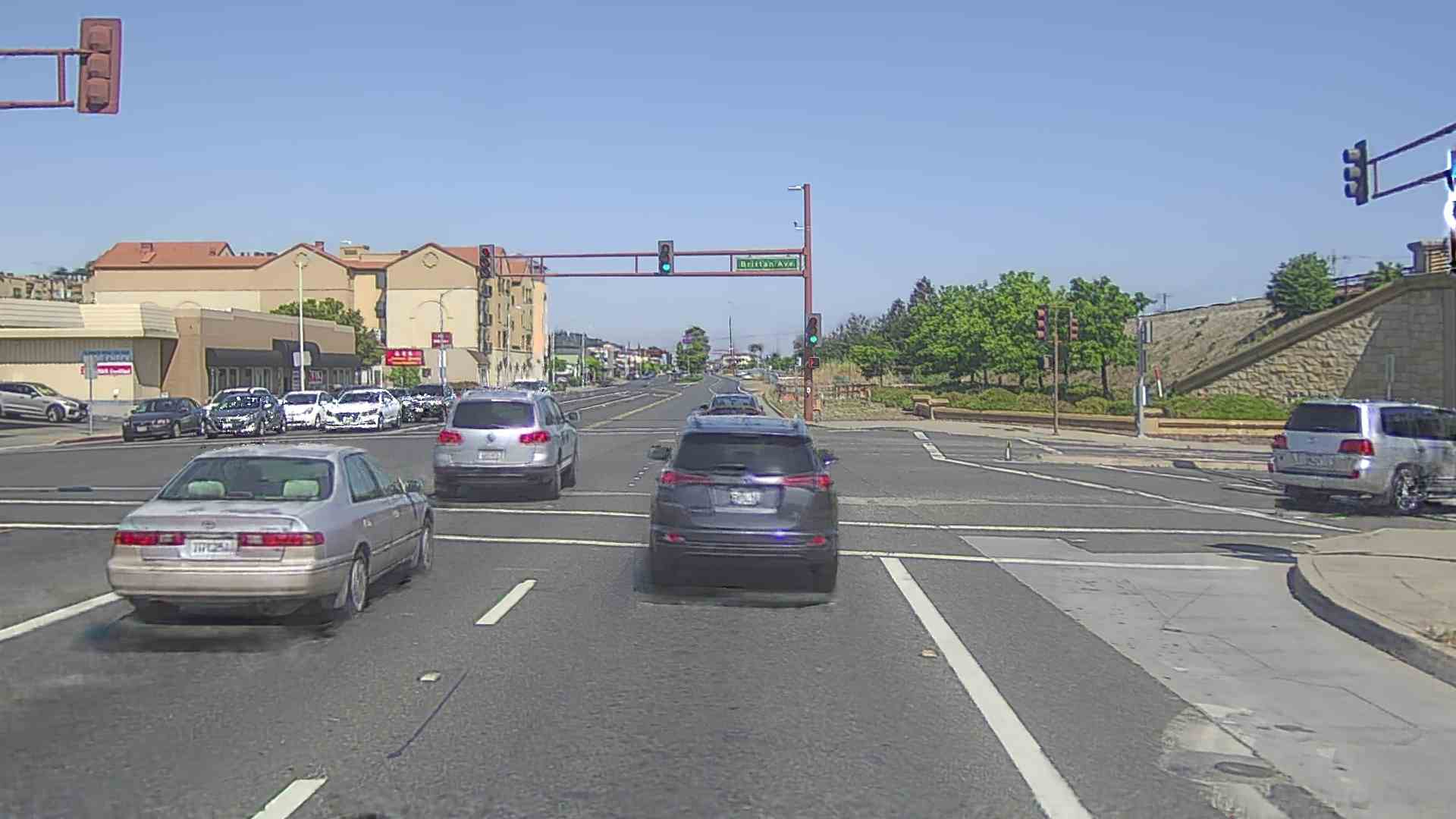} \\

        \includegraphics[clip=false, trim={0 0 0 0},width=\fgsize\columnwidth]{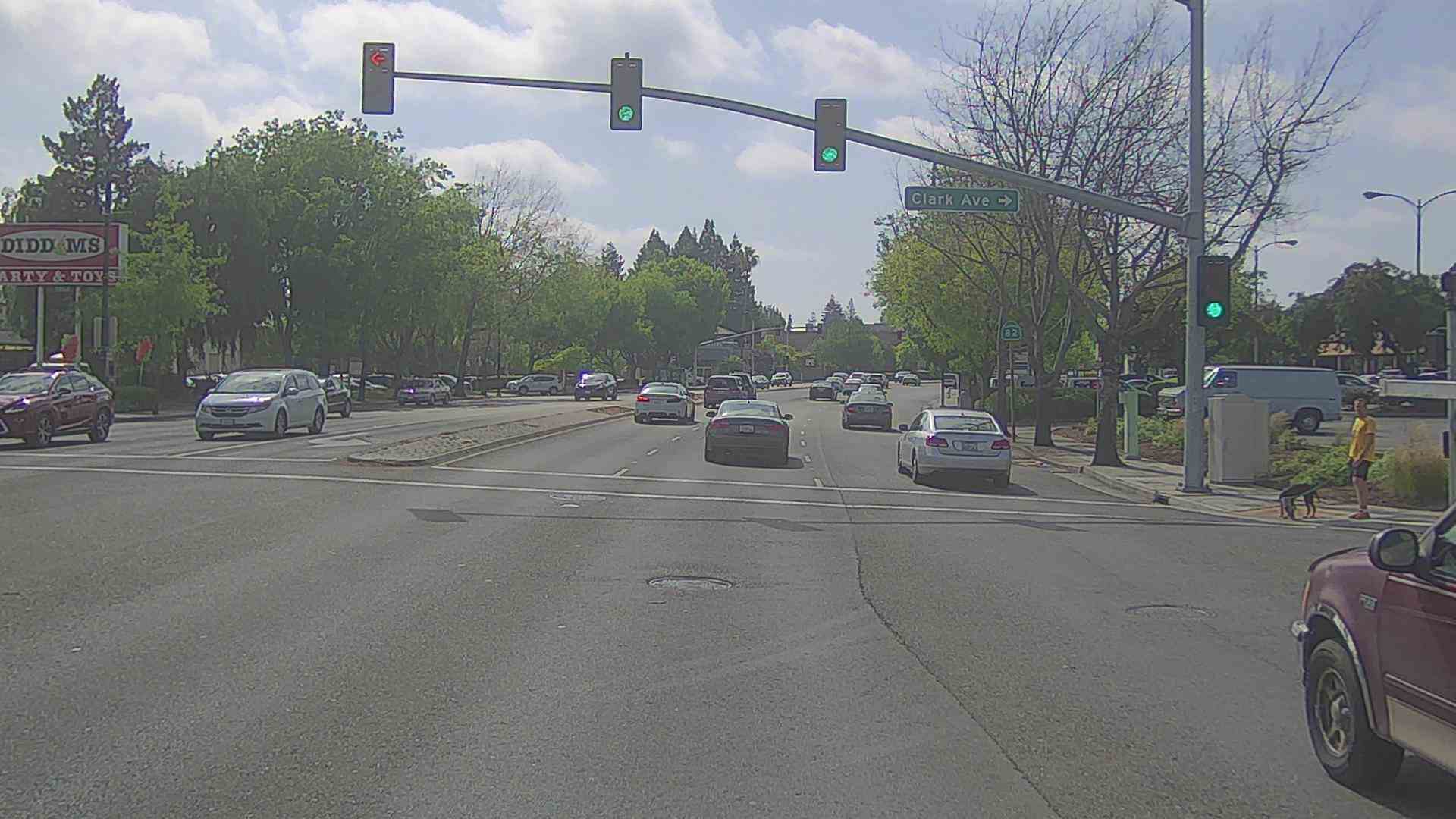} & 
        \includegraphics[clip=false, trim={0 0 0 0},,width=\fgsize\columnwidth]{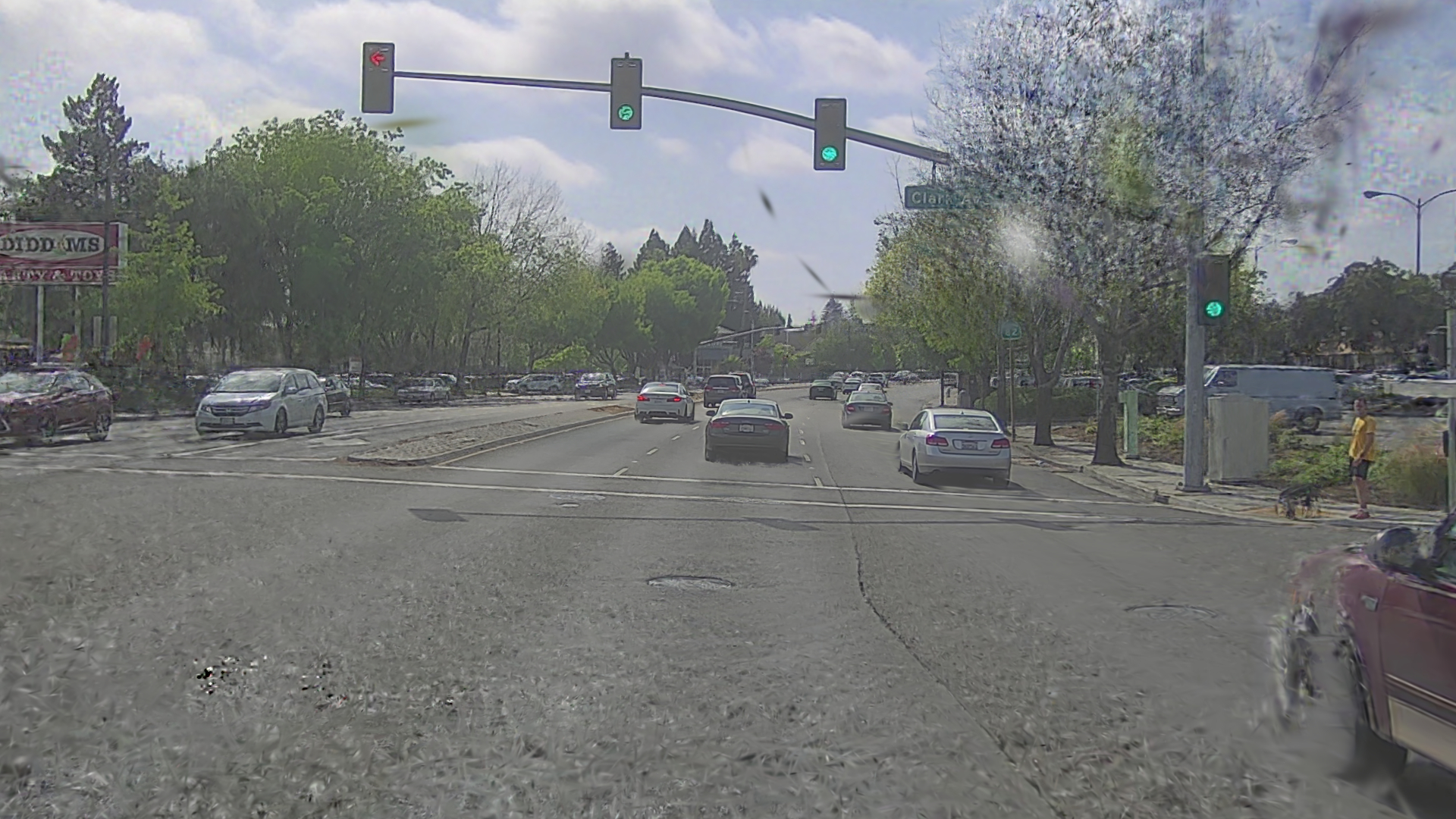} & 
        \includegraphics[clip=false, trim={0 0 0 0},,width=\fgsize\columnwidth]{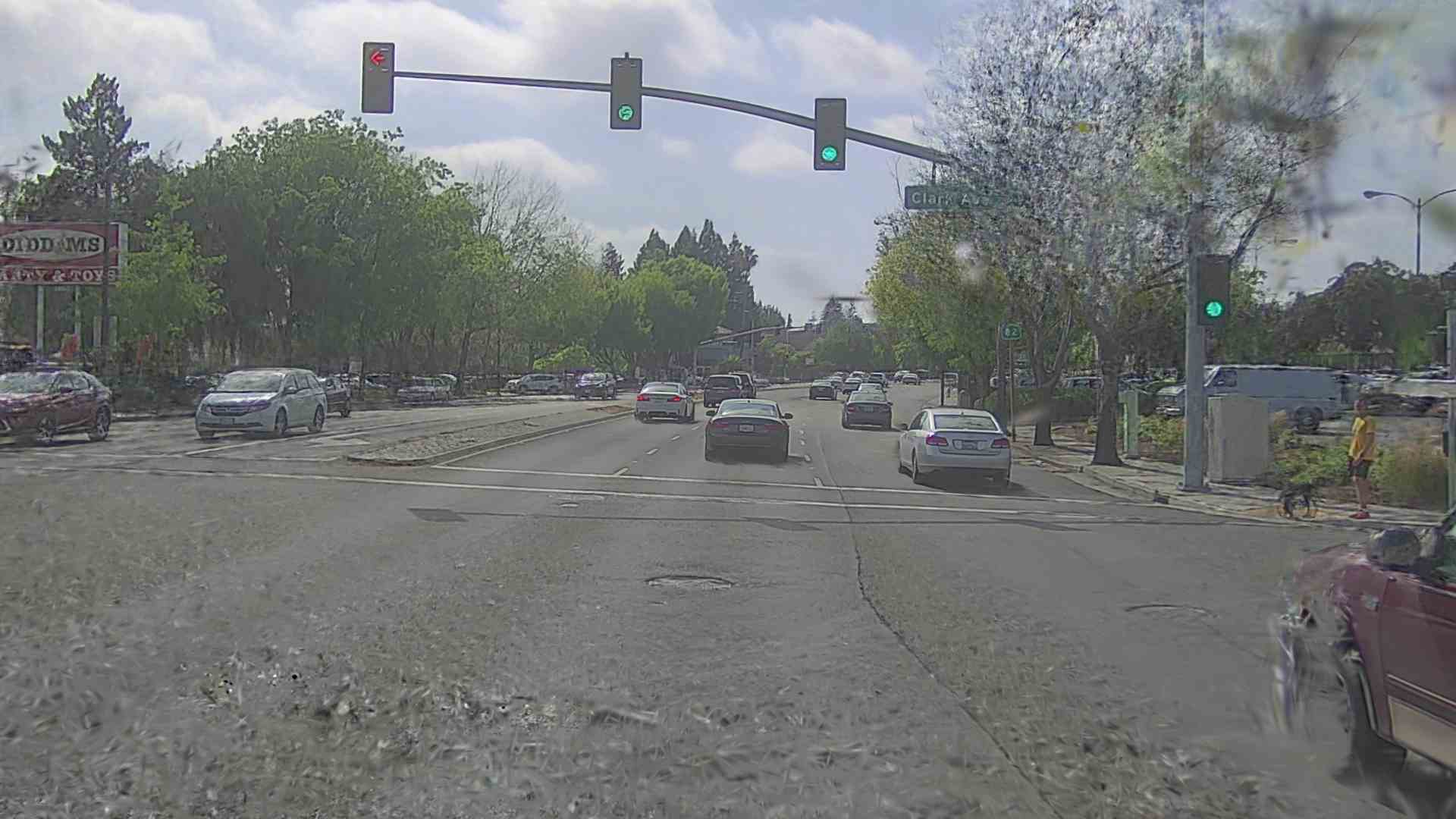} & 
        \missingfigure{0.19\linewidth}{1.75cm} & 
        \includegraphics[clip=false, trim={0 0 0 0},,width=\fgsize\columnwidth]{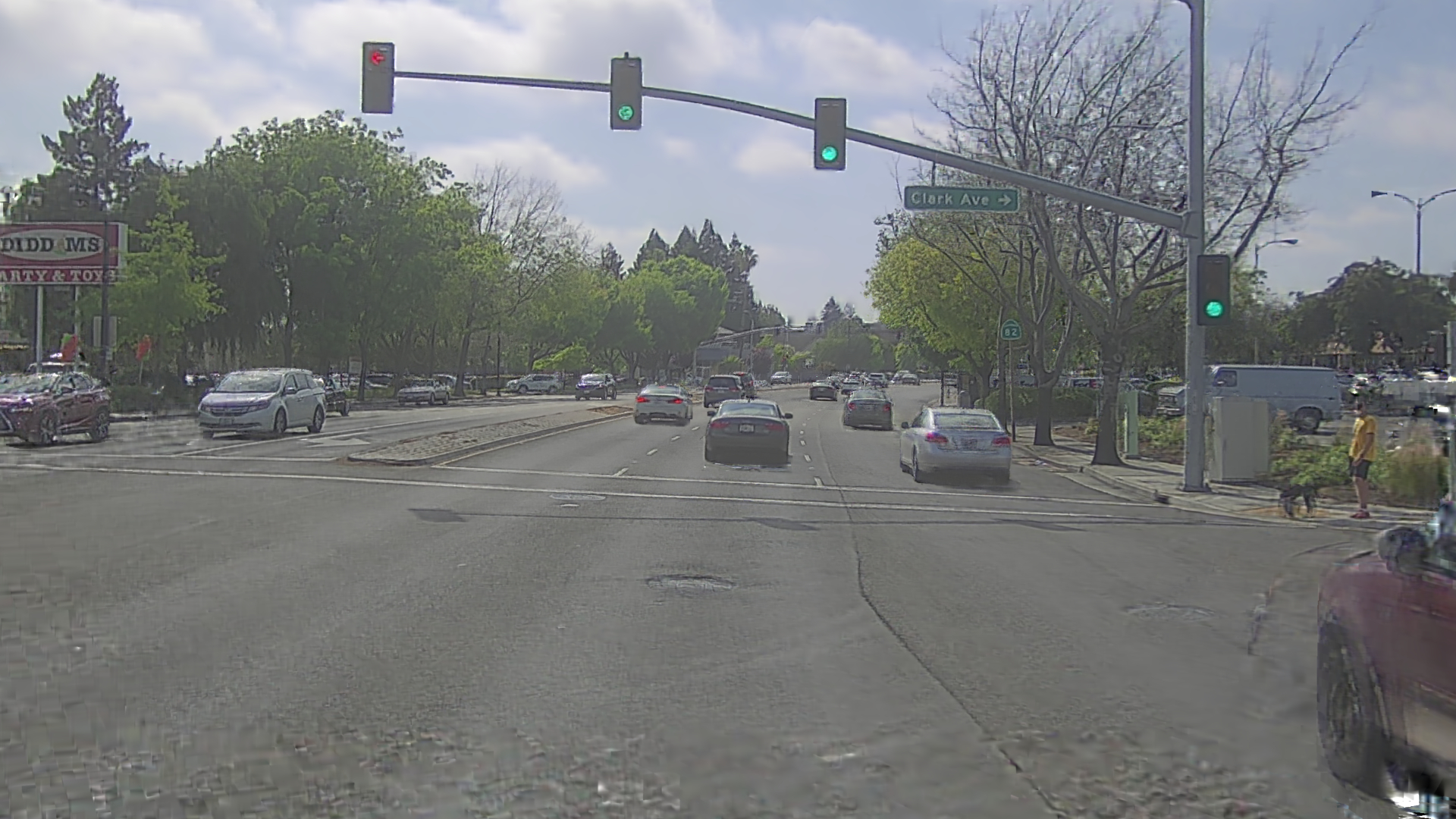} \\
        
    \end{tabular}
    \vspace{0.05cm}
    {\\
    \scriptsize $^{\ddagger}$ Figures taken from \cite{chen2025salf} original manuscript.}
    \vspace{-0.2cm}
    \caption{\textbf{Qualitative Comparison on Dynamic Driving Scenes.} Across multiple PandaSet sequences\cite{xiao2021pandaset}, \method~preserves coherent vehicle geometry and cleaner scene structure while avoiding the floating artifacts and ghosting visible in other baselines.}

    \label{fig:qualitative_results}

\end{figure*}

%% file: tabs/fid_shift_lane.tex
\begin{wraptable}[13]{r}{0.5\textwidth}
\centering

\vspace{-13pt}

\resizebox{0.48\linewidth}{!}{%
\begin{tabular}{lc}
    \toprule
    Method & FID $\downarrow$ \\ 
    \midrule
    UniSim$^{\ddagger}$ & 74.7 \\
    NeuRAD$^{\ddagger}$ & 72.3 \\
    StreetGS$^{*}$ & 72.6 \\
    OmniRe$^{*}$ & 71.6 \\
    \midrule
    \method{} & \underline{54.6} \\ 
    \method{} @20k & \textbf{53.5} \\ 
    \bottomrule
\end{tabular}%
}

{\scriptsize
$^{\ddagger}$ Results reported in~\cite{tancik2023neurad}.\\
$^{*}$ Obtained using DriveStudio~\cite{chen2025omnire}.
}
 \caption{\textbf{Shifted-Lane Extrapolation.} On PandaSet \method~reaches the lowest FID under a 2\,m lateral trajectory shift. 
}
\label{tab:geometry_extrapolation}
\end{wraptable}

%% file: tabs/ablation_init.tex
\begin{table*}

\centering
\resizebox{0.7\textwidth}{!}{%
\begin{tabular}{lcccccc}
    \cmidrule(l{0pt}r{0pt}){2-7}
    & \multicolumn{3}{c}{Perceptual} & & \multicolumn{2}{c}{Geometry} \\ 
    \cmidrule(l{2pt}r{2pt}){2-4} \cmidrule(l{2pt}r{2pt}){6-7}
    Method 
    & PSNR$\uparrow$ 
    & SSIM$\uparrow$ 
    & LPIPS$\downarrow$
    & & CD(m)$\downarrow$
    & F1@0.1(\%m)$\uparrow$ \\
    \midrule
w/o LiDAR subdivision & 28.11 & 0.845 & 0.203 & & 0.204 & 40.57 \\
Binary density init   & \underline{28.31} & \underline{0.853} & \underline{0.188} & & \textbf{0.190} & \textbf{43.86} \\
\midrule
Full (Ours @ 20k)     & \textbf{28.32} & \textbf{0.854} & \textbf{0.186} & & \underline{0.193} & \underline{43.23} \\
\midrule
\end{tabular}}
      \caption{\textbf{Background Initialization Ablation.} On five background-dominant PandaSet sequences, LiDAR-guided subdivision provides most of the gain, while binary density initialization perform similarly.
}
\label{tab:ablation_init}
\end{table*}

%% file: tabs/ablation_losses.tex
\begin{table*}[!t]

\centering
\resizebox{0.67\textwidth}{!}{%
\begin{tabular}{lcccccc}
    \cmidrule(l{0pt}r{0pt}){2-7}
    & \multicolumn{3}{c}{Perceptual} & & \multicolumn{2}{c}{Geometry} \\ 
    \cmidrule(l{2pt}r{2pt}){2-4} \cmidrule(l{2pt}r{2pt}){6-7}
    Method & PSNR$\uparrow$ & SSIM$\uparrow$ & LPIPS$\downarrow$ & & CD((m))$\downarrow$ & F1@0.1(\%)$\uparrow$ \\ 
    \midrule
    
    w/o $\mathcal{L}_{\text{depth}}$    & \textbf{25.32} & \textbf{0.806} & \textbf{0.234} & & 0.644 & 52.2 \\
    w/o $\mathcal{L}_{\text{normal}}$   &\underline{25.00} & \underline{0.797} & 0.246 & & 0.319 & 49.0 \\
    w/o sky model & 24.90 & 0.795 & 0.245 & & \underline{0.233} & \underline{66.1} \\
    \midrule
    \method @20k (ours)  & 24.84 & 0.793 & \underline{0.244} & & \textbf{0.231} & \textbf{66.3} \\
    \midrule
\end{tabular}}
\caption{\textbf{Supervision Loss Ablation.} LiDAR depth supervision is most important for geometry, without which the model degrades in CD and F1.
  }
\label{tab:ablation_losses}
\end{table*}

%% file: figs_tex/depth_ablation.tex
\begin{figure*}[t] 
    \centering
    \def\fgsize{0.24}
    \scriptsize
    \setlength{\tabcolsep}{0.0020\linewidth}
    \renewcommand{\arraystretch}{0.00}
    \begin{tabular}{ccc@{\hspace{7pt}}c}%
         w/o $\mathcal{L}_{\text{normal}}$ & w/o $\mathcal{L}_{\text{normal}}$ & w/o sky Model & \method \\ 
        
        \includegraphics[clip=false, trim={0 0 0 0},width=\fgsize\columnwidth]{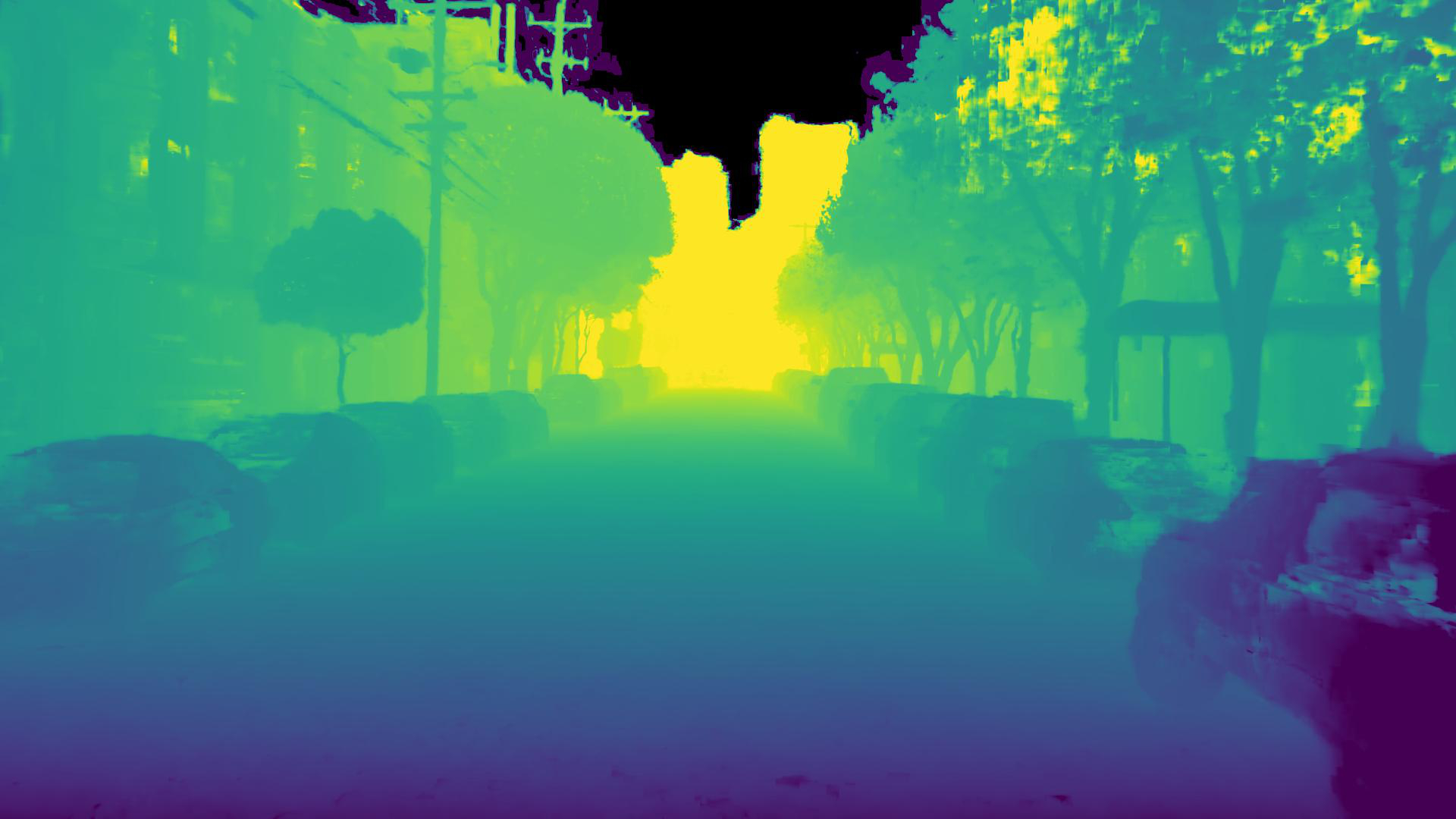} & 
        \includegraphics[clip=false, trim={0 0 0 0},,width=\fgsize\columnwidth]{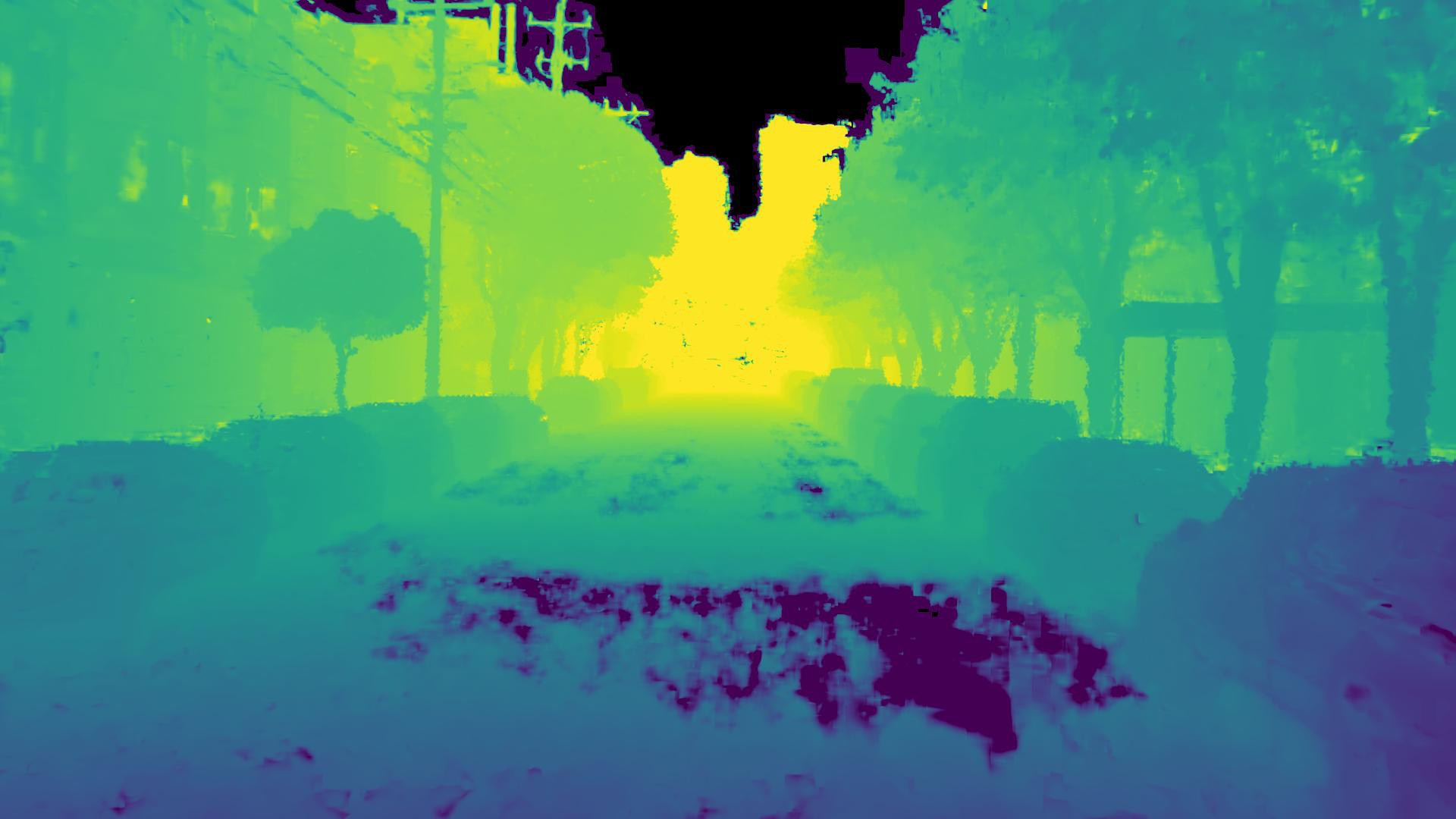} & 
        \includegraphics[clip=false, trim={0 0 0 0},,width=\fgsize\columnwidth]{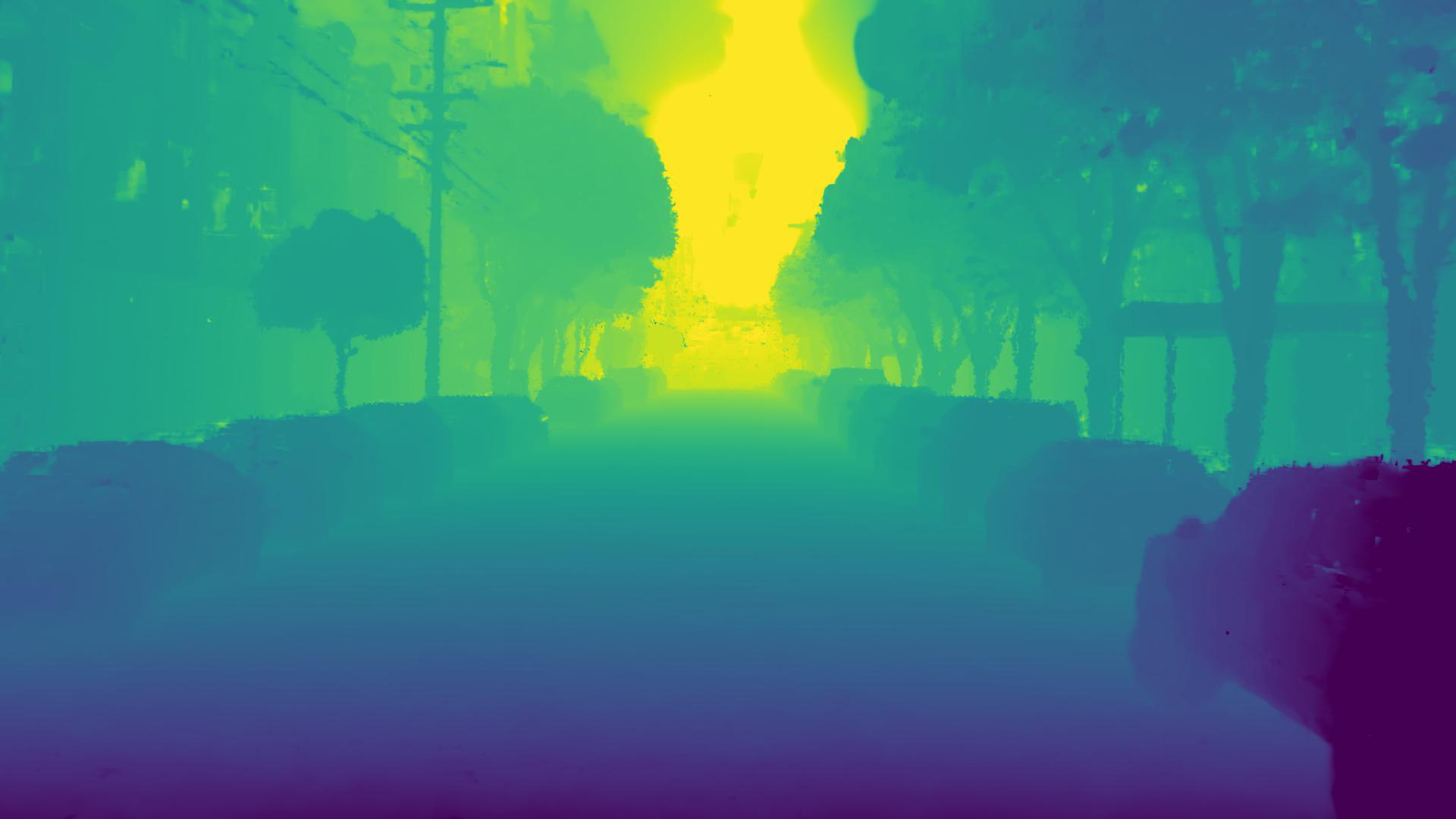} & 
        \includegraphics[clip=false, trim={0 0 0 0},,width=\fgsize\columnwidth]{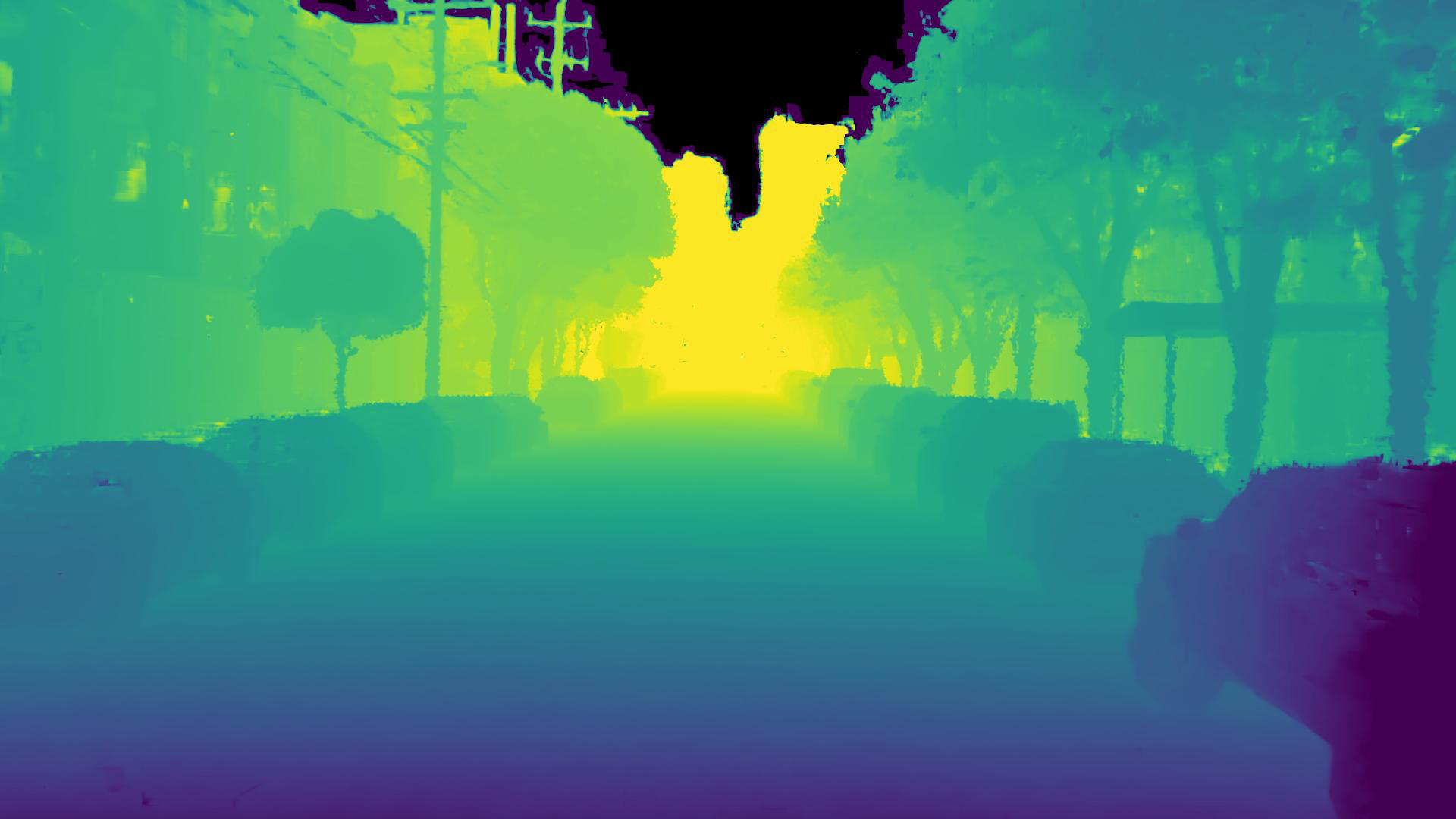} \\      
       
    \end{tabular}
    \caption{\textbf{Effect of Each Supervision Signal on Reconstructed Depth.} Quantitatively, removing normal supervision ($\mathcal{L}_{\text{normal}}$) leads to a substantial degradation in geometric metrics, yielding the lowest overall F1@0.1 score. Disabling LiDAR depth supervision ($\mathcal{L}_{\text{depth}}$) similarly compromises global structural boundaries, while omitting the sky model results in inaccurate far-depth allocations in the background. Our full configuration effectively combines these complementary signals to produce the most geometrically coherent and uniform depth maps.}

    \label{fig:ablation_depth_images}

\end{figure*}

%% file: secs/5_conclusion.tex
\subsection{Limitations}

Despite its strong geometric accuracy and efficient training, our method has several limitations.
First, the explicit voxel representation favors geometric fidelity but can limit perceptual quality compared to point-based splatting methods, which can more freely allocate primitives to capture fine appearance details.
Second, our pipeline assumes fixed camera poses and object tracks, making it sensitive to calibration, synchronization, and tracking errors which strongly affect accuracy~\cite{Herau2025PoseOF}.
Third, the current formulation is restricted to rigid dynamic actors and does not explicitly model non-rigid motion.
Finally, the minimum voxel resolution is coupled to the scene AABB size, creating a trade-off between representing distant structures and preserving fine geometric detail.

\section{Conclusion}

We presented \method, a compositional sparse voxel rendering framework for dynamic driving scene reconstruction. Our method extends sparse voxel rasterization~\cite{svr} to multi-octree environments, enabling a single-pass rendering formulation that decouples a static background from independently moving actors. Combined with a LiDAR-guided structural initialization and ray-interval sorting, this design effectively reduces floating artifacts while maintaining efficient training.

Experiments on the PandaSet benchmark~\cite{xiao2021pandaset} demonstrate state-of-the-art geometric reconstruction and competitive perceptual quality, while achieving faster training than prior methods. Beyond reconstruction accuracy, the proposed structured representation provides a stable foundation for downstream applications such as scene editing and semantic understanding, albeit with some trade-offs in perceptual flexibility due to its explicit voxel design.

Future work includes joint optimization of camera and object poses, modeling of non-rigid dynamics, and decoupling distant scene regions from the global AABB to improve scalability and reconstruction fidelity in large-scale environments.

%% file: secs/supp.tex
\appendix

\section{Algorithmic Multi-Octree Render}
\label{sec:algorithm}
Our rendering pipeline handles a static background ($\mathcal{O}_{bg}$) and a dynamic sequence of $N$ independently moving foreground asset octrees $\{\mathcal{O}_i\}_{i=1}^N$ simultaneously. 
Rather than performing complex runtime spatial tree sorting, we implement a macro-level ray-interval decomposition strategy. For each camera ray $\mathbf{r}(\tau) = \mathbf{o} + \tau\mathbf{d}$, we calculate explicit intersections with the oriented 3D bounding boxes of all active dynamic foreground assets to obtain a front-to-back pre-ordered sequence of intervals $\mathcal{I}$. 

Each consecutive segment is processed by SVRaster volume rendering accumulation function $\mathcal{F}$~\cite{svr}, which updates the global pixel properties state tuple $\text{S} = (\text{C}, \text{D}, \text{N}, \text{T}, \tau_{curr})$ where $\text{C}$ tracks accumulated color, $\text{D}$ denotes rendered surface depth, $\text{N}$ aggregates voxel normals, $\text{T}$ tracks residual transmittance, and $\tau_{curr}$ marks the active ray distance boundary.
Specifically, $\mathcal{F}$ processes the camera ray parameters to filter the sorted voxel list of the target octree, isolating and evaluating only the primitives that reside within the active bounded interval $[\tau_{in}^i, \tau_{out}^i]$. By tracking and continuously passing this state tuple $S$ across consecutive interval calls, our design guarantees mathematically correct multi-volume occlusion and exact transmittance propagation across all coordinate boundaries. This step-by-step composition pass is formally detailed in \cref{alg:rendering}.

\section{Datasets}
\label{sup:sup_atasets}
Our evaluation is done in Pandaset sequences~\cite{xiao2021pandaset}. All experiments are done at original resolution using only the front-facing camera. Each sequence is split chronologically into 40 training and 40 evaluation frames via even/odd indexing, resulting in total 400 training and 400 evaluation frames.
\begin{itemize}
    \item \textbf{Main Dataset} Our primary evaluation follows the protocol defined by SaLF \cite{chen2025salf}, using the following sequences: 001, 011, 016, 028, 053, 063, 084, 106, 123, and 158.
    \item \textbf{Static-Dominant Split:} Consists of five background-dominant PandaSet sequences containing sparse dynamic actors (028, 029, 53, 55 and 63), specifically chosen to isolate background geometry from vehicle movements.
    \item \textbf{Supervision Ablation Dataset:} A targeted subset of the primary benchmark sequences, consisting of sequences 011, 016, 028, 106, 158.
\end{itemize}

All methods and baseline benchmarks are evaluated on hardware equivalent to an NVIDIA RTX 3090 GPU to ensure direct performance parity with reported execution timelines.

\begin{algorithm}[t]
\caption{Scene Rendering with Segmented Volume Rendering at a timestep $t$.}
\label{alg:rendering}

\begin{algorithmic}[1]
\State \textbf{Input:} Background octree $\mathcal{O}_{bg}$, asset octrees $\{\mathcal{O}_i\}_{i=1}^N$, transformations $\{T_t^i\}_{i=1}^N$, camera ray $\mathbf{r}$, sorted intervals $\mathcal{I}$.
\vspace{1pt}

\State $S \gets (\text{C}=0,\ \text{D}=0,\ \text{N}=0,\ \text{D}=1 )\ $, $\tau_{\mathrm{curr}} \gets 0$

\For{$i \in \{1,\dots,N\}$}

    \Comment{Render background before object entry}
    \State $\text{S} \gets \mathcal{F}(\mathcal{O}_{bg},\, \mathbf{r},\, [\tau_{\mathrm{curr}},\, \tau_{\mathrm{in}}^i],\, \text{S})$

    \Comment{Render object in canonical local space}
    \State $\mathbf{r}_{\mathrm{local}} \gets (T_t^i)^{-1}(\mathbf{r})$
    \State $\text{S} \gets \mathcal{F}(\mathcal{O}_i, \mathbf{r}_{\mathrm{local}},\, [\tau_{\mathrm{in}}^i, \tau_{\mathrm{out}}^i],\, \text{S})$

    \Comment{Advance to next interval}
    \State $\tau_{\mathrm{curr}} \gets \tau_{\mathrm{out}}^i$

    \If{$\text{T} < \varepsilon_{\mathrm{stop}}$}
        \State \textbf{break}
    \EndIf

\EndFor

\Comment{Render remaining background}
\State $\text{S} \gets \mathcal{F}(\mathcal{O}_{bg},\, \mathbf{r},\, [\tau_{\mathrm{curr}},\, \tau_{\max}], \text{S})$

\end{algorithmic}
\end{algorithm}

\section{Baselines}
We benchmark our framework against representative approaches from three established lines of work in dynamic driving scene reconstruction:

\begin{itemize}
    \item \textbf{Continuous Volumetric Frameworks:} UniSim~\cite{yang2023unisim} and NeuRAD~\cite{tancik2023neurad} model driving environments via continuous neural radiance fields optimized for closed-loop simulation. Following standard benchmarking practices, we report the performance metrics for both baselines directly as cited in the SaLF paper~\cite{chen2025salf}.
    \item \textbf{Unconstrained Point-Based Splatting:} OmniRe~\cite{chen2025omnire} and StreetGS~\cite{yan2024street} leverage object-centric, anisotropic 3D Gaussian primitives. While achieving interactive rendering speeds, their unconstrained optimization layout frequently causes geometric degradation, floating artifacts, and unpredictable memory spikes in large outdoor environments. We evaluate these baselines using the implementations within the DriveStudio codebase under the \textit{paper legacy} configuration, maintaining default hyperparameter schedules to ensure direct hardware performance parity.
    \item \textbf{Hybrid Voxel-Implicit Layouts:} SaLF~\cite{chen2025salf} (both Base and Large configurations) use a sparse voxel layout but embeds a local implicit field inside each voxel node. Our framework completely inherits this benchmark protocol. However, because their official source code is currently unavailable, obtaining a full set of metrics across all evaluation dimensions remains difficult, restricting our direct comparison to the values provided in their original publication.
\end{itemize}

\section{Optimization Details}

\paragraph{Training Losses}\label{sec:losses}
Our multi-octree framework and spherical sky model are jointly optimized end-to-end via a multi-modal supervision loss function:
\begin{equation}
    \mathcal{L} = \lambda_{\text{color}}\mathcal{L}_{\text{color}} + \lambda_{\text{depth}}\mathcal{L}_{\text{depth}} + \lambda_{\text{normal}}\mathcal{L}_{\text{normal}} + \lambda_{\text{sky}}\mathcal{L}_{\text{sky}}
\end{equation}
\paragraph{Color Reconstruction Loss ($\mathcal{L}_{\text{color}}$)}
The primary photometric objective is a combination of an $L_1$ pixel loss and a structural similarity (SSIM) term computed between the rendered image $C$ and the ground-truth camera observation $C_{\text{gt}}$:
\begin{equation}
    \mathcal{L}_{\text{color}} = \lambda_{\text{MSE}} \lambda_{\text{MSE}} + \lambda_{\text{SSIM}} (1 - \text{SSIM}(C, C_{\text{gt}}))
\end{equation}
\paragraph{Sparse Depth Supervision Loss ($\mathcal{L}_{\text{depth}}$)}
To enforce strict geometric constraints and eliminate volumetric alignment ambiguities, we apply an $L_1$ depth loss:
\begin{equation}
    \mathcal{L}_{\text{depth}} = \| D - D_{\text{LiDAR}} \|_1
\end{equation}
where $D$ is the accumulated voxel depth along the camera ray path and $D_{\text{LiDAR}}$ represents the sparse, time-synchronized ground-truth depth values generated by projecting the $360^{\circ}$ LiDAR point cloud directly into the camera frame.
\paragraph{Dense Surface Normal Loss ($\mathcal{L}_{\text{normal}}$)}
To supplement areas where LiDAR points are sparse or surface texturing is uniform, we incorporate a geometric alignment term:
\begin{equation}
    \mathcal{L}_{\text{normal}} = 1 - \frac{N \cdot N_{\text{pseudo}}}{\|N\|_2 \|N_{\text{pseudo}}\|_2}
\end{equation}
This loss maximizes the cosine similarity between our rendered voxel surface normals $N$ and dense pseudo-ground-truth surface normals $N_{\text{pseudo}}$ generated  by using the DepthAnythingV2~\cite{depth_anything_v2} foundation model and following SVRaster ~\cite{svr} computation from depth maps to normal maps.

\paragraph{Sky Transmittance Mask Loss ($\mathcal{L}_{\text{sky}}$)}
To prevent the invalid allocation of sparse voxels in unbounded free space, we isolate the sky region using a binary mask $\mathcal{M}_{\text{sky}}$ derived from zero-depth LiDAR boundaries. We enforce maximum global ray transmittance via a mean squared error (MSE) objective :
\begin{equation}
    \mathcal{L}_{\text{sky}} = \frac{1}{|\mathcal{M}_{\text{sky}}|} \sum_{i \in \mathcal{M}_{\text{sky}}} (T_{\text{final}, i} - 1)^2
\end{equation}
where $T_{\text{final}}$ is the residual transmittance accumulated at the terminal boundary of the multi-octree composition pass. This ensures that background primitives are completely penalized within the sky dome, routing all background color generation exclusively through the 2D  environment map.

\subsection{Adaptive Optimization Schedule}
\label{sec:sup_implementation}

We adopt the ascending loss schedule of SVRaster~\cite{svr}. To dynamically determine convergence, we monitor structural rendering improvements every 1,000 iterations using $\Delta_{\text{SSIM}} = \text{SSIM}_t - \text{SSIM}_{t-1000}$. Once $\Delta_{\text{SSIM}}$ falls below a threshold $\epsilon$, background octree ($\mathcal{O}_{\text{bg}}$) subdivision is deactivated. The background octree then undergoes two pruning cycles (2,000 iterations total) to eliminate residual low-density voxels, followed by a final 3,000-iteration refinement phase with frozen subdivision and pruning. In contrast, the foreground asset octrees $\{\mathcal{O}_i\}$ bypass early deactivation and subdivide independently throughout training to accommodate their longer optimization window.

\section{Additional Results}
\paragraph{Depth Map Visualization} In~\cref{fig:depth_scenes}, we provide extended qualitative visualizations of the reconstructed depth maps across multiple dynamic sequences, which are generated in a single pass by evaluating our multi-volume composite rendering function.
\input{figs_tex/depths_scenes}

\begin{figure}[!t]
\centering
\includegraphics[width=0.4\textwidth]{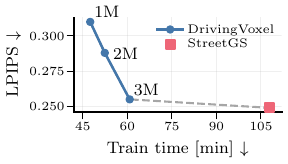}
\caption{\textbf{Quality-efficiency tradeoff.} Comparison of LPIPS versus training time for \method~under different voxel budgets (1M, 2M, and 3M) against StreetGS.  \method~achieves a highly competitive perceptual quality while requiring significantly less training time, allowing a predictable control over the performance-efficiency tradeoff.}
\label{fig:tradeoff}
\end{figure}

\paragraph{Quality-Efficiency Tradeoffs}Fig.~\ref{fig:tradeoff} reports LPIPS and training time for DrivingVoxel under different maximum voxel budgets (1M, 2M, and 3M) alongside StreetGS. As shown in the plot, \method~ reaches competitive perceptual quality much faster than StreetGS, with training time ranging from approximately 47 to 60 minutes compared to StreetGS's 107 minutes. While increasing the voxel budget from 1M to 3M drastically lowers the LPIPS error to approach StreetGS quality, it only marginally increases training overhead. These results confirm that limiting the total voxel allocation gives users direct and predictable control over the quality-efficiency tradeoff. This is a distinct advantage that Gaussian-based methods like StreetGS cannot offer, as their memory usage and primitive counts grow randomly during training.

%% file: figs_tex/depths_scenes.tex
\begin{figure*}[!t] 
    \centering
    \def\fgsize{0.27} %
    \scriptsize
    \setlength{\tabcolsep}{0.0020\linewidth}
    \renewcommand{\arraystretch}{0.00}
    \begin{tabular}{ccc}
         OmniRe~\cite{chen2025omnire} & StreetGS~\cite{yan2024street} & \method ~(ours) \\ 
         
        \includegraphics[clip=false, trim={0 0 0 0},width=\fgsize\columnwidth]{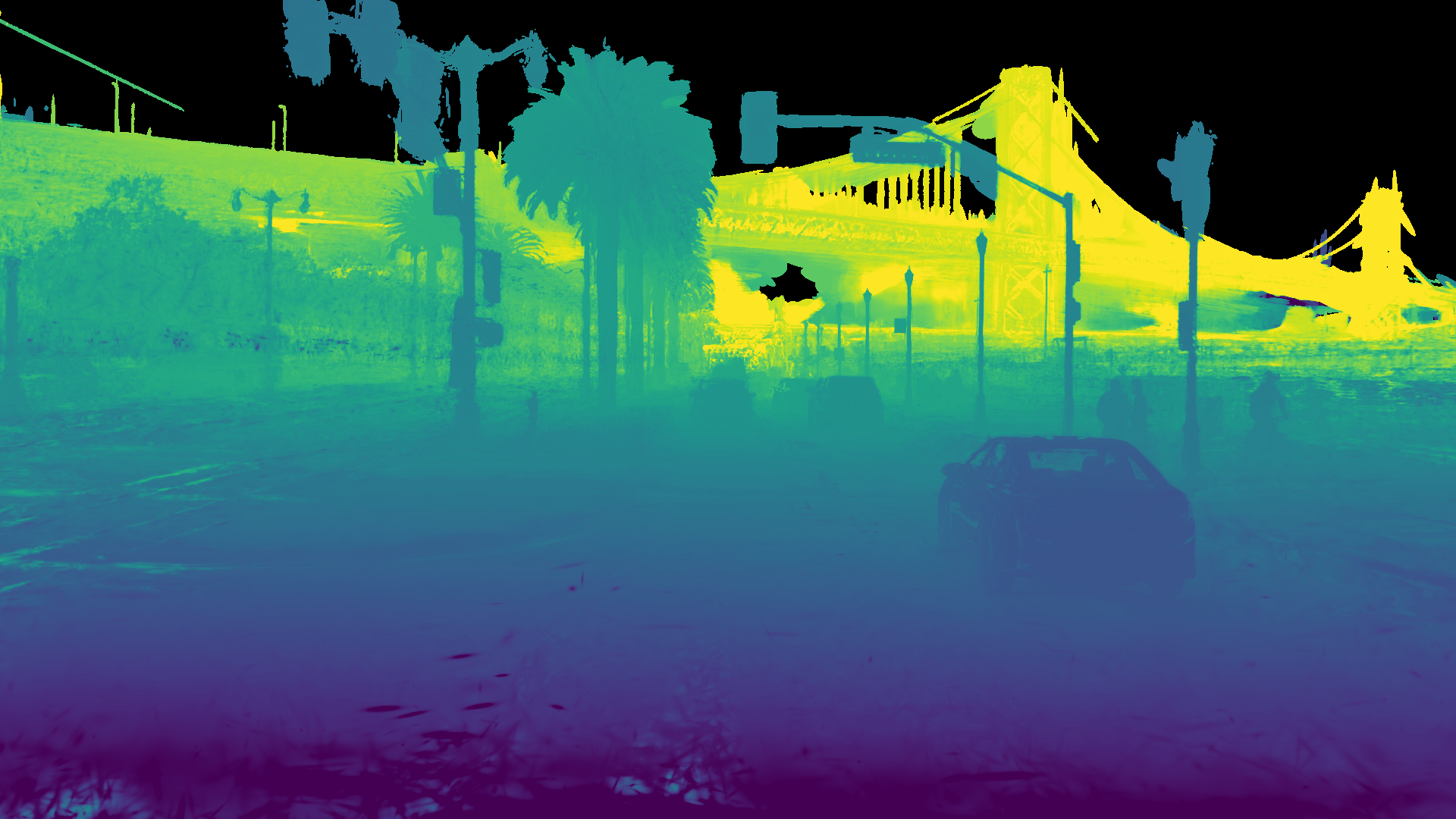} & 
        \includegraphics[clip=false, trim={0 0 0 0},width=\fgsize\columnwidth]{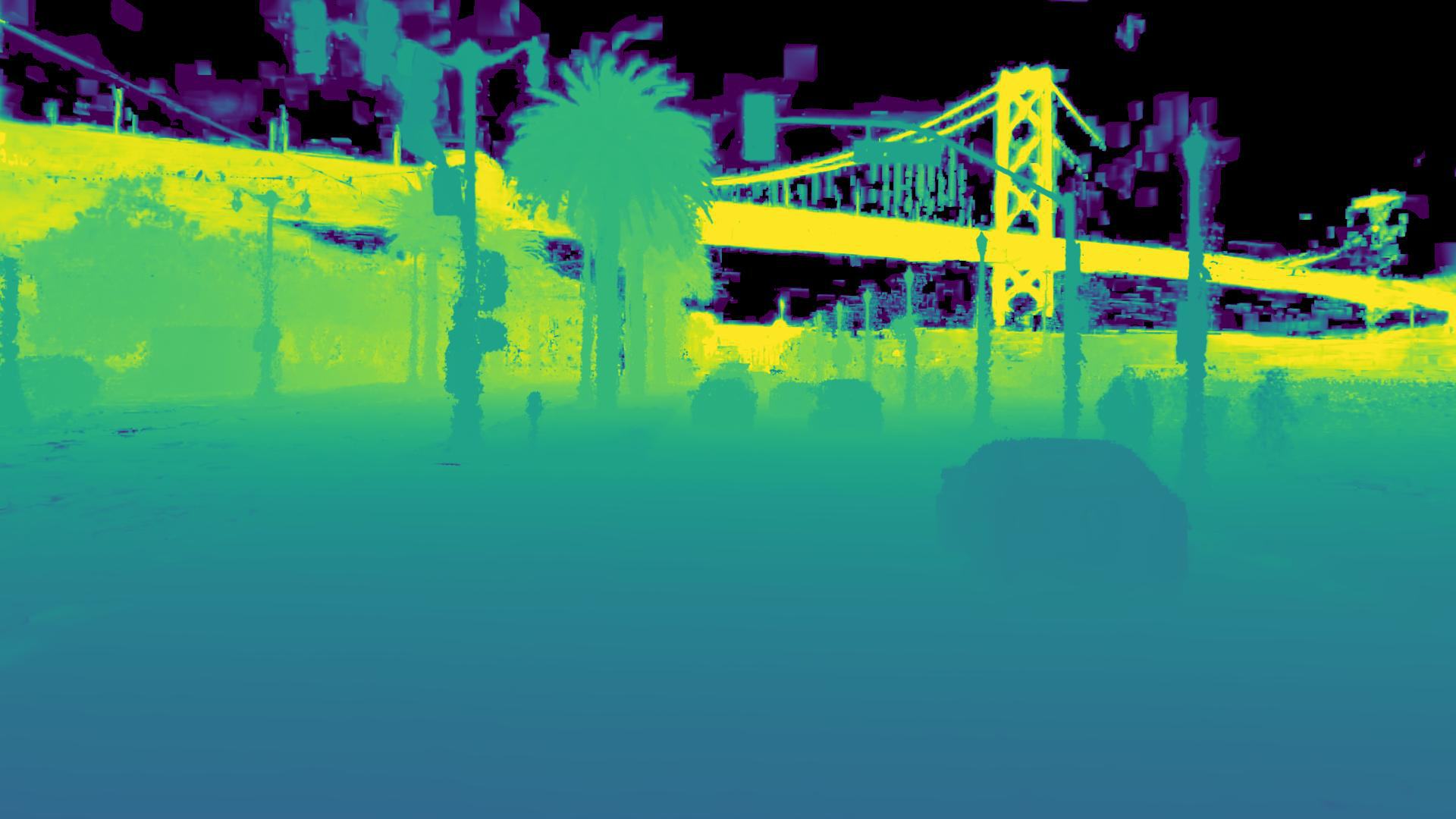} & 
        \includegraphics[clip=false, trim={0 0 0 0},width=\fgsize\columnwidth]{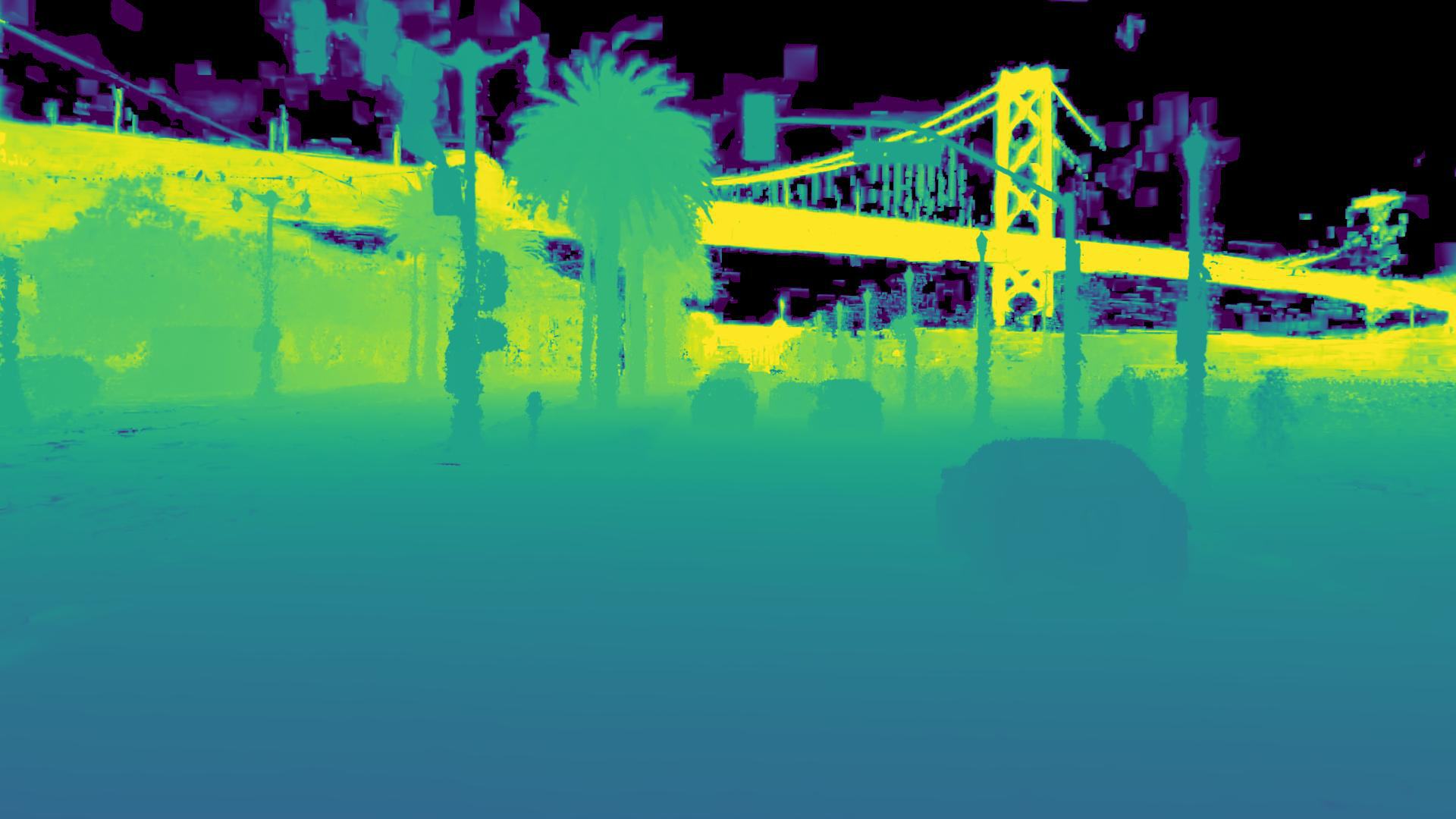} \\      

        \includegraphics[clip=false, trim={0 0 0 0},width=\fgsize\columnwidth]{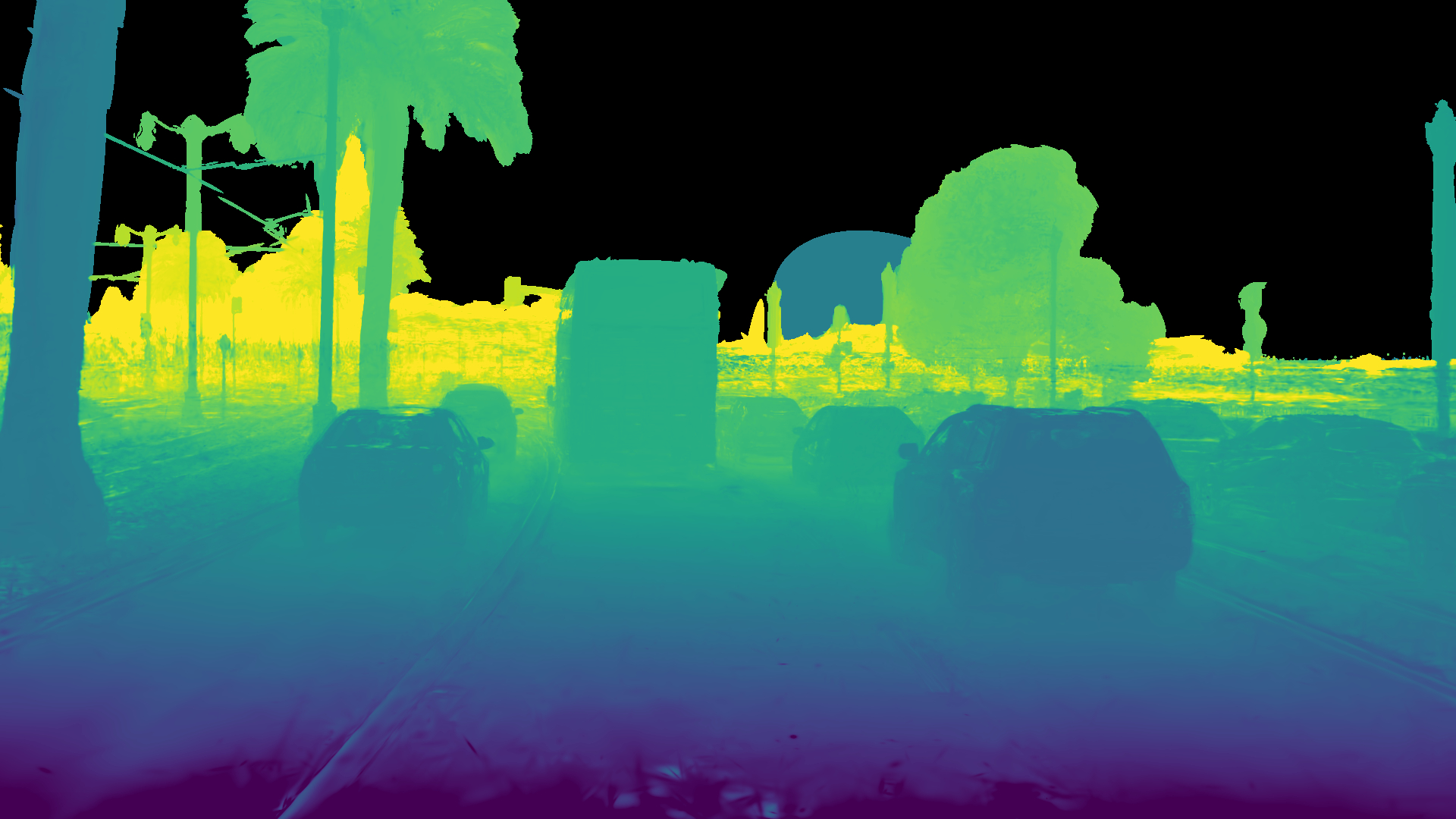} &     
        \includegraphics[clip=false, trim={0 0 0 0},width=\fgsize\columnwidth]{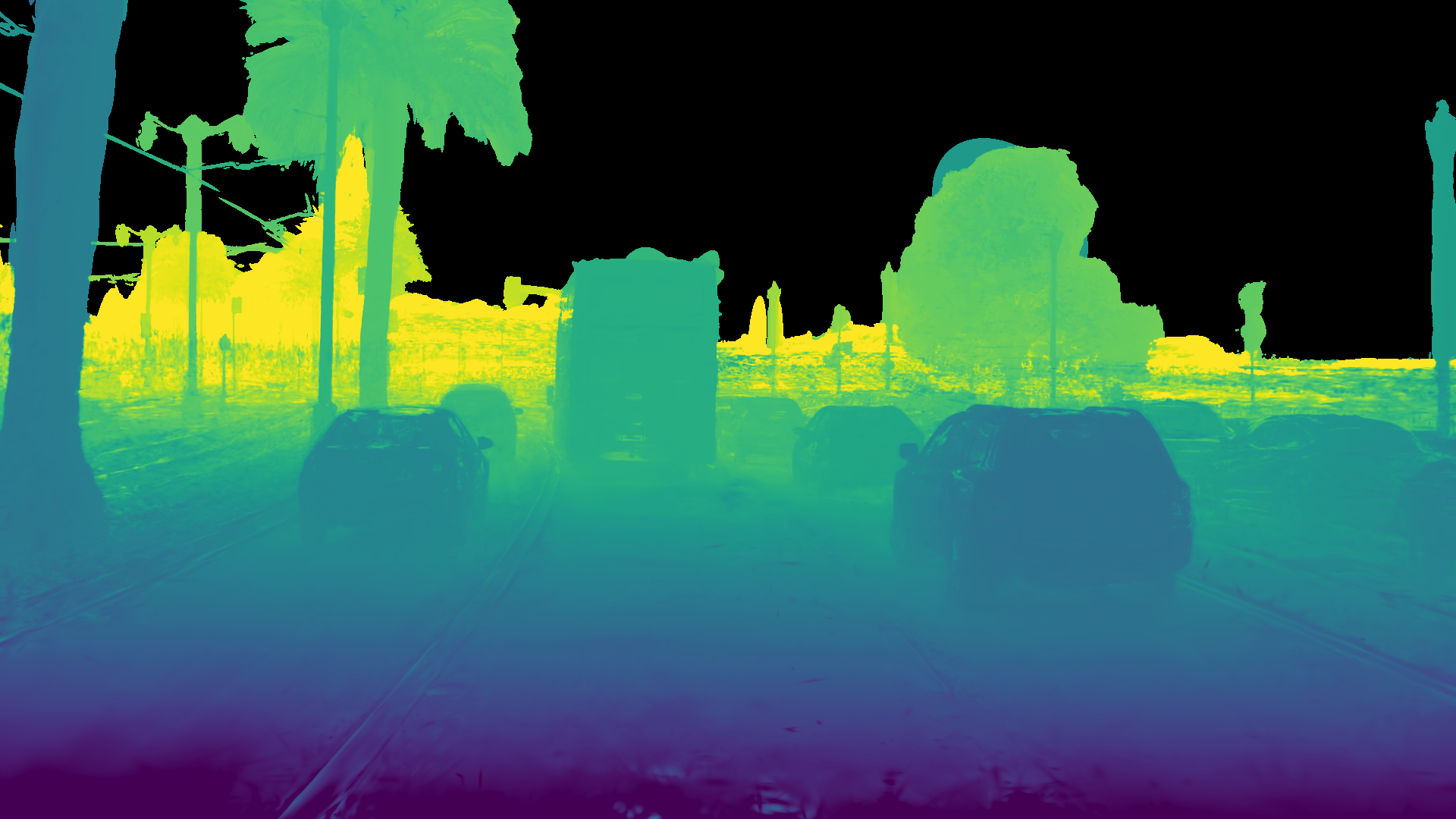} & 
        \includegraphics[clip=false, trim={0 0 0 0},width=\fgsize\columnwidth]{figs/render_results_3x3/scene_16/streetgs_depth_0025.jpg} \\

        \includegraphics[clip=false, trim={0 0 0 0},width=\fgsize\columnwidth]{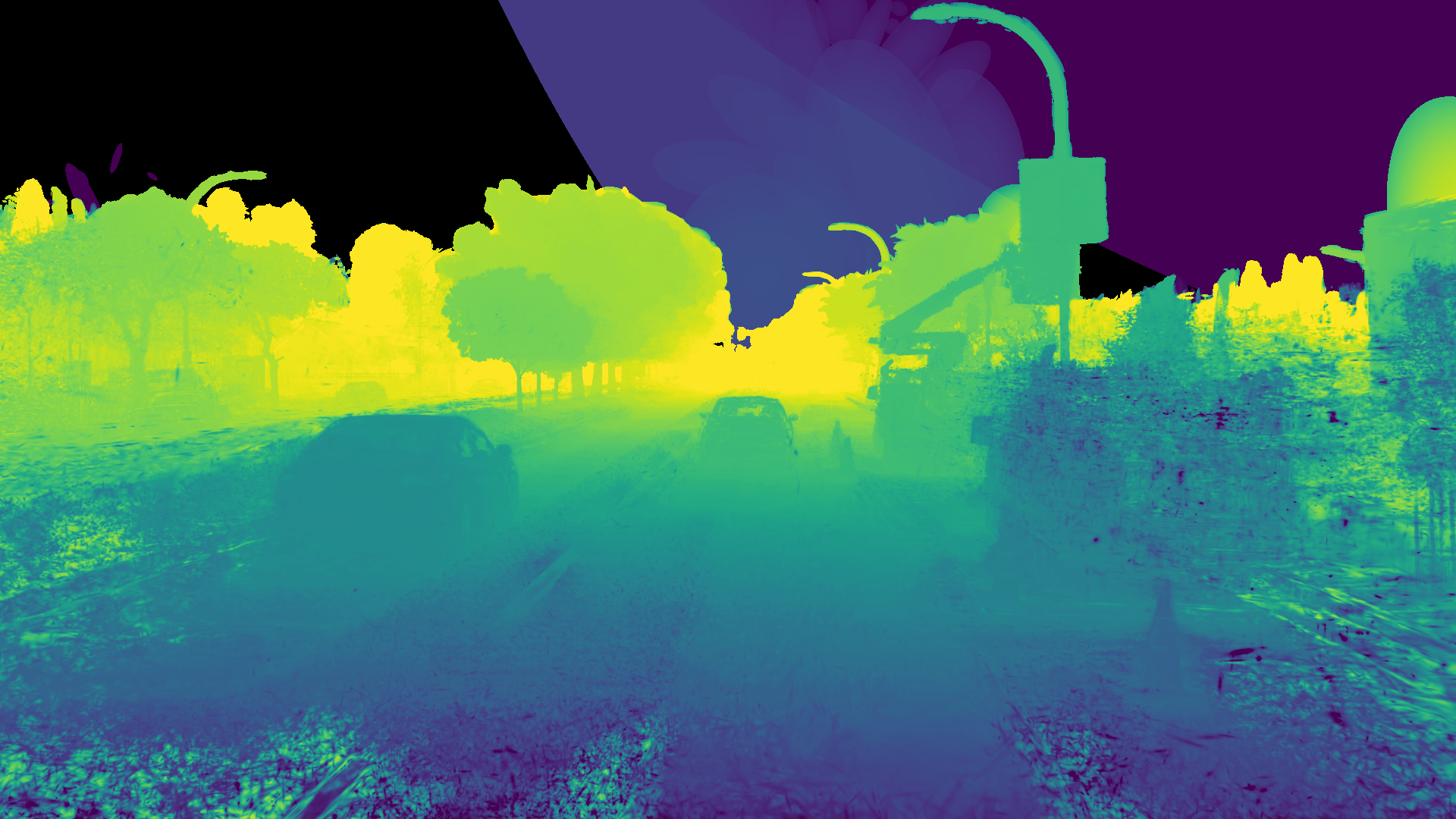} & 
        \includegraphics[clip=false, trim={0 0 0 0},width=\fgsize\columnwidth]{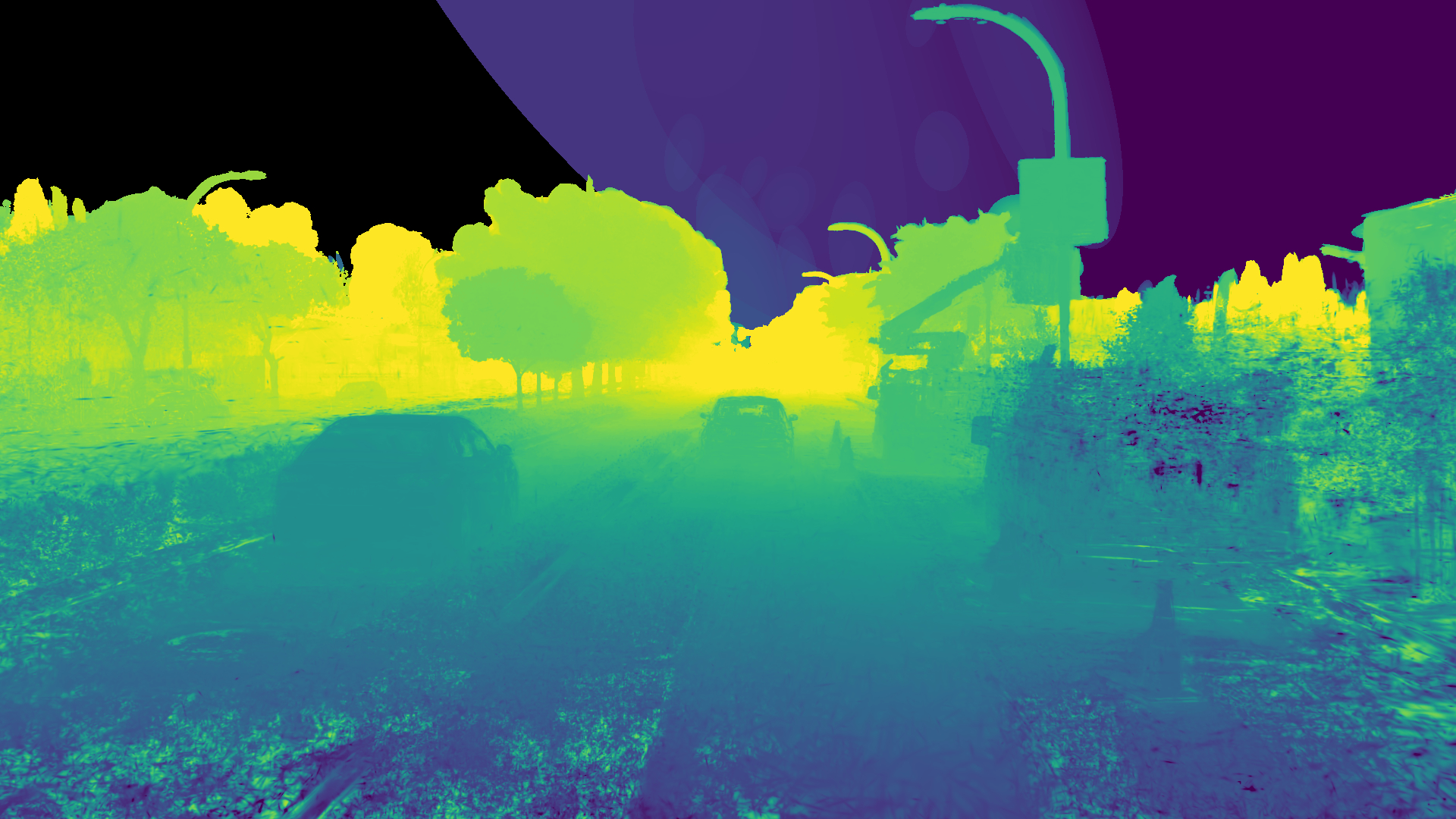} & 
        \includegraphics[clip=false, trim={0 0 0 0},width=\fgsize\columnwidth]{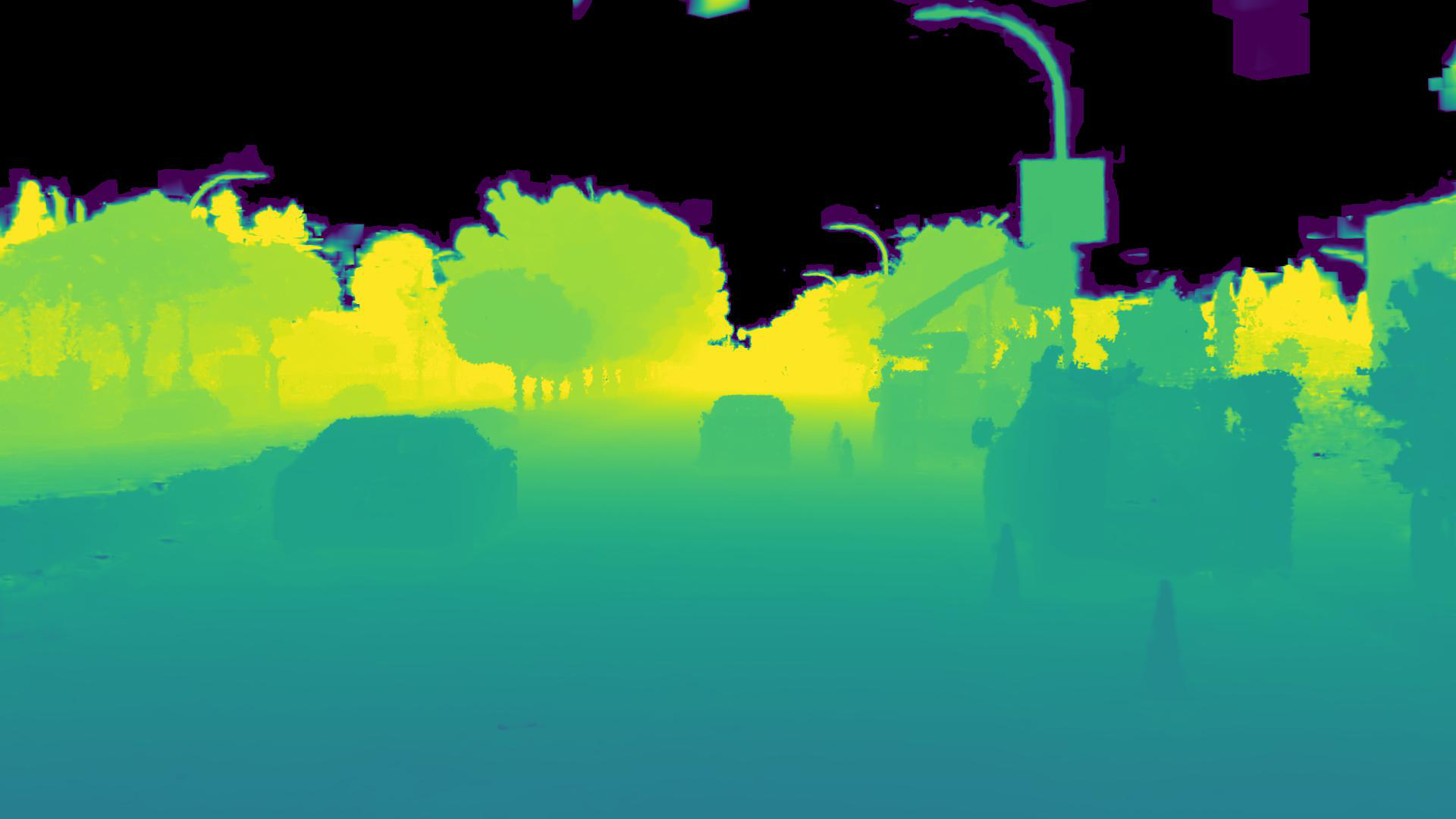} \\

        \includegraphics[clip=false, trim={0 0 0 0},width=\fgsize\columnwidth]{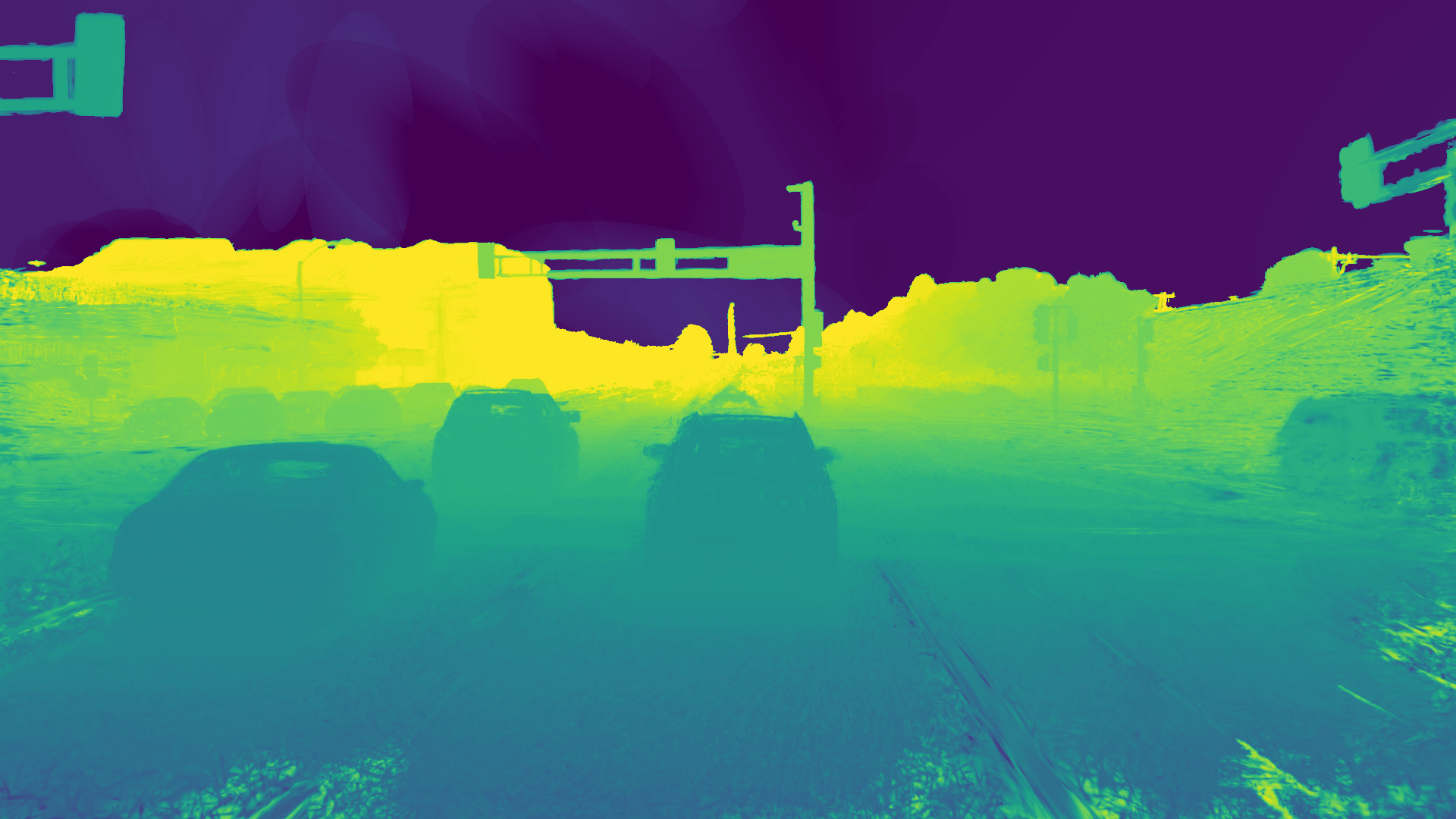} & 
        \includegraphics[clip=false, trim={0 0 0 0},width=\fgsize\columnwidth]{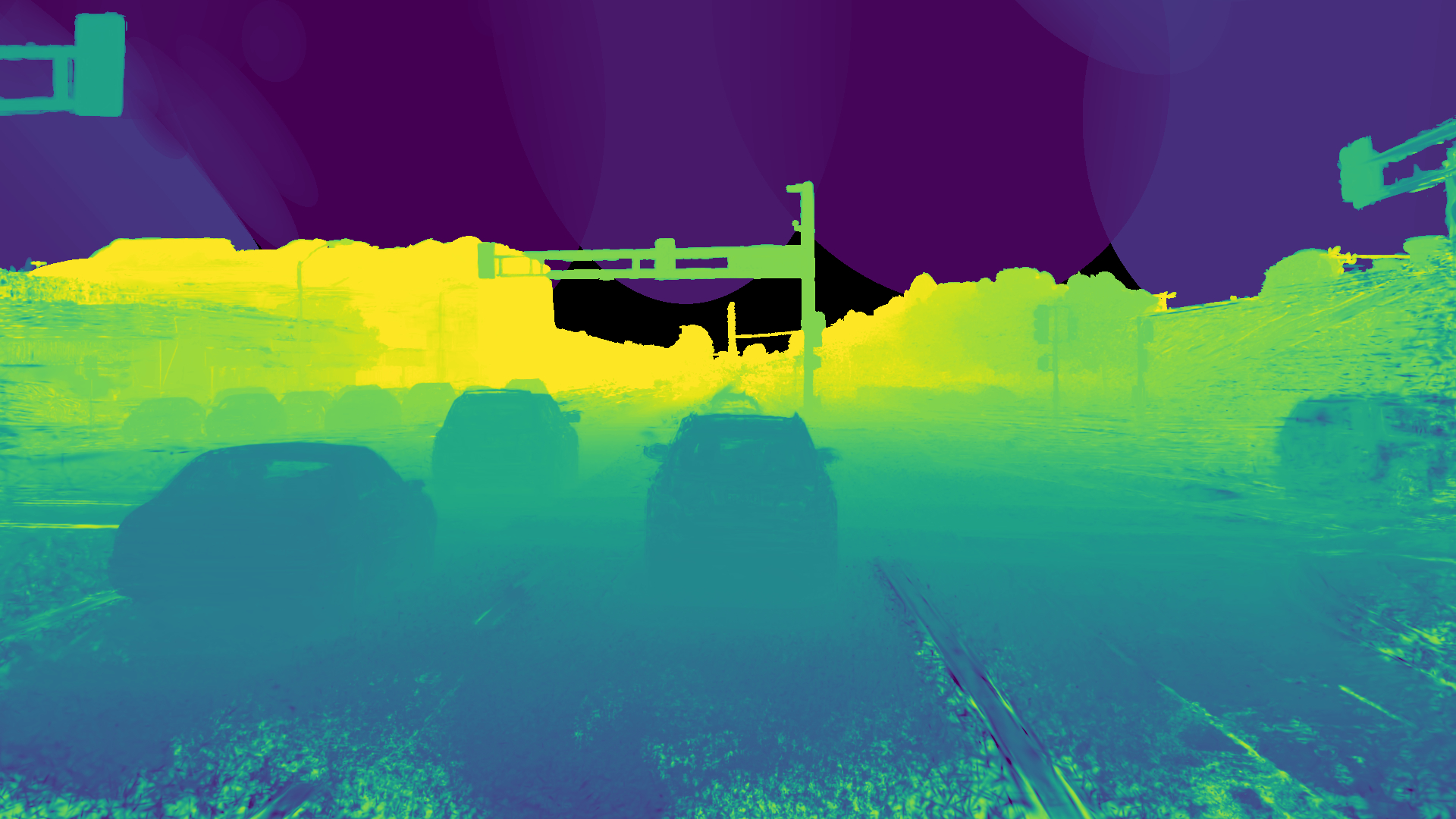} & 
        \includegraphics[clip=false, trim={0 0 0 0},width=\fgsize\columnwidth]{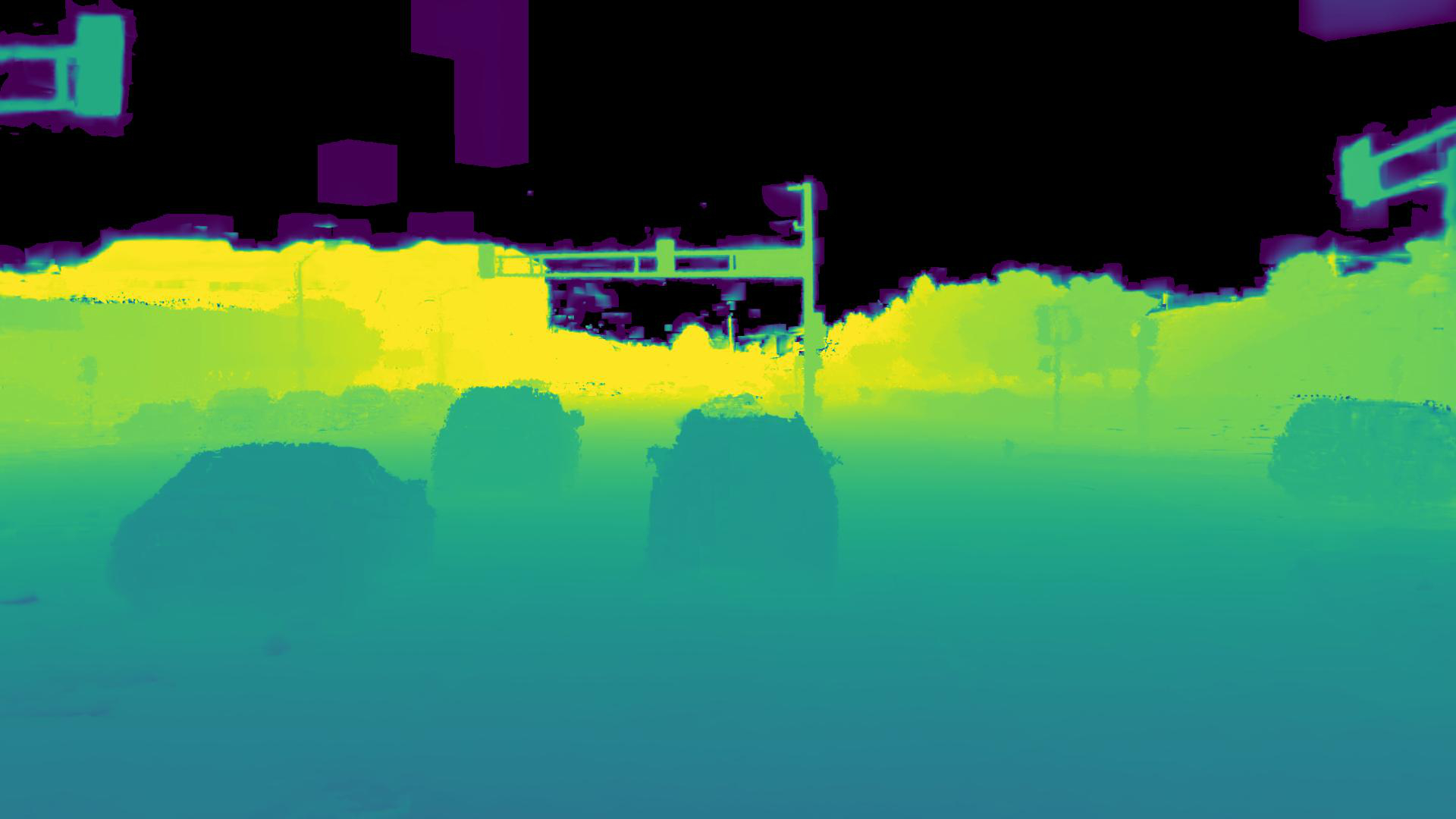} \\

        \includegraphics[clip=false, trim={0 0 0 0},width=\fgsize\columnwidth]{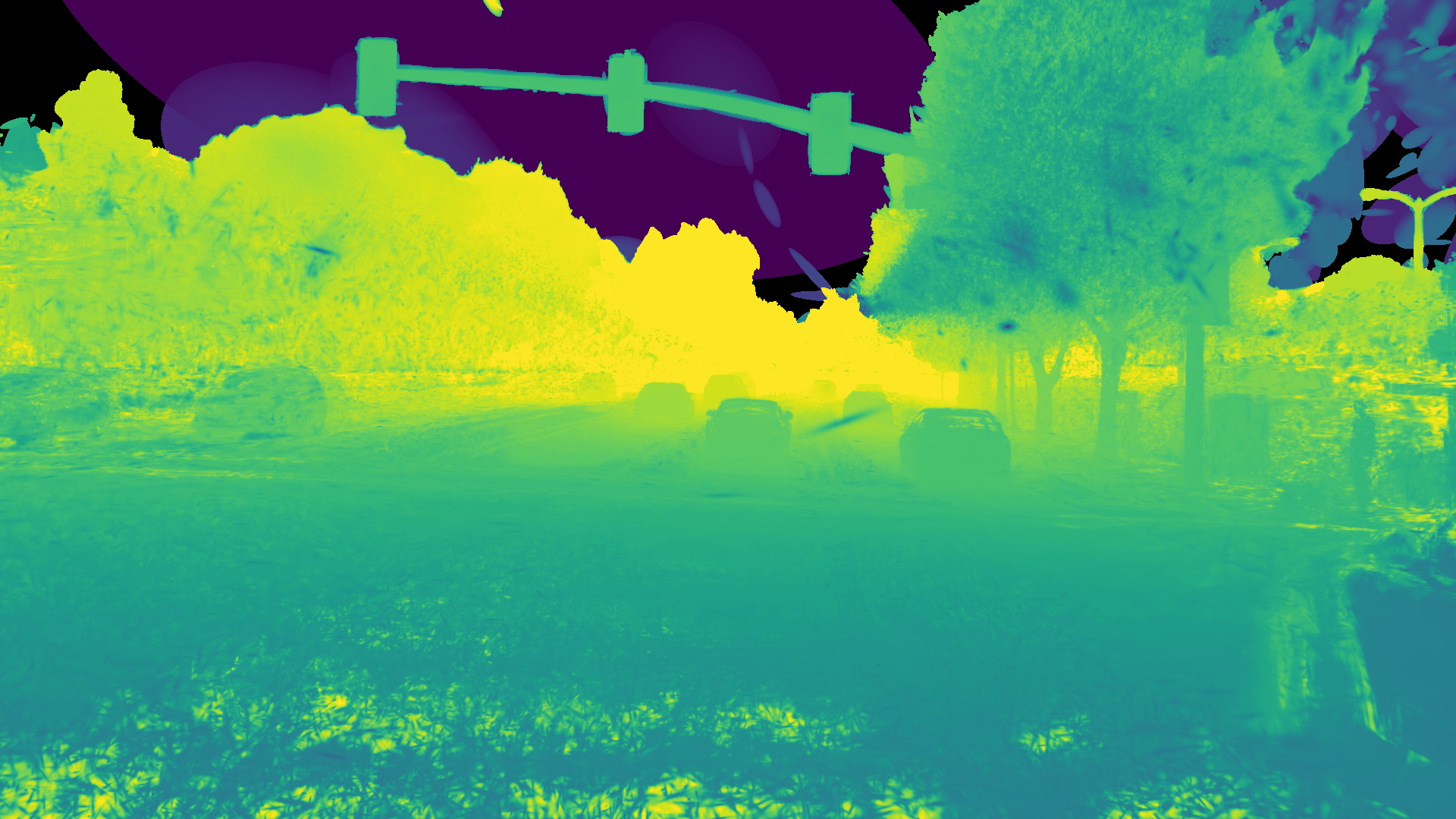} & 
        \includegraphics[clip=false, trim={0 0 0 0},width=\fgsize\columnwidth]{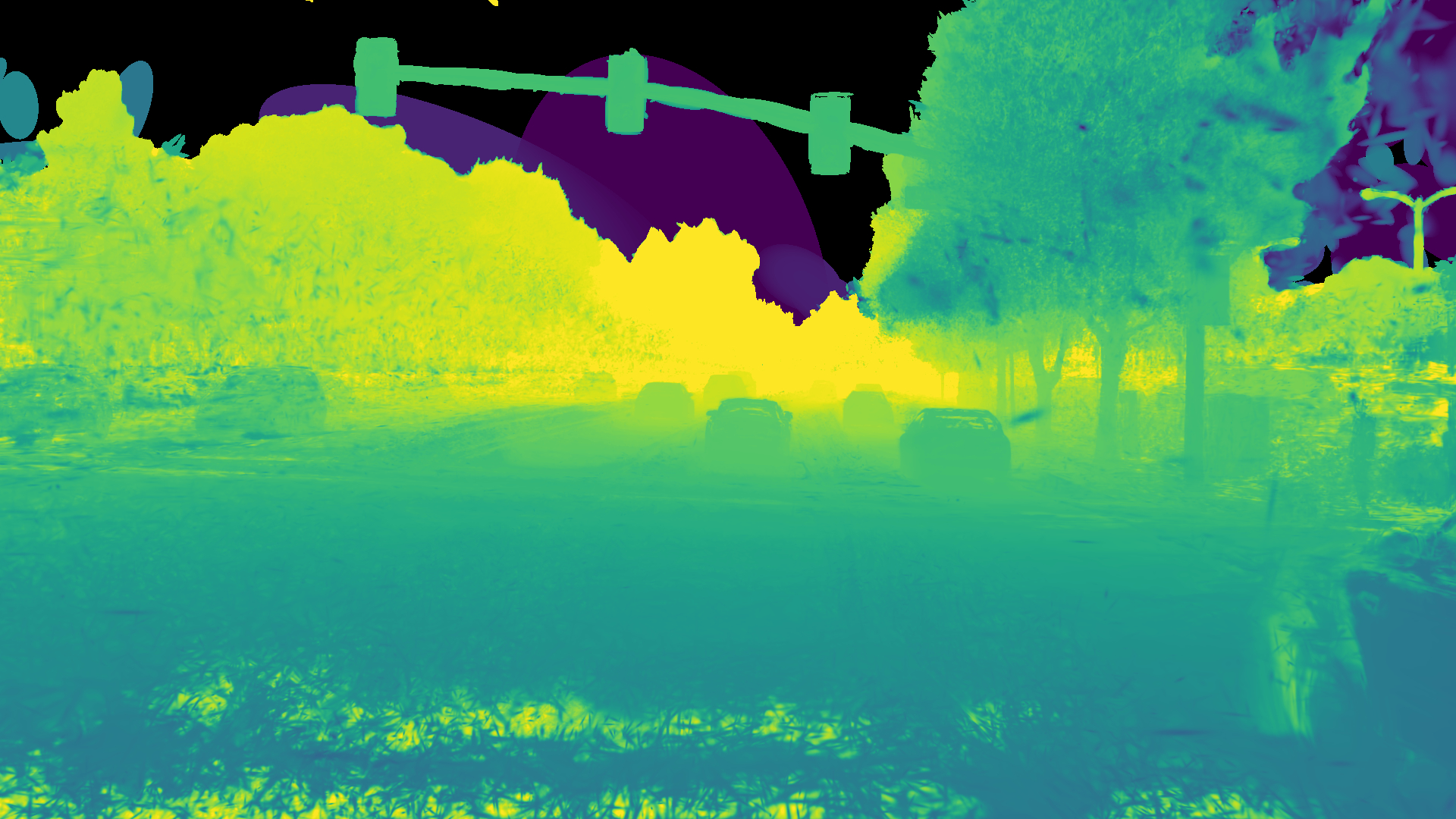} & 
        \includegraphics[clip=false, trim={0 0 0 0},width=\fgsize\columnwidth]{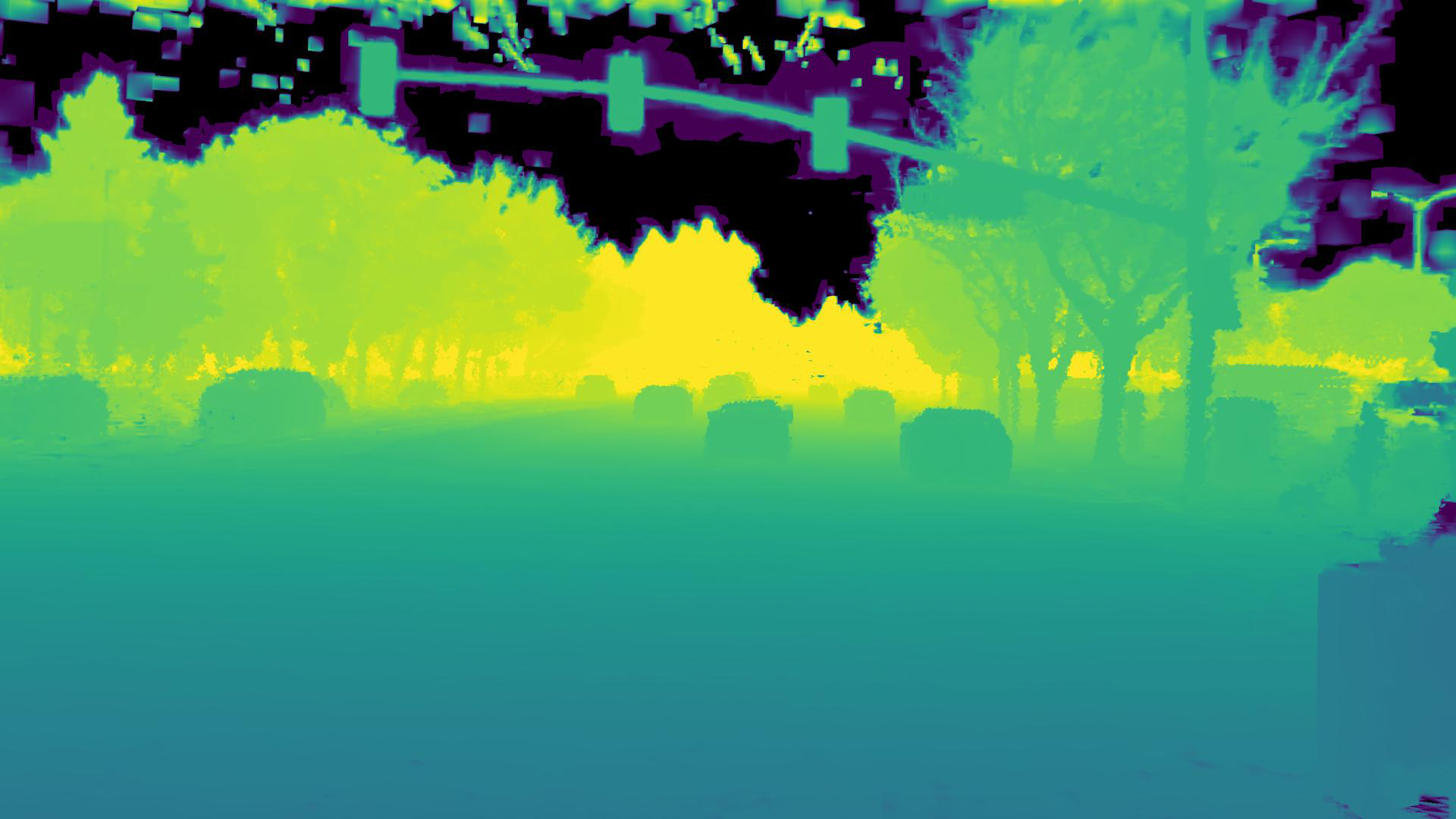} \\
        
    \end{tabular}

    \caption{\textbf{Qualitative Comparison of Reconstructed Depth Maps.} Across multiple PandaSet sequences\cite{xiao2021pandaset}, \method~preserves sharp object boundaries, coherent vehicle geometry, and cleaner scene layout structures compared to point-based baseline alternatives.}
    \label{fig:depth_scenes}
\end{figure*}